\documentclass[nohyperref]{article}

\usepackage{microtype}
\usepackage{graphicx}
\usepackage{subcaption}
\usepackage{booktabs, tabularx} 
\usepackage{bm}
\usepackage{tikz}
\usepackage{pgfplots, import}
\usepackage{pgf}
\usepackage{subcaption, caption}
\usepackage{graphicx}
\usepackage{babel,adjustbox, multirow,makecell}
\usepackage{colortbl}
\usetikzlibrary{matrix}
\usepackage{makecell}

\usetikzlibrary{calc,matrix,positioning, patterns}

\usepackage{hyperref}
\hypersetup{
    colorlinks,
    linkcolor = blue,
    citecolor =blue,
}

\usepackage[accepted]{icml2022}

\usepackage{amsmath}
\usepackage{amssymb}
\usepackage{mathtools}
\usepackage{amsthm}

\usepackage[capitalize,noabbrev]{cleveref}

\theoremstyle{plain}
\newtheorem{theorem}{Theorem}[section]

\newtheorem{lemma}[theorem]{Lemma}

\theoremstyle{definition}

\newtheorem{condition}[theorem]{Condition}
\newtheorem{example}[theorem]{Example}

\theoremstyle{remark}

\newcommand{\wstdmy}[2]{\makecell{#1\small$\pm$#2}} 

\crefname{condition}{condition}{conditions}
\Crefname{condition}{Condition}{Conditions}
\crefname{example}{example}{example}
\Crefname{example}{Example}{Example}
\Crefname{section}{Section}{Section} 
\Crefname{figure}{Figure}{Figure} 
\Crefname{table}{Table}{Table} 
\Crefname{equation}{Equation}{Equation} 
\newcommand{\scatterOP}[0]{\mathcal{I}}
\newcommand{\bx}[0]{\bm{x}}
\newcommand{\xh}[0]{\bm{x}_h}
\newcommand{\xl}[0]{\bm{x}_l}
\newcommand{\xhp}[0]{\bm{x}_h^\prime}
\newcommand{\xlp}[0]{\bm{x}_l^\prime}
\newcommand{\mask}[0]{\bm{M}}
\newcommand{\mi}[2]{I\left(#1;#2 \right)}
\newcommand{\mic}[3]{I\left(#1;#2\middle| #3 \right)}

\newcommand{\added}[0]{\textcolor{black}}

\DeclareMathOperator{\supp}{supp}
\DeclareMathOperator{\acc}{Acc}
\newcommand{\support}[1]{\supp \left\{ #1 \right\}} 

\usepackage{pgf}

\usepackage[textsize=tiny]{todonotes}

\icmltitlerunning{A Consistent and Efficient Evaluation Strategy for Attribution Methods}

\begin{document}

\twocolumn[
\icmltitle{A Consistent and Efficient Evaluation Strategy  \\
           for Attribution Methods}



\icmlsetsymbol{equal}{*}

\begin{icmlauthorlist}
\icmlauthor{Yao Rong}{equal,yyy}
\icmlauthor{Tobias Leemann}{equal,yyy}
\icmlauthor{Vadim Borisov}{yyy}
\icmlauthor{Gjergji Kasneci}{yyy}
\icmlauthor{Enkelejda Kasneci}{yyy}
\end{icmlauthorlist}

\icmlaffiliation{yyy}{Department of Computer Science, University of Tübingen, Tübingen, Germany}

\icmlcorrespondingauthor{Yao Rong}{yao.rong@uni-tuebingen.de}
\icmlcorrespondingauthor{Tobias Leemann}{tobias.leemann@uni-tuebingen.de}

\icmlkeywords{Machine Learning, ICML}

\vskip 0.3in
]



\printAffiliationsAndNotice{\icmlEqualContribution} 

\begin{abstract}
With a variety of local feature attribution methods being proposed in recent years, follow-up work suggested several evaluation strategies. To assess the attribution quality across different attribution techniques, the most popular among these evaluation strategies in the image domain use pixel perturbations. However, recent advances discovered that different evaluation strategies produce conflicting rankings of attribution methods and can be prohibitively expensive to compute. 
In this work, we present an information-theoretic analysis of evaluation strategies based on pixel perturbations. Our findings reveal that the results
are strongly affected by information leakage through the shape of the removed pixels as opposed to their actual values.
Using our theoretical insights, we propose a novel evaluation framework termed Remove and Debias (ROAD) which offers two contributions: First, it mitigates the impact of the confounders, which entails higher consistency among evaluation strategies. Second, ROAD does not require the computationally expensive retraining step and saves up to 99\,\% in computational costs compared to the state-of-the-art. \added{We release our source code at \url{https://github.com/tleemann/road_evaluation}}.

\end{abstract}

\section{Introduction}
Explainable Artificial Intelligence (XAI) has become a widely discussed research topic \citep{adadi2018peeking}. Specifically, feature attribution methods \citep{springenberg2014striving,ribeiro2016should, lundberg2017unified,sundararajan2017axiomatic, selvaraju2017grad} that quantify the importance of input features to a model's decision are widely used. Such local explanations can help to analyze and debug predictive models \citep{bhatt2020explainableMLdeploy, adebayo2020debugging}, e.g., in the medical domain \citep{eitel2019testing}, in recommender systems \citep{afchar2020making}, and many other applications. With an increasing number of feature attribution methods proposed in the literature, the need for sound strategies to evaluate these methods is also increasing \citep{nguyen2020metrics, hase2020evaluating, yeh2019fidelity, Hooker2019ROAR}.

\begin{figure}[t]
\centering
\resizebox{.85\linewidth}{!}{
\tikzset{ 
    table/.style={
        matrix of nodes,
        row sep=-\pgflinewidth,
        column sep=-\pgflinewidth,
        nodes={
            rectangle,
            draw=black,
            align=center
        },
        minimum height=1.5em,
        text depth=0.5ex,
        text height=2ex,
        nodes in empty cells,
        every even row/.style={
            nodes={fill=gray!20}
        },
        column 1/.style={
            nodes={text width=3.2em,font=\bfseries}
        },
        row 1/.style={
            nodes={
                fill=gray,
                text=white,
                font=\bfseries
            }
        }
    }
}
\definecolor{color0}{RGB}{55,162,219}
\definecolor{color3}{RGB}{232,187,44}
\definecolor{color4}{RGB}{25,161,108}
\newcommand{\tabspace}[0]{3.0}
\newcommand{\tabspacenoffset}[0]{0.2}
\newcommand{\outerboxoffset}[0]{5.5}

\begin{tikzpicture}

\matrix (start) [table,text width=3.5em]
{
Rank & 1   & 2 & 3 \\
MoRF  & |[draw,fill=color3!40]| \small{IG} & |[draw,fill=color4!40]| \small{IG-Var} & |[draw,fill=color0!40]| \small{IG-SG}  \\
LeRF  & |[draw,fill=color0!40]| \small{IG-SG} & 
|[draw,fill=color3!40]| \small{IG} &
|[draw,fill=color4!40]|\small{IG-Var}
 \\
};
\node at (\outerboxoffset,-\tabspacenoffset) {\parbox[c]{4cm}{\textbf{Removal evaluation \\ strategy (e.g., ROAR)}
\begin{itemize}
    \setlength\itemsep{-0.3em}
    \item Consistency: \textcolor{red}{Low}
    \item Computation : \textcolor{red}{$\sim$60 min}
\end{itemize}}};




\matrix at (0,-\tabspace) (debias) [table,text width=3.5em]
{
Rank & 1   & 2 & 3 \\
MoRF  & |[draw,fill=color0!40]|\small{IG-SG} & |[draw,fill=color3!40]|\small{IG} & |[draw,fill=color4!40]|\small{IG-Var}  \\
LeRF   & |[draw,fill=color0!40]|\small{IG-SG} & |[draw,fill=color3!40]|\small{IG} & |[draw,fill=color4!40]|\small{IG-Var}  \\
};

\node at (\outerboxoffset, -\tabspace-\tabspacenoffset) {\parbox[c]{4cm}{\textbf{Debiased removal \\evaluation strategy}
\begin{itemize}
    \setlength\itemsep{-0.3em}
    \item Consistency: \textcolor{blue}{High}
    \item Computation : \textcolor{red}{$\sim$60 min}
\end{itemize}}};

\draw [->] (start.south) to [out=-75,in=75] (debias.north) node[above=2pt,xshift=35pt] {\textbf{debiasing}};

\matrix at (0,-2*\tabspace) (road) [table,text width=3.5em]
{
Rank & 1   & 2 & 3 \\
MoRF  & |[draw,fill=color0!40]| \small{IG-SG} & |[draw,fill=color3!40]| \small{IG} & |[draw,fill=color4!40]| \small{IG-Var}  \\
LeRF  & |[draw,fill=color0!40]| \small{IG-SG} & 
|[draw,fill=color3!40]| \small{IG} &
|[draw,fill=color4!40]|\small{IG-Var}
 \\
};

\draw [->] (road.north) to [out=75,in=-75] (debias.south) node[above=-20pt,xshift=36pt] {\textbf{agrees with}};

\node at (\outerboxoffset, -2*\tabspace-\tabspacenoffset) {\parbox[c]{4cm}{\textbf{ROAD (ours)}
\begin{itemize}
    \setlength\itemsep{-0.3em}
    \item No retraining
    \item Consistency: \textcolor{blue}{High} 
    \item Computation : \textcolor{blue}{33 sec}
\end{itemize}}};
\end{tikzpicture}

}
\vspace{-0.2cm}
\caption{Comparison between previous removal and retraining evaluation strategies (\textbf{Top}) and ours (\textbf{Bottom}). Previously, rankings of different attribution methods, Integrated Gradients (IG)\ \citep{sundararajan2017axiomatic} and its two variants SmoothGrad (IG-SG)\ \citep{smilkov2017smoothgrad}, SmoothGrad$^{2}$ (IG-SQ)\ \citep{Hooker2019ROAR}, are highly inconsistent with respect to hyperparameters such as the removal orders Most Relevant First (MoRF) and Least Relevant First (LeRF). Our ROAD strategy achieves a consistent ranking using only 1\% of the previously required resources.}
\label{fig:teaser}
\end{figure}

Evaluation strategies, proposed to compare different attribution methods, commonly follow an ablation approach by perturbing the input features, e.g., image pixels, deemed most or least important. Specifically, perturbing pixels assigned high importance should decrease predictive quality whereas perturbing unimportant pixels, should hardly affect the predictions. 
These measures aim to capture the \textit{fidelity} of explanations \citep{tomsett2020sanity}, i.e., how well the explanation genuinely reflects the prediction of the underlying model. Fidelity based on a single data sample is known as local fidelity, while global fidelity is measured on the whole data set \citep{tomsett2020sanity}.

The outcome of evaluation strategies is highly sensitive to parameters such as the perturbation function and order. Depending on the order chosen, i.e., \textit{most relevant pixels first} or \textit{least relevant pixels first}, such removal strategies often lead to highly contradictory results. For instance, local attribution methods that seem to perform well in one order may perform rather poorly in the other \citep{tomsett2020sanity, haug2021baselines,Hooker2019ROAR}. This inconsistency makes it hard for researchers to impartially compare between different attribution methods and it is not well understood where the inconsistencies stem from. Moreover, for conducting the global fidelity check, a retraining step is required by some methods \citep{Hooker2019ROAR}, which is prohibitively expensive in practice \citep{tomsett2020sanity}. These two drawbacks and our improvements are illustrated in \cref{fig:teaser}.



In this paper, we aim to overcome these shortcomings and make the evaluation more consistent and efficient. To this end, we propose a new debiased strategy that compensates for confounders causing inconsistencies. Furthermore, we show that in the debiased setting, we can skip the retraining without significant changes in the results. This results in drastic efficiency gains as shown in the lower part of \cref{fig:teaser}. We argue that it is crucial for the community to have sound evaluation strategies that do not suffer from limited accessibility due the required compute capacity.
Specifically, we make the following contributions:
\begin{itemize}
    \item We examine the mechanisms underlying the evaluation strategies based on perturbation by conducting a rigorous information-theoretic analysis, and formally reveal that results can be significantly confounded.
    \item To compensate for this confounder, we propose the Noisy Linear Imputation strategy and empirically prove its efficiency and effectiveness. The proposed strategy significantly decreases the sensitivity to hyperparameters such as the removal order. 
    
    \item We generalize our findings to a novel evaluation strategy, ROAD (RemOve And Debias), which can be used to objectively and efficiently evaluate several attribution methods. Compared to previous evaluation strategies requiring retraining, e.g., Remove and Retrain (ROAR) \cite{Hooker2019ROAR}, ROAD saves 99\,\% of the computational costs. 
\end{itemize}






\section{Related Work}
\label{sec:related_work}
There is a plethora of works on different explanation techniques \citep{tjoa2020survey}, especially attribution methods that assign importance scores to each input features. Popular approaches have been proposed by \citet{springenberg2014striving,bach2015pixel,ribeiro2016should,kasneci2016licon,sundararajan2017axiomatic,fong2017interpretable,shrikumar2017learning,smilkov2017smoothgrad, petsiuk2018rise, adebayo2018sanity,chen2018shapley,xu2020attribution, covert2021explaining}, and many more.

With the growing number of attribution methods, various scholars have presented desiderata that explanations should fulfill \citep{bhatt2020evaluating, nguyen2020metrics, fel2021good, afchar2021towards, nauta2022anecdotal}.  \citet{doshi2017towards} consider two subcategories in this field, namely \textit{human-grounded} metrics relying on human judgment and \textit{functional-grounded} metrics. The latter do not require a human-generated ground truth \added{that can be hard or even impossible to obtain. Metrics of this type frequently} rely on the idea that if the most important part of the image is changed, the output probability of the given black-box model should also change in return. Examples include the Sensitivity-n measure proposed by \citet{ancona2017towards} and the infidelity and max-sensitivity metrics by \citet{yeh2019fidelity}. \citet{samek2016evaluating} and \citet{petsiuk2018rise} also propose to perturb the pixels in the input image according to the importance scores. However, \citet{Hooker2019ROAR} show that the perturbation introduces artifacts and results in a distribution shift, putting these no-retraining approaches in question. They propose the Remove and Retrain (ROAR) framework with an extensive model retraining step to adapt to the distribution shift. Therefore, we distinguish between evaluation methods with \textit{retraining} and \textit{no-retraining} approaches. ROAR has been adopted in several recent studies \citep{hartley2020explaining,izzo2020baseline,meng2021mimic, schramowski2020making,srinivas2019full} and variations are being proposed in concurrent work \cite{shah2021input}. 

Only few papers have used and compared different evaluation strategies for attribution methods and a sound theoretical explanation for the differences between them is still missing. \added{\citet{sturmfels2020visualizing} assess different baselines for feature attribution applying the Integrated Gradient method \citep{sundararajan2017axiomatic}.} They also observe that changing the hyperparameter settings can lead to varying results. \citet{haug2021baselines} draw the same conclusion for attributions on tabular data. \citet{tomsett2020sanity} compute the consistency among different, no-retraining evaluation strategies and report an alarmingly low agreement. In this work, we conduct a rigorous analysis of reasons for existing inconsistency and provide a solution to reduce it, which is not studied in previous works. Moreover, our solution also reduces high computational costs caused by retraining.



\section{Preliminaries}
In this section, we formally define the pixel-perturbation strategies considered by the following analysis. 

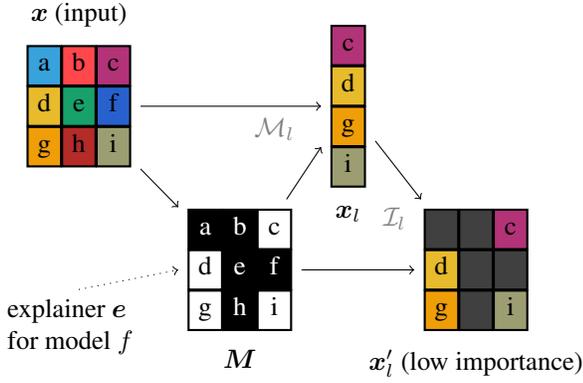
\begin{figure}
    \centering
    \begin{tikzpicture}[
    squarednode/.style={rectangle, draw=black,  thick, minimum size=4mm, text height=.6em,text width=.6em, text centered,text depth=.5ex},
    ]
    
    \newcommand{\cellfc}[0]{darkgray}
    \newcommand{\belowdist}[0]{0.00cm}
    \definecolor{color0}{RGB}{55,162,219}
    \definecolor{color1}{RGB}{255,71,76}
    \definecolor{color2}{RGB}{178,51,120}
    \definecolor{color3}{RGB}{232,187,44}
    \definecolor{color4}{RGB}{25,161,108}
    \definecolor{color5}{RGB}{40,97,205}
    \definecolor{color6}{RGB}{240,158,7}
    \definecolor{color7}{RGB}{180,43,40}
    \definecolor{color8}{RGB}{156,158,114}

    \matrix (m1) at (0,0) [matrix of nodes,nodes={squarednode},column sep=-\pgflinewidth, row sep=-\pgflinewidth]{
        |[draw,fill=color0]| a & |[draw,fill=color1]| b & |[draw,fill=color2]| c \\
        |[draw,fill=color3]| d & |[draw,fill=color4]| e & |[draw,fill=color5]| f \\
        |[draw,fill=color6]| g & |[draw,fill=color7]| h & |[draw,fill=color8]| i \\
        };
    \node [above =\belowdist of m1]{$\bm{x}$ (input)};    
    \node (explan)[below =1.5 cm of m1]{\parbox[c]{1.9cm}{\raggedright explainer $\bm{e}$ for model $f$}}; 
    
    \matrix (m2) [below right = 0.3cm and 0.5cm of m1, matrix of nodes,  nodes={squarednode}, column sep=-\pgflinewidth, row sep=-\pgflinewidth]{
    |[draw,fill=black]| \textcolor{white}{a} & |[draw,fill=black]| \textcolor{white}{b}  & |[draw]| c \\
    |[draw]| d & |[draw,fill=black]| \textcolor{white}{e}  & |[draw,fill=black]| \textcolor{white}{f}  \\
    |[draw]| g & |[draw,fill=black]| \textcolor{white}{h}  & |[draw]| i \\
    };
    \node [below =\belowdist of m2]{$\mask$};
    
    \matrix (mxl) [right= 2.4cm of m1, matrix of nodes,  nodes={squarednode}, column sep=-\pgflinewidth, row sep=-\pgflinewidth]{
    |[draw,fill=color2]| c\\
    |[draw,fill=color3]| d\\
    |[draw,fill=color6]| g\\
    |[draw,fill=color8]| i\\
    };
    \node [below = \belowdist of mxl]{$\bm{x}_l$};
    
    \matrix (mxlp) [right = 1.5cm of m2, matrix of nodes,  nodes={squarednode}, column sep=-\pgflinewidth, row sep=-\pgflinewidth]{
        |[draw,fill=\cellfc]| \textcolor{\cellfc}{0} & |[draw,fill=\cellfc]| \textcolor{\cellfc}{1} & |[draw,fill=color2]| c \\
        |[draw,fill=color3]| d & |[draw,fill=\cellfc]| \textcolor{\cellfc}{4} & |[draw,fill=\cellfc]| \textcolor{\cellfc}{5}  \\
        |[draw,fill=color6]| g & |[draw,fill=\cellfc]| \textcolor{\cellfc}{7} & |[draw,fill=color8]| i \\
    };
    \node [below =\belowdist of mxlp]{$\bm{x}_l^\prime$ (low importance)};
    
    \node at (4.2, -1.5) {\textcolor{gray}{$\scatterOP_l$}};
    \node at (2.6, -0.3) {\textcolor{gray}{$\mathcal{M}_l$}};
    
    \draw[->, dotted](explan.north) -- (m2.west);
    \draw[->](m1) -- (m2);
    \draw[->](m1) -- (mxl);
    \draw[->](m2) -- (mxl);
    \draw[->](m2) -- (mxlp);
    \draw[->](mxl) -- (mxlp);
    \end{tikzpicture}
    \vspace{-0.2cm}
    \caption{Our analytical model of feature removal evaluation (MoRF order shown): The input image $\bm{x}$ (9 pixels a--i) is explained by an explanation method that returns a mask $\mask$ indicating important pixels (black). The remaining, less important pixel values $\xl$ can be extracted from the image using the masking operator $\mathcal{M}_l$ and transformed via the imputation operator $\scatterOP_l$ to an imputed variant of the input $\xlp$, which determines the evaluation outcome. This model allows to separate the information in the feature \textit{values} from that contained in the binary mask $\mask$. \label{fig:functionalsetup}}
\end{figure}

\subsection{Retraining Evaluation Strategies}
We consider a pixel removal strategy, where pixels are successively replaced by imputed values.
Consistent with the literature \citep{tomsett2020sanity,samek2016evaluating}, we consider two removal orders: \textbf{MoRF} (Most Relevant First) or \textbf{LeRF} (Least Relevant First), where the subsequent removal starts with the most important pixels for the former and the least important ones for the latter. We now provide a formal definition of MoRF with retraining, i.e., the ROAR benchmark, that will be used throughout our analysis. We always use the MoRF order in the analysis presented in this paper. However, an analogous analysis of its counterpart LeRF is possible without much additional effort and can be found in the appendix.

To ease our derivations, we describe the procedure by a series of operations that can be analyzed independently. A classifier $f: \mathbb{R}^d \rightarrow \left\{1,\ldots,c\right\}$ maps inputs $\bm{x} \in \mathbb{R}^d$ to labels $C \in \left\{1,\ldots,c\right\}$, where $c$ is the number of classes. A feature attribution explanation for the prediction assigns each input dimension an importance value. In the MoRF setting, the features are ordered in a descending order of importance. Subsequently, the $k$ most important features per instance are selected for removal, where $0 \leq k \leq d$ is successively increased during the benchmark. However, for the moment we consider only one fixed value of $k$. Thus, we can model the explanation $\bm{e}_k$ as a choice of features via a binary mask $\mask = \bm{e}_k\left(f, \bm{x}\right) \in \left\{ 0,1 \right\}^d$, with the corresponding value set to one, if the corresponding feature is among the top-$k$, and to zero otherwise. Furthermore, suppose $\mathcal{M}_{l}: \left\{ 0,1 \right\}^d \times  \mathbb{R}^d \rightarrow \mathbb{R}^{d-k}$ to be the selection operator for the \textbf{\underline{l}}east important dimensions indicated in the mask and $\bm{x}_{l}=\mathcal{M}_{l}\left(\bm{M}, \bm{x}\right)$ to be a vector containing only the remaining features as shown in \cref{fig:functionalsetup}. We suppose that the features preserve their internal order in $\bm{x}_{l}$, i.e., features are ordered ascendingly by their original input indices. This definition allows to separately consider the information flow in the feature \textit{mask} $\bm{M}$ and that in the feature \textit{values} $\bm{x}_l$.

The ROAR approach measures the accuracy of a newly trained classifier $f'$ on modified samples $\bm{x}^\prime_l \coloneqq \scatterOP_{l}\left(\bm{M},\bm{x_l}\right)$, where $\scatterOP_{l}: \left\{ 0,1 \right\}^d\times\mathbb{R}^{d-k} \rightarrow \mathbb{R}^{d}$ is an imputation operator that redistributes all inputs in the vector $\bm{x}_{l}$ to their original positions and sets the remainder to some filling value. In the special case of zero imputation, $\bm{x}^\prime_l = \scatterOP_{l}\left(\bm{M}, \mathcal{M}_{l}\left(\bm{M}, \bm{x}\right)\right) = \left(1-\bm{M}\right) \odot \bm{x}$. This means the top-$k$ features are discarded. For a better evaluation result, the accuracy should drop quickly with increasing $k$, indicating that the most influential features were successfully removed. 
\begin{table}[t]
    \centering
    \resizebox{.5\textwidth}{!}{ 
    \begin{tabular}{r|l}
        $C$ & Class label random variable \\
        $I$ & Mutual information \\
        $\scatterOP$ & Imputation operator \\ 
        $\mask$ & Binary mask in $\left\{0,1\right\}^d$\\
        $\mathcal{M}$ & Mask selection operator (takes out relevant features)\\$\bm{x}$ & Input features in $\mathbb{R}^d$ \\
        $\xl$ & \added{Low importance features only in $\mathbb{R}^{d-k}$} \\
        $\xlp$ & Imputed low importance features in $\mathbb{R}^{d}$ \\
    \end{tabular}
    }
    \caption{Overview of the notation used in this work.\label{tab:notation}}
\end{table}
\subsection{Information Theory}
We now briefly revisit the central concepts of information theory that will be handy for our analysis and introduce the notation. The fundamental quantity in information theory is the entropy $H$ of a discrete random variable $X$ with support $\support{X}$,
\begin{equation}
    H(X) \vcentcolon= -\sum_{x\in \support{X}} P(X=x)\log P(X=x).
\end{equation}
The entropy corresponds to the information gained through observation of a realization of this variable. If the random variable considered can be easily inferred, we use $p(x)$ as a shorthand for $P(X=x)$. Furthermore, we denote the joint entropy between random variables $X$ and $Y$ by $H(X,Y)$, which is equivalent to the entropy of their joint distribution. In accordance with \citet{Cover2006}, we always separate random variables by comma to denote the joint distribution of multiple of variables.

The conditional entropy $H(X|Y)$ is the expected amount of information left in a variable, given the observation of a condition $Y$.
The most central concept in our analysis will be mutual information (MI), i.e., the amount of information in one random variable shared with another. For example, by $I(\bm{x};C) \coloneqq H(C) - H\left( C \middle| \bm{x} \right)$,
we denote the MI between the complete feature vector and the class variable $C$. We separate arguments by a semicolon and allow single random variables or sets of random variables as arguments to all the defined quantities. For sets, we always consider the joint distribution of their member variables. Please confer \citet{Cover2006} for a more profound introduction. We provide a short overview of our notation in \cref{tab:notation}.

\section{Analysis}
\label{sec:analysis}
\begin{figure}[t!]
  \centering
  \includegraphics[width=0.4\textwidth]{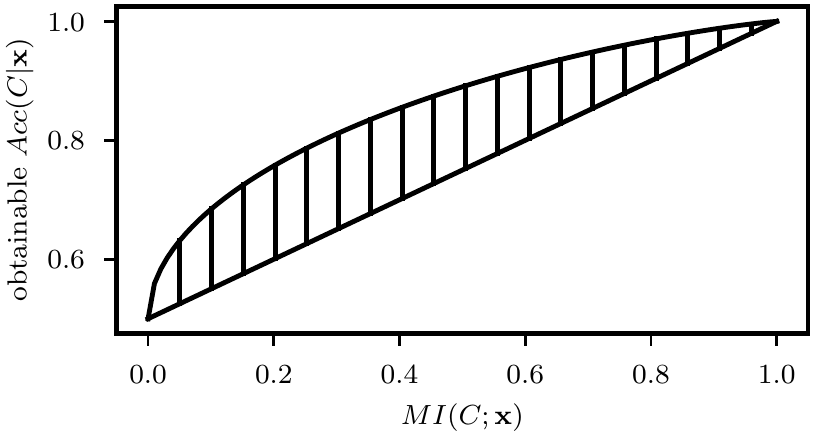}
  \vspace{-0.2cm}
  \caption{Relation between Mutual Information (MI) and obtainable accuracy for the two-class problem with equal class priors. The knowledge of the MI $I(\bx;C)$ implies strong bounds for the obtainable accuracy. This connection permits to use MI as a surrogate for the obtainable accuracy in the perturbation strategy in our analysis. Figure adapted from \citet{Meyen2016masterThesis}.\label{fig:accandmi}
  }
  \vspace{-0.2cm}
\end{figure}
In this section, we show that the pixel perturbation strategies are susceptible to a previously unknown confounder: The binary mask itself can leak class information that might in not be present in the feature values. After making the connection between the accuracy and mutual information as a theoretical tool in \cref{sec:mivsacc}, we formally derive the confounder and identify this leakage on real data in \cref{sec:maskbias}. We subsequently show how to mitigate it through Minimally Revealing Imputation in \cref{sec:biasreduction}.

\subsection{On the Relation Between Accuracy and Mutual Information}
\label{sec:mivsacc}
To begin our analysis of the presented strategies and their underlying mechanisms, we first establish the relation between classification accuracy and the mutual information. It is well-known that the classification performance of an optimal classifier in the Bayesian sense (assigning the class with the highest posterior) is dependent on the MI between features and labels \citep{Hellman1970perror, Vergara2014reviewMIfeatureSelection, Meyen2016masterThesis}. Nevertheless, the relationship is not a function, but comes in form of upper and lower bounds of the obtainable accuracy. 
For the simple two-class problem, the bounds are shown in \cref{fig:accandmi} (cf. \Cref{sec:formulationbounds} for derivations). They impose strong limits on the optimal classification performance, if the mutual information $I(\bx;C)$ is known.

For the pixel removal strategies that use retraining, this allows us to analyze the frameworks using MI as a surrogate for the attainable accuracy because higher MI almost always leads to higher accuracy.
In the MoRF setting with retraining, $I(\xlp; C)$ will play a key role, because it quantifies the information left in the least important features and thus determines obtainable accuracy which is the outcome of the evaluation. Low mutual information $I(\xlp; C)$ results in a sharp drop in accuracy and good benchmarking results:
\[
\downarrow I(\xlp; C)~~\Rightarrow~~\uparrow \text{MoRF benchmark.}
\]
Therefore, in the MoRF setting low mutual information of $\xlp$ and $C$ is desirable\footnote{In LeRF, a higher accuracy and thus higher $I(\xlp; C)$ is beneficial}.
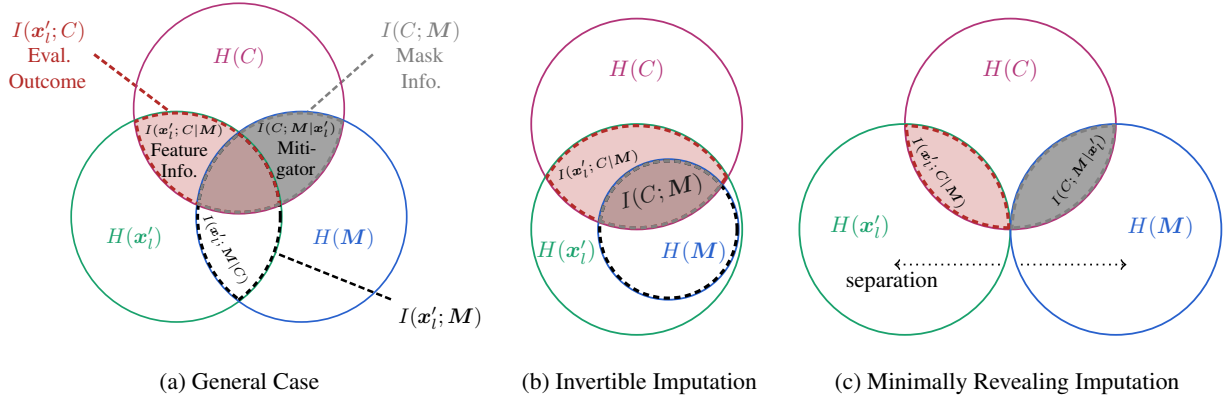
\begin{figure*}[h!]
\begin{center}
    \begin{subfigure}[b]{0.42\textwidth}
    \centering
    \scalebox{0.8}{
\begin{tikzpicture}
\newcommand{\scale}[0]{0.6}
\newcommand{\circrad}[0]{1.75}
\newcommand{\thickwidth}[0]{1.0mm}
\newcommand{\centerx}[0]{1.73}
\definecolor{color0}{RGB}{55,162,219}
\definecolor{color1}{RGB}{255,71,76}
\definecolor{color2}{RGB}{178,51,120}
\definecolor{color3}{RGB}{232,187,44}
\definecolor{color4}{RGB}{25,161,108}
\definecolor{color5}{RGB}{40,97,205}
\definecolor{color6}{RGB}{240,158,7}
\definecolor{color7}{RGB}{180,43,40}
\definecolor{color8}{RGB}{156,158,114}

\begin{scope}
  \clip (0,\scale*2) circle (\circrad); 
  \fill[gray!80,opacity=0.9] (\scale*\centerx,-\scale*1) circle (\circrad);
\end{scope}

\begin{scope}
  \clip (0,\scale*2) circle (\circrad); 
  \fill[color7!50,opacity=0.5] (-\scale*\centerx,-\scale*1) circle (\circrad);
\end{scope}

\node[text width=1.2cm] at (-1.8*\scale, 0.85*\scale) {\parbox{1.5cm}{\centering\tiny$I(\xlp;C|\mask)$\\\small {Feature~~\\Info.~~}}};

\node[text width=1.3cm] at (1.4*\scale, 0.85*\scale) {\parbox{1.5cm}{\centering\tiny$I(C;\mask|\xlp)$\\\small{Miti-\\gator}}};

\node[text width=2cm,  rotate=-60] at (-0.1*\scale, -2.4*\scale) {\tiny$I(\xlp;\mask|C)$};
\node[text width=1.3cm, color=color7] at (-5.3*\scale, 3.5*\scale) (texticx) {\parbox{1.3cm}{\centering $I(\xlp;C)$ \\Eval. Outcome}};

\node[text width=1.3cm, color=gray] at (5*\scale, 3.5*\scale) (texticm) {\parbox{1.3cm}{\centering $I(C;\mask)$ \\Mask Info.}};

\node[text width=1.3cm, color=black] at (5.5*\scale, -3.8*\scale) (textixm) {$I(\xlp;\mask)$};

\draw[dotted,line width=0.5mm, color=color7,dash pattern=on 3pt off 1.5pt] (texticx.east) -- (-1.9*\scale, 1.9*\scale);

\draw[dotted, very thick, color=gray, dash pattern=on 3pt off 1.5pt] (texticm.west) -- (+1.9*\scale, 1.9*\scale);

\draw[dotted, very thick, color=black, dash pattern=on 3pt off 1.5pt] (textixm.north west) -- (1*\scale, -2*\scale);

\draw[color=color2, thick] (0,\scale*2) circle (\circrad) node [label={[shift={(0.0,0.4)}]$H(C)$}] {};
\draw[color=color4, thick] (-\scale*\centerx,-\scale*1) circle (\circrad) node [label={[shift={(-0.7,-0.8)}]$H(\xlp)$}] {};
\draw[color=color5, thick] (\scale*\centerx,-\scale*1) circle (\circrad) node [label={[shift={(+0.7,-0.8)}]$H(\mask)$}] {};

\begin{scope}
  \clip (\scale*\centerx,-\scale*1) circle (\circrad); 
  \begin{scope}
  \clip(-\scale*\centerx,-\scale*1) circle (\circrad);
  \draw[dotted, line width=1mm]
  (-\scale*\centerx,-\scale*1) circle (\circrad);
  \draw[dotted, line width=1mm] (\scale*\centerx,-\scale*1) circle (\circrad);
  \end{scope}
\end{scope}


\begin{scope}
  \clip (0,\scale*2) circle (\circrad); 
  \begin{scope}
  \clip(-\scale*\centerx,-\scale*1) circle (\circrad);
  \draw[color7, dotted, line width=1mm]
  (-\scale*\centerx,-\scale*1) circle (\circrad);
  \draw[color7, dotted, line width=1mm] (0,\scale*2) circle (\circrad);
  \end{scope}
\end{scope}

\begin{scope}
  \clip (0,\scale*2) circle (\circrad); 
  \begin{scope}
  \clip(\scale*\centerx,-\scale*1) circle (\circrad);
  \draw[gray, dotted, line width=1mm]
  (\scale*\centerx,-\scale*1) circle (\circrad);
  \draw[gray, dotted, line width=1mm] (0,\scale*2) circle (\circrad);
  \end{scope}
\end{scope}
\end{tikzpicture}}
    \vspace{-0.5em}
    \caption{General Case\label{fig:quantitiesmask}}
    \end{subfigure}
    \begin{subfigure}[b]{0.19\textwidth}
    \centering
    \scalebox{0.85}{
\begin{tikzpicture}
\newcommand{\scale}[0]{0.55}
\newcommand{\circrad}[0]{1.65} 
\newcommand{\circradm}[0]{1.1} 
\newcommand{\thickwidth}[0]{1.0mm}
\newcommand{\centerx}[0]{0}
\newcommand{\centerxm}[0]{0.90}
\definecolor{color0}{RGB}{55,162,219}
\definecolor{color1}{RGB}{255,71,76}
\definecolor{color2}{RGB}{178,51,120}
\definecolor{color3}{RGB}{232,187,44}
\definecolor{color4}{RGB}{25,161,108}
\definecolor{color5}{RGB}{40,97,205}
\definecolor{color6}{RGB}{240,158,7}
\definecolor{color7}{RGB}{180,43,40}
\definecolor{color8}{RGB}{156,158,114}

\begin{scope}
  \clip (0,\scale*2) circle (\circrad); 
  \fill[gray!80,opacity=0.9] (\scale*\centerxm,-\scale*1) circle (\circradm);
\end{scope}

\begin{scope}
  \clip (0,\scale*2) circle (\circrad); 
  \fill[color7!50,opacity=0.5] (-\scale*\centerx,-\scale*1) circle (\circrad);
\end{scope}

\node[text width=1.2cm,  rotate=15] at (-1.2*\scale, 0.8*\scale) {\tiny$I(\xlp;C|\mask)$};

\node[text width=2cm,  rotate=15, color=black!90] at (1.3*\scale, 0.2*\scale) {$I(C;\mask)$};






\draw[color=color2, thick] (0,\scale*2) circle (\circrad) node [label={[shift={(0.0,0.4)}]$H(C)$}] {};
\draw[color=color4, thick] (\scale*-\centerx,-\scale*1) circle (\circrad) node [label={[shift={(-1.1,-0.8)}]$H(\xlp)$}] {};
\draw[color=color5, thick] (\scale*\centerxm,-\scale*1) circle (\circradm) node [label={[shift={(+0.4,-0.8)}]$H(\mask)$}] {};

\begin{scope}
  \clip (\scale*\centerxm,-\scale*1) circle (\circradm); 
  \draw[dotted, line width=1mm]
   (\scale*\centerxm,-\scale*1) circle (\circradm);
\end{scope}

\begin{scope}
  \clip (0,\scale*2) circle (\circrad); 
  \begin{scope}
  \clip(-\scale*\centerx,-\scale*1) circle (\circrad);
  \draw[color7, dotted, line width=1mm]
  (-\scale*\centerx,-\scale*1) circle (\circrad);
  \draw[color7, dotted, line width=1mm] (0,\scale*2) circle (\circrad);
  \end{scope}
\end{scope}

\begin{scope}
  \clip (\scale*\centerxm,-\scale*1) circle (\circradm); 
  \begin{scope}
  \clip (0,\scale*2) circle (\circrad);
  \draw[gray, dotted, line width=1mm]
  (\scale*\centerxm,-\scale*1) circle (\circradm);
  \draw[gray, dotted, line width=1mm]
  (0,\scale*2) circle (\circrad);
  \end{scope}
\end{scope}
\end{tikzpicture}}
    \vspace{-0.5em}
    \caption{Invertible Imputation\label{fig:invertibleimp}}
    \end{subfigure}
    \begin{subfigure}[b]{0.37\textwidth}
    \centering
    \scalebox{0.85}{
\begin{tikzpicture}
\newcommand{\scale}[0]{0.55}
\newcommand{\circrad}[0]{1.65}
\newcommand{\thickwidth}[0]{1.0mm}
\newcommand{\centerx}[0]{3}
\definecolor{color0}{RGB}{55,162,219}
\definecolor{color1}{RGB}{255,71,76}
\definecolor{color2}{RGB}{178,51,120}
\definecolor{color3}{RGB}{232,187,44}
\definecolor{color4}{RGB}{25,161,108}
\definecolor{color5}{RGB}{40,97,205}
\definecolor{color6}{RGB}{240,158,7}
\definecolor{color7}{RGB}{180,43,40}
\definecolor{color8}{RGB}{156,158,114}

\begin{scope}
  \clip (0,\scale*2) circle (\circrad); 
  \fill[gray!80,opacity=0.9] (\scale*\centerx,-\scale*1) circle (\circrad);
\end{scope}

\begin{scope}
  \clip (0,\scale*2) circle (\circrad); 
  \fill[color7!50,opacity=0.5] (-\scale*\centerx,-\scale*1) circle (\circrad);
\end{scope}

\node[text width=1.2cm,  rotate=-55] at (-2.1*\scale, 0.7*\scale) {\tiny$I(\xlp;C|\mask)$};

\node[text width=2cm,  rotate=55] at (2.3*\scale, 1.2*\scale) {\tiny$I(C;\mask|\xlp)$};






\draw[color=color2, thick] (0,\scale*2) circle (\circrad) node [label={[shift={(0.0,0.4)}]$H(C)$}] {};
\draw[color=color4, thick] (\scale*-\centerx,-\scale*1) circle (\circrad) node [label={[shift={(-0.7,-0.4)}]$H(\xlp)$}] {};
\draw[color=color5, thick] (\scale*\centerx,-\scale*1) circle (\circrad) node [label={[shift={(+0.7,-0.4)}]$H(\mask)$}] {};

\draw[->, dotted, thick]  (-0.1*\centerx*\scale,-2*\scale) -- (-1.1*\centerx*\scale,-2*\scale);
\draw[->, dotted, thick] (0.1*\centerx*\scale,-2*\scale)-- (1.1*\centerx*\scale,-2*\scale);
\node[text width=3.5cm] at (-0.5*\centerx*\scale,-2.5*\scale) (textsep) {separation};

\begin{scope}
  \clip (0,\scale*2) circle (\circrad); 
  \begin{scope}
  \clip(-\scale*\centerx,-\scale*1) circle (\circrad);
  \draw[color7, dotted, line width=1mm]
  (-\scale*\centerx,-\scale*1) circle (\circrad);
  \draw[color7, dotted, line width=1mm] (0,\scale*2) circle (\circrad);
  \end{scope}
\end{scope}

\begin{scope}
  \clip (0,\scale*2) circle (\circrad); 
  \begin{scope}
  \clip(\scale*\centerx,-\scale*1) circle (\circrad);
  \draw[gray, dotted, line width=1mm]
  (\scale*\centerx,-\scale*1) circle (\circrad);
  \draw[gray, dotted, line width=1mm] (0,\scale*2) circle (\circrad);
  \end{scope}
\end{scope}
\end{tikzpicture}}
    \vspace{-0.5em}
    \caption{Minimally Revealing Imputation\label{fig:optimal imp}}
    \end{subfigure}
    \vspace{-0.5em}
    \caption{The Evaluation Outcome $I(\xlp; C)$ (red area),  is confounded by the Mask Information $I(C; \mask)$ (gray area) when there is some overlap (a). Only the Feature Information $I(\xlp; C|\mask)$, the part of the Outcome not overlapping (light red area), should actually be assessed. In the worst case (which we term Invertible Imputation), the Mask Information is entirely contained in the Outcome (b). Separating the information in the imputed image $\xlp$ and the mask $\mask$ allows to reduce the overlap and the influence (c). \label{fig:masking}}
\end{center}
\end{figure*}
\subsection{Class Information Leakage through Masking}
\label{sec:maskbias}
We demonstrate that it is easily possible to leak class information only through the mask's shape and to harshly manipulate the evaluation score. Therefore, we start by separating the influence of the mask from that of the feature values. \added{Our derivation relies on the multi-information $I(C; \bm{x}_l^\prime; \bm{M})$, which is defined by \citet{Vergara2014reviewMIfeatureSelection} as follows}:
\begin{align}
I(C; \bm{x}_l^\prime; \bm{M}) &= I(C;\bm{x}_l^\prime| \bm{M}) -I(C;\bm{x}_l^\prime) \label{eqn:2} \\
I(C; \bm{x}_l^\prime; \bm{M}) &= I(C; \bm{M}| \bm{x}_l^\prime) -I(C; \bm{M}). \label{eqn:3}
\end{align}
\added{Setting \cref{eqn:2} and \cref{eqn:3} equal, we arrive at the identity:}
\begin{equation}
\label{eqn:maskgeneral}
   \underbrace{I(\xlp; C)}_\text{Eval. Outcome} = \underbrace{I(C;\xlp|\mask)}_\text{Feature Info.} +  \underbrace{I(C;\mask)}_\text{Mask Info.} - \underbrace{I(C;\mask|\xlp)}_\text{Mitigator}.
\end{equation}

The quantities involved are visualized in \cref{fig:quantitiesmask}. The first term ``Feature Information" is the class information contained in the features (and not in the mask) that we wish to estimate. 
The second term ``Mask Information" shows that class-discriminative information in the mask can have a high impact on the result. This influence can be compensated by the ``Mitigator" term.

\paragraph{Class Information Leakage} If the Mask Information term is superior to the Mitigator, $I(C;\mask) > I(C;\mask|\xlp)$, the evaluation outcome is unfairly increased to a value not justified by the selected features. We term this phenomenon \textit{Class Information Leakage}, as some discriminative information is ``leaked'' through the used binary mask $\mask$.

The Mitigator can entirely vanish when the mask is perfectly inferable from the imputed image $\xlp$. This results in a non-compensated effect of Class Information Leakage. We define this imputation operation as follows:
\begin{condition}
\label{ass:condinvertibleinfilling}
\textit{Invertible Imputation. Let $\scatterOP_l: \left\{ 0,1 \right\}^d \times  \mathbb{R}^{d-k} \rightarrow \mathbb{R}^{d}$ be the imputation operator that takes the least important features as an input. We suppose that there are inverse functions $\scatterOP_{l,M}^{-1}$ and $\scatterOP_{l,x}^{-1}$, such that}
\[
\bm{x}_l^\prime = \scatterOP_l\left(\bm{M}, \bm{x}_l\right) \Leftrightarrow \mask = \scatterOP_{l,M}^{-1}(\bm{x}_l^\prime) \wedge \bm{x}_l = \scatterOP_{l,x}^{-1}(\bm{x}_l^\prime).
\]
\end{condition}

If, for instance, the pixels removed are set to some reserved value indicating their absence, the imputation operator is invertible, as the mask can be reconstructed. Therefore, $H(\mask|\xlp){=} H\left(\scatterOP_{l,M}^{-1}(\xlp)|\xlp\right){=}0$. In this case, also the Mitigator $\mic{C}{\mask}{\xlp} = 0$, because it is bounded by $0 = H(\mask|\xlp) \geq  \mic{C}{\mask}{\xlp} \geq 0$.
The Feature Information term is constrained to be positive. Thus, the Mask Information has a non-negligible impact on the Evaluation Outcome because a higher Mask Information term will always increase it. This case is depicted in \cref{fig:invertibleimp}.

We can create a simple example that shows how evaluation scores are influenced: Imagine a two-class problem that consists of detecting whether an object is located on the left or the right side of an image. A reasonable attribution method masks out pixels on the left or the right depending on the location of the object. In this case, the retraining step can lead to a classifier that infers the class just from the location of the masked out pixels and obtain high accuracy. This explanation map will be rated far worse in MoRF (no accuracy drop) than it might actually be. \added{In the context of amortized explanation methods, a similar  finding has been made by \citet{jethani2021have}. We theoretically showed that this problem also arises in evaluation strategies and empirically demonstrate that the leakage is significant for popular attribution methods on real data in \cref{sec:masking impact}.}


\subsection{Reduction of Information Leakage}
\label{sec:biasreduction}
To tackle this problem, we follow an intuitive approach: If we cannot guarantee that there is no class information contained in the mask itself, we have to stop it from leaking the class information into the imputed images. Therefore, we make sure that the mask used cannot be easily inferred from the imputed image. We would like to set $I(\xlp;\mask)=0$, i.e., the mask is independent of the imputed vector allowing to separate the effects as shown in \cref{fig:optimal imp}. Unfortunately, this is not possible in general: If both should be dependent on the class label, they will also have to share a minimal amount of information (that regarding the class). However, we can demand conditional independence and make $I(\xlp;\mask)$ as small as possible.
\begin{condition} 
\label{ass:minimallyrevealing}
\textit{Minimally Revealing Imputation. Let $\scatterOP_l: \left\{ 0,1 \right\}^d \times  \mathbb{R}^{d-k} \rightarrow \mathbb{R}^{d}$ be the infilling operator that takes the least important features as an input. Suppose $\xlp$ and $\mask$ are independent given the class information $I\left(\xlp;\mask|C\right)=0$ and $I\left(\xlp;\mask \right) \approx 0$}.
\end{condition}
In this case, $I(C;\mask) - \mic{C}{\mask}{\xlp} = I\left(\xlp;\mask \right) - \mic{\xlp}{\mask}{C} \approx 0$,
which implies $I(C;\mask) \approx \mic{C}{\mask}{\xlp}$ (also cf. \cref{fig:optimal imp}), indicating that the Mitigator effectively compensates the Mask Information term.

\section{Debiasing Evaluation Strategies for Local Attribution Methods}
\added{With the theoretical analysis in \cref{sec:analysis}, we can better understand where the biases come from, and thus mitigate them. 
Building on the derivations, we now show the strong impact of the Class Information Leakage introduced in \cref{sec:maskbias} on a real-world data set to highlight the necessity to compensate for this confounder.} We explain how we reduce its influence by proposing a novel imputation operator termed \textit{Noisy Linear Imputation}.

\begin{figure}[t]
    \begin{subfigure}[h]{0.23\textwidth}
        \centering
        \includegraphics[width=0.95\textwidth]{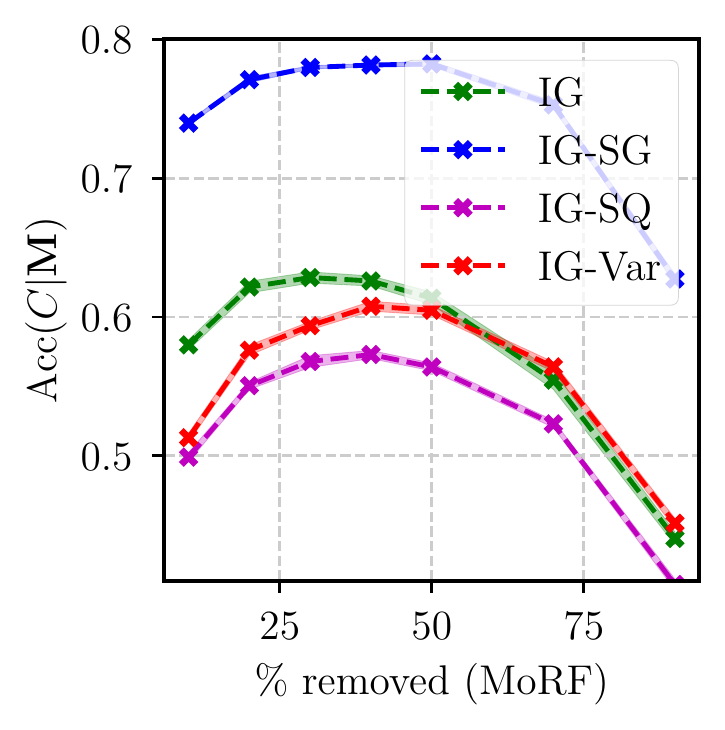}
        \caption{Integrated Gradients \newline (IG)}
    \end{subfigure}
    \hfill
    \begin{subfigure}[h]{0.23\textwidth}
        \centering
        \includegraphics[width=0.95\textwidth]{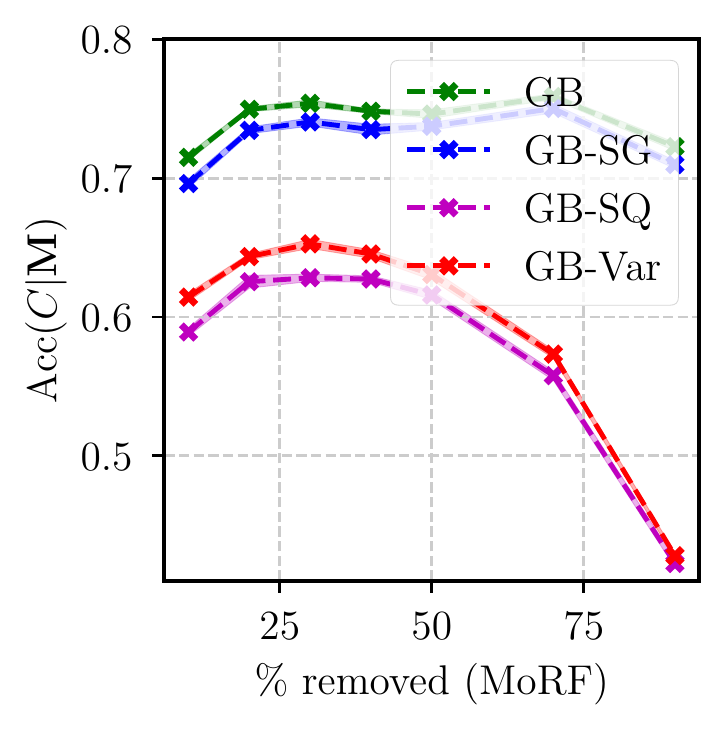}
        \caption{Guided Backpropagation \newline (GB)}
    \end{subfigure}
    \caption{Accuracy of a trained classifier only using the binary masks $\mask$ without feature values as input on the CIFAR-10 data set. Binary masks $\mask$ were computed for different variants of IG and GB. Only the masks contain enough information to reach an accuracy of almost up to 80\,\% (compared to 85\,\% with full images) highlighting that the feature values do not play an important role in the evaluation. This underlines the necessity to compensate for this confounder.}
    \label{fig:bitmask}
\end{figure}
\subsection{Extent of Class Information Leakage}
\label{sec:masking impact}

To empirically confirm our findings, we performed experiments on CIFAR-10  \citep{krizhevsky2009learning}. We use the same attribution methods as in \citet{Hooker2019ROAR}: Integrated Gradients (IG)\ \citep{sundararajan2017axiomatic} and Guided Backprop (GB)\ \citep{springenberg2014striving} serve as base explanations, and three ensembling strategies for each are used in addition: SmoothGrad (SG)\ \citep{smilkov2017smoothgrad}, SmoothGrad$^{2}$ (SQ)\ \citep{Hooker2019ROAR} and VarGrad (Var) \ \citep{adebayo2018sanity}. In total, we consider eight attribution methods and provide details and parameters in the supplementary material.

We empirically show that with fixed value imputation with the global mean, the explanation masks are leaking class information. This takes two steps: \added{(1) We show that the Mask Information $I(C;\mask)$ is extremely high. (2) We verify that the Mitigator is small by testing the \textit{Invertible Imputation} Condition, which implies that class information is leaked into the evaluation outcome through $I(C;\mask)$.}

To assess the class information in the mask, we train a ResNet-18 \citep{he2016deep} that uses only binary masks $\mask$ (no pixel values $\xl$) to predict the class. \added{As we discussed previously, the accuracy of a classifier can be used as a surrogate for the calculation of MI, which is prohibitively expensive for high-dimensional data.} The curves\footnote{Standard Errors are indicated by shaded areas in all figures. However, they are often hardly visible due to their low magnitude.} are shown in \cref{fig:bitmask}. Stunningly, the mask alone results in high accuracy curves that reach almost 80\,\% for IG-SG, only some percent below the accuracy of the classifier on the full inputs. This allows us to conclude that the Mask Information $\mi{C}{\mask}$ is almost as high as our Evaluation Outcome $\mi{C}{\xlp}$.

To show that the Mitigator is almost zero which leads to class information leakage, we test the \textit{Invertible Imputation} condition. Therefore, the inverse function $\scatterOP_{l,M}^{-1}$ that predicts the imputation mask from the imputed image is required (having this function, finding $\scatterOP_{l,x}^{-1}$ is trivial). For the fixed value imputation, an approximate inverse is simple: Setting all pixels in the mask to $0$ if the corresponding image pixel has the filling value (which has to be inferred from the distribution). For a stronger verification, we train an imputation predictor network consisting of three convolutional layers, which predicts for each pixel if it was imputed or original. As \Cref{fig:impupredictionacc} (blue curve) shows, the miss-classification rate when using fixed value imputation is almost zero, i.e., the network can easily recognize the pixels that were imputed. According to our analysis, in this setting close to \textit{Invertible Imputation}, the Mitigator will be negligibly small.

This leads us to the conclusion that the mask-related leakage fundamentally influences many previous evaluations using fixed value imputation \citep{shrikumar2017learning,petsiuk2018rise, Hooker2019ROAR} and it is essential to stop the information leakage through the masks.

\subsection{Debiasing with Noisy Linear Imputation}
\label{sec:linear imputation}
To reduce the Class Information Leakage, we propose a better-suited imputation operator $\scatterOP_l$ that adheres to the \textit{Minimally Revealing Imputation} condition we derived. The remaining process is left unchanged and stays as depicted in \cref{fig:functionalsetup}. However, we face three requirements: (1) We have to get closer to the theoretical condition of Minimally Revealing Imputation. (2) The imputation strategy needs to be highly efficient, since the imputation module has to be run for each image in the data set. (3) We wish to have as few hyper-parameters as possible (preferably none to rule out another confounding factor). 

\begin{figure}[h]
\begin{subfigure}[h]{0.15\textwidth}
    \centering
    \includegraphics[width=\textwidth]{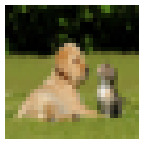}
    \caption{Original}
    \label{fig:inpaint1}
\end{subfigure}
\begin{subfigure}[h]{0.15\textwidth}
    \centering
    \includegraphics[width=\textwidth]{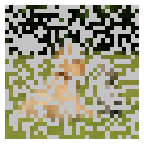}
    \caption{Fixed Imp.}
    \label{fig:inpaint2}
\end{subfigure}
\begin{subfigure}[h]{0.15\textwidth}
    \centering
    \includegraphics[width=\textwidth]{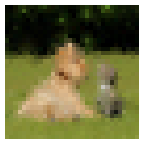}
    \caption{Noisy Lin. Imp.}
    \label{fig:inpaint3}
\end{subfigure}
\begin{subfigure}[b]{0.15\textwidth}
    \centering
\definecolor{color0}{RGB}{55,162,219}
\definecolor{color1}{RGB}{255,71,76}
\definecolor{color2}{RGB}{178,51,120}
\definecolor{color3}{RGB}{232,187,44}
\definecolor{color4}{RGB}{25,161,108}
\scalebox{0.9}{
\begin{tikzpicture}[every node/.style={inner sep=0,outer sep=0}]
    \newcommand{\numnodes}{4} 
    \newcommand{\spacing}{0.3}
    \newcommand{\legendspacing}{0.5}
    \newcommand{\legendoffset}{0.5}
    \newcommand{\legendxoffset}{1.2}
    \foreach \x in {0,...,\numnodes}{
        \draw (0, \spacing*\x) -- (\spacing*\numnodes, \spacing*\x);
        \draw (\spacing*\x, 0) -- (\spacing*\x, \spacing*\numnodes);
    }
    \foreach \x in {0,...,\numnodes}
        \foreach \y in {0,...,\numnodes}
            \filldraw[fill=white, very thick] (\spacing*\x,\spacing*\y) circle (0.07);
    \filldraw[color=blue, fill=blue!15, very thick] (1*\spacing,1*\spacing) circle (0.1);
    \filldraw[color=color4, fill=color4!15, very thick] (1*\spacing,2*\spacing) circle (0.1);
    \filldraw[color=blue, fill=blue!15, very thick] (1*\spacing,3*\spacing)  circle (0.1);
    \filldraw[color=color4, fill=color4!15, very thick] (2*\spacing,1*\spacing) circle (0.1);
    \filldraw[color=color2, fill=color2!15, very thick] (2*\spacing,2*\spacing) circle (0.1);
    \filldraw[color=color4, fill=color4!15, very thick] (2*\spacing,3*\spacing) circle (0.1);
    
    \filldraw[color=blue, fill=blue!15, very thick] (3*\spacing,1*\spacing) circle (0.1);
    \filldraw[color=color4, fill=color4!15, very thick] (3*\spacing,2*\spacing) circle (0.1);
    \filldraw[color=blue, fill=blue!15, very thick] (3*\spacing,3*\spacing) circle (0.1);

    \filldraw[color=color2, fill=color2!15, very thick] (-2*\spacing,-\legendoffset) circle (0.1);
    \node[draw=none, align = left, text width=26mm] at (\legendxoffset,-\legendoffset) {image pixel};
    
    \filldraw[color=color4, fill=color4!15, very thick] (-2*\spacing,-\legendoffset-1*\legendspacing) circle (0.1);
    \node[draw=none, align = left, text width=26mm] at (\legendxoffset,-\legendoffset-1*\legendspacing) {direct neighbor};
    
    \filldraw[color=blue, fill=blue!15, very thick]
    (-2*\spacing,-\legendoffset-2*\legendspacing) circle (0.1);
    \node[draw=none, align = left, text width=26mm] at (\legendxoffset,-\legendoffset-2*\legendspacing) {indirect neighbor};
    
\end{tikzpicture}
}
    \caption{Graphical model used to derive our imputation \label{fig:graphstructure}}
\end{subfigure}
\hfill
\begin{subfigure}[b]{0.3\textwidth}
    \centering
    \includegraphics[width=\textwidth]{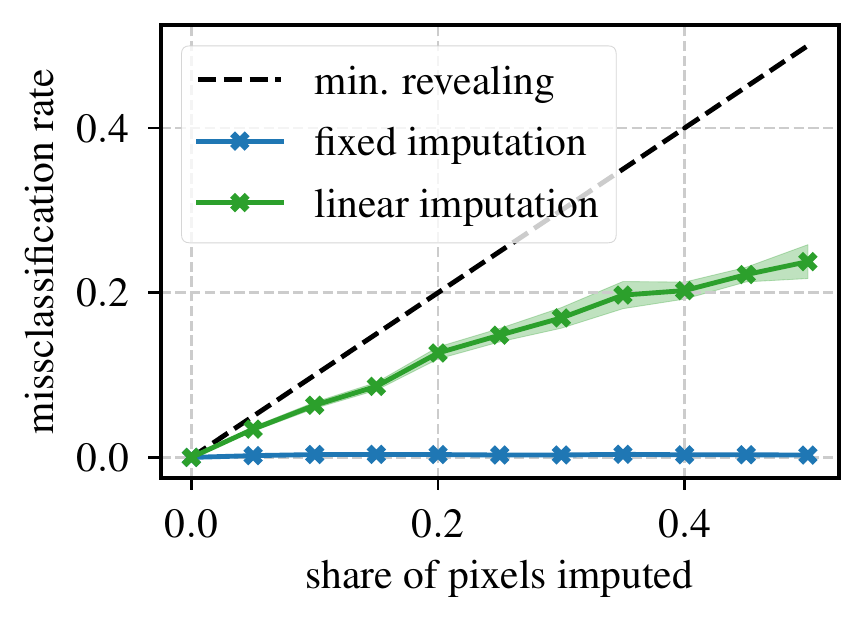}
    \caption{Misclassification rate of the imputation predictor for different shares of randomly imputed pixels (CIFAR-10)}
    \label{fig:impupredictionacc}
\end{subfigure}

\caption{The considered imputation operators. When 50\,\% of the original image (a) are removed, they can either be imputed by a fixed value (b) or by our proposed Noisy Linear strategy (c,d). Training of an imputation predictor (e) shows that it is much harder to tell which pixels are original and which were imputed when using our proposed imputation model. \added{This is closer to the optimal, minimally revealing imputation (black).} Hence, by using imputed samples of this kind, Class Information Leakage is reduced.\label{fig:inpainingexample}}
\end{figure}

We devise a new strategy called \textit{Noisy Linear Imputation}, which fulfills the above goals. In this way, our model addresses some of the fundamental problems of existing strategies. Intuitively, we search a way to make more subtle imputations that cannot be easily recognized and result in lower $\mi{\xlp}{\mask}$. To this end, we suppose that each pixel can be approximated by the weighted mean of its neighbors (cf. \cref{fig:graphstructure}) as image pixels are highly correlated\footnote{In fact, for direct and indirect neighbors, $\rho{=}0.89$ and $\rho{=}0.82$ respectively on CIFAR-10}:
\begin{equation*}
\begin{split}
    \bm{x}_{i,j}&=w_d\left(\bm{x}_{i,j+1}+\bm{x}_{i,j-1}+\bm{x}_{i+1,j}+\bm{x}_{i-1,j}\right)\\
    &+w_i\left(\bm{x}_{i+1,j+1}+\bm{x}_{i-1,j+1}+\bm{x}_{i+1,j-1}+\bm{x}_{i-1,j-1} \right)
\end{split}
\end{equation*}
where $w_d, w_i$ are constant coefficients for direct neighbors and indirect, diagonal neighbors. When setting up a single equation for each removed pixel we arrive at an equation system. \added{For known pixels, we directly plug in their values and only consider each removed pixel as an unknown variable. When neighboring pixels are removed, the equations become connected and cannot be solved independently.} Nevertheless, the resulting system is sparse and can be efficiently solved, even for a large number of missing pixels. To choose the neighbor weights for the linear interpolation, we draw inspiration from the graph structure \added{(see \cref{fig:graphstructure})}: Indirect neighbors have distance 2 from the original node in the graph and direct neighbors have distance 1. Hence, we gave the direct neighbors twice the weight of the diagonal ones. Because the weights need to some up to 1 for a weighted interpolation, this leads to $w_d{=}\frac{1}{6}$ and $w_i{=}\frac{1}{12}$. We add a small random noise ($\sigma=0.1$) to the solution to ensure that the linear dependency cannot be learned by the model.

\Cref{fig:inpainingexample} (top) provides an example of an imputed sample. From the imputed version in \cref{fig:inpaint3}, inference on the mask is significantly harder than the one imputed with fixed values as in \cref{fig:inpaint2}. We again train the imputation predictor for verification and show the results in \cref{fig:impupredictionacc}. We confirm that our strategy lies significantly closer to the optimal, Minimally Revealing Imputation. 
\added{Admittedly, there are even more sophisticated imputation strategies, for example building on Generative Adversarial Networks (GANs) such as Generative Adversarial Imputation Nets (GAIN) proposed by \citet{yoon2018gain}. However, our strategy already achieves considerable improvements and is highly efficient, because it does not require training of a GAN model. For completeness, we include additional experiments with GAN imputation in \cref{sec: gain}.}


\section{Experiments}
\label{sec:road}
Having established that our Noisy Linear Imputation fulfills its purpose, in this section, \added{we show that it entails even more benefits in practice}. We first highlight how it makes results among different evaluation strategies more consistent in \cref{sec:consistency}. We then present another considerable advantage in \cref{sec:efficiency}: its agreement with a no-retraining evaluation strategy is sufficiently high, so that the retraining step is no longer required. 
We name this debiased and no-retraining evaluation framework ROAD (RemOve And Debias). 
All experiments in this section were conducted on CIFAR-10 using the eight attribution methods mentioned. \added{We also use Food-101 \cite{bossard2014food}, a large-scale dataset of high-resolution images, to validate the generalizability of our method. To this end, we train over 1000 models from scratch on data imputed using the strategies, explanations and removal percentages. Since the results on Food-101 also support the findings from CIFAR-10, we include them in \cref{sec: results food}.}

\subsection{Consistency under Removal Orders}\label{sec:consistency}

\begin{figure}[t]
\begin{subfigure}[h]{0.48\linewidth}
    \centering
    \includegraphics[width=\linewidth, trim=0cm 0.3cm 0cm 0cm, clip ]{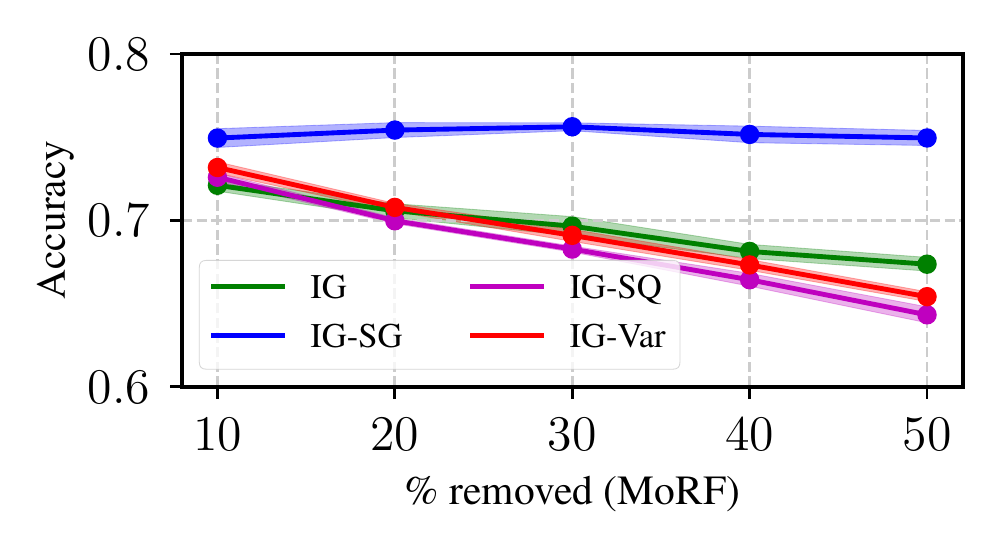}
    \caption{MoRF: Fixed Imp.}
    \label{fig: ig_morf_ret_fixed_main}
\end{subfigure}
\begin{subfigure}[h]{0.48\linewidth}
    \centering
    \includegraphics[width=\linewidth, trim=0cm 0.3cm 0cm 0cm, clip]{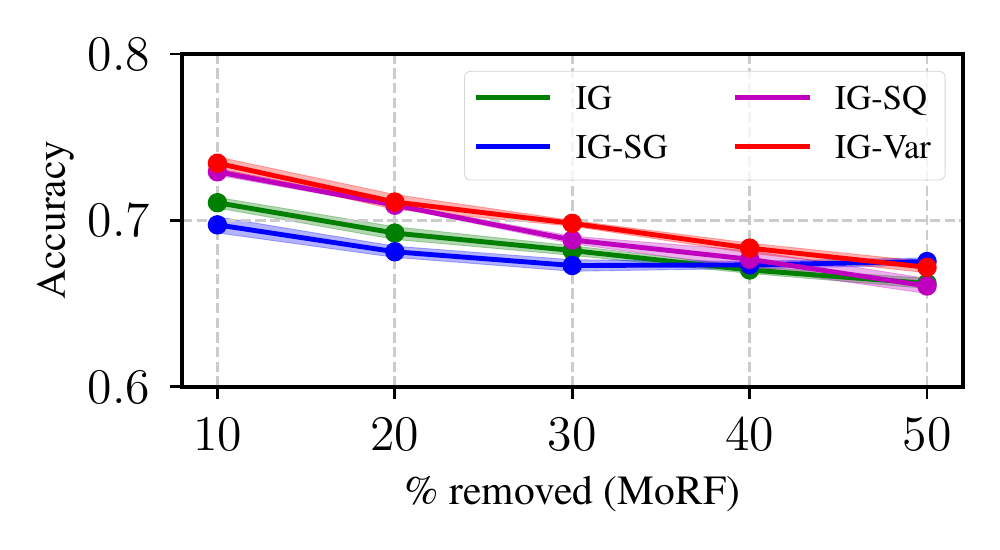}
    \caption{MoRF: Noisy Lin. Imp.}
    \label{fig: ig_morf_ret_linear_main}
\end{subfigure}
\\
\begin{subfigure}[h]{0.48\linewidth}
    \centering
    \includegraphics[width=\linewidth, trim=0cm 0.3cm 0cm 0cm, clip]{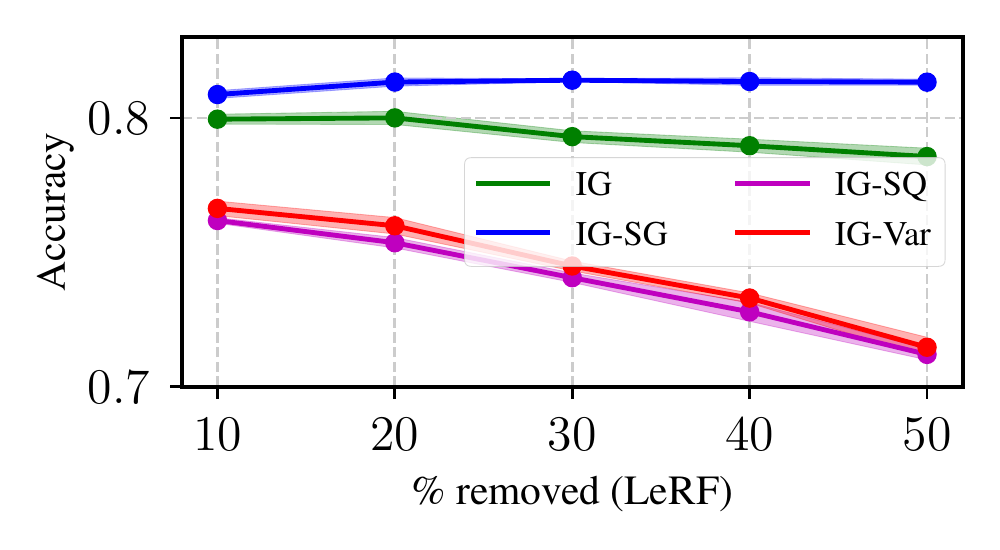}
    \caption{LeRF: Fixed Imp.}
    \label{fig: ig_lerf_ret_fixed_main}
\end{subfigure}
\begin{subfigure}[h]{0.48\linewidth}
    \centering
    \includegraphics[width=\linewidth, trim=0cm 0.3cm 0cm 0cm, clip]{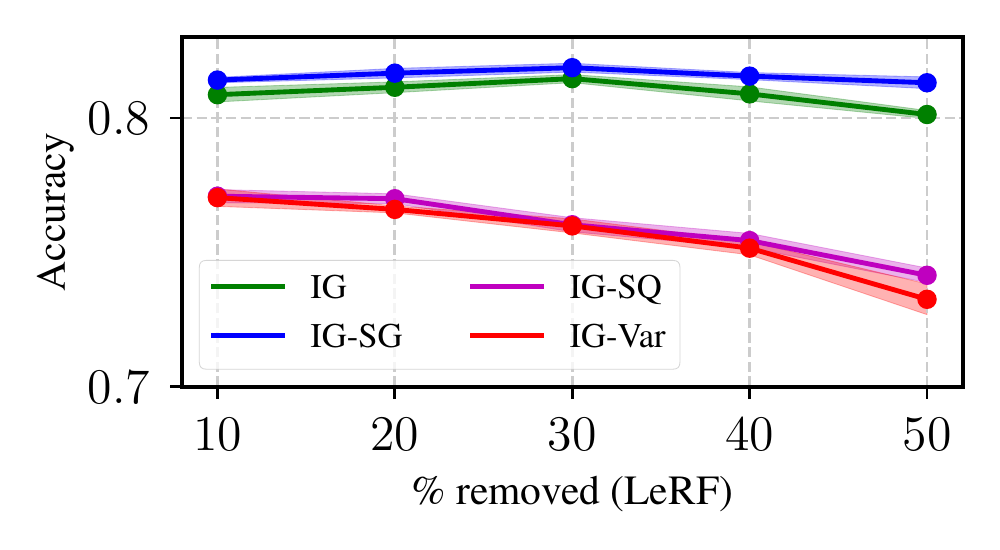}
\caption{LeRF: Noisy Lin. Imp.}
\label{fig: ig_lerf_ret_linear_main}
\end{subfigure}
\caption{Consistency comparison using fixed value vs. Noisy Linear Imputation. The higher accuracy is better in LeRF, while the lower is better in MoRF. Comparing (a) and (c), fixed value imputation gives different rankings in MoRF and LeRF orders: IG-SG is the best in LeRF but the worst in MoRF. Comparing (b) and (d), Noisy Linear Imputation changes the outcome considerably and yields a consistent ranking in MoRF and LeRF. \label{fig:consistency}}
\vspace{-0.3cm}
\end{figure}

As we aim for evaluation strategies that are less prone to the hyperparameter setting and allow for a consistent ranking, we study the consistency of evaluation results under the different removal orders MoRF and LeRF. \cref{fig:consistency} depicts the obtained curves \added{(using ``Retrain")}. For a clear view, we only show four curves of attribution methods based on IG with retraining and up to 50\% pixels are removed. We include the full curves for the IG with its derivatives as well as GB with derivatives in \added{\cref{sec: cifar appendix}.}
The results using the common fixed value imputation shown in \cref{fig: ig_morf_ret_fixed_main} and \cref{fig: ig_lerf_ret_fixed_main}. The results with our Noisy Linear Imputation are shown in \cref{fig: ig_morf_ret_linear_main} and \cref{fig: ig_lerf_ret_linear_main}. In MoRF, a sharp drop in the beginning indicates a better attribution method, while a slight drop is desirable in LeRF. Hence, using fixed imputation, the ranking in MoRF is IG, IG-Var, IG-SQ, IG-SG, whereas the ranking in LeRF is IG-SG, IG, IG-SQ, and IG-Var. We see, for instance, that IG-SG is the worst in MoRF and the best in LeRF. When using the Noisy Linear Imputation, the inconsistency vanishes. The ranking in MoRF is: IG-SG, IG, IG-SQ, and IG-Var, which is the same as in LeRF. 

We quantitatively compute the consistency among all eight attribution methods with and without retraining. Concretely, we compute the ranks (from 1=best to 8=worst) of our explanation methods for each percentage of perturbed pixels. We then calculate the Spearman Rank correlation between different evaluation strategies. 
As shown in \cref{tab:consistency}, the correlation score of the fixed value imputation is \added{$-0.01$} when using retraining and $0.01$ when no retraining is applied. This indicates no consistency in the rankings. When we deploy our Noisy Linear Imputation, the results change drastically: The correlation score is improved to \added{$0.61$} and \added{$0.58$} with and without retraining, respectively. This might imply that the information leakage is responsible for a major share of the inconsistency.
\newcommand{\wstd}[2]{#1\small{$\pm$#2}}
\begin{table}[t]
\vspace{0.5cm}
\centering
\scalebox{0.9}{
\begin{tabular}{*{4}{>{\centering\arraybackslash}m{1.6cm}}}
    \toprule
    \multicolumn{2}{c}{Retrain} &\multicolumn{2}{c}{No-Retrain}
    \\
    \multicolumn{2}{c}{MoRF vs. LeRF} &  \multicolumn{2}{c}{MoRF vs. LeRF}\\
    \cmidrule(lr){1-2} \cmidrule(lr){3-4}
    fixed & lin & fixed & lin \\
    \cmidrule(lr){1-2} \cmidrule(lr){3-4}
    \wstd{-0.01}{0.01} & \wstd{\textbf{0.61}}{0.01}  
    & \wstd{0.01}{0.00} & \wstd{\textbf{0.58}}{0.01}\\
    \bottomrule
\end{tabular}}
\caption{Spearman rank correlation between evaluation strategies. There is almost no agreement between MoRF and LeRF when using fixed imputation (as in previous works). When using our imputation (``lin``), consistency across MoRF and LeRF orders increases drastically.}
\label{tab:consistency}
\vspace{-0.3cm}
\end{table}

\subsection{Efficiency}
\label{sec:efficiency}
\newcommand{\hide}[1]{}

\begin{table}[]
\hide{\begin{subtable}{0.5\textwidth}
\centering
\scalebox{0.9}{
\begin{tabular}{|c|c|c|c|}
\cline{3-4}
\multicolumn{2}{c|}{\multirow{3}{*}{}} & \multicolumn{2}{c|}{Retrain}\\
\multicolumn{2}{c|}{}  & \multicolumn{2}{c|}{MoRF}   \\ \cline{3-4} \multicolumn{2}{c|}{} & fix   & lin   
\\ \hline
\multirow{2}{*}{\begin{tabular}[c]{c@{}c@{}} No-Retrain\\ MoRF\end{tabular}} & fix &  0.14 & 0.39   
\\ \cline{2-4} 
&lin &0.65     &\textbf{0.87}  
\\ \hline
\end{tabular}}
\caption{Evaluation strategy MoRF}
\end{subtable}
\begin{subtable}{0.5\textwidth}
\vspace{0.5cm}
\centering
\scalebox{0.9}{
\begin{tabular}{*{4}{>{\centering\arraybackslash}m{1.4cm}}}
\cline{3-4}
\multicolumn{2}{c|} &\multicolumn{2}{c|}{Retrain}\\
\multicolumn{2}{c|}{}  & \multicolumn{2}{c|}{LeRF}  \\ \cline{3-4} \multicolumn{2}{c|}{} & fix    & lin   
\\ \hline
\multirow{2}{*}{\begin{tabular}[c]{c@{}c@{}} No-Retrain\\ LeRF\end{tabular}} & fix &  0.05 & 0.13   
\\ \cline{2-4} 
&lin &0.82     &\textbf{0.95}  
\\ \hline
\end{tabular}}

\caption{Evaluation strategy LeRF}
\end{subtable}}

\centering
    \scalebox{0.9}{\begin{tabular}{*{4}{>{\centering\arraybackslash}m{1.6cm}}}
    \toprule
    \multicolumn{2}{c}{MoRF} &\multicolumn{2}{c}{LeRF}
    \\
    \multicolumn{2}{c}{Retain vs. No-Retr.} &  \multicolumn{2}{c}{Retain vs. No-Retr.}\\
    \cmidrule(lr){1-2}\cmidrule(lr){3-4}
    fixed & lin 
    & fixed & lin \\
    \cmidrule(lr){1-2} \cmidrule(lr){3-4}
    \wstd{0.15}{0.01} & \wstd{\textbf{0.84}}{0.01}
    & \wstd{0.09}{0.01} & \wstd{\textbf{0.94}}{0.01} \\
    \bottomrule
\end{tabular}}

\caption{Spearman rank correlation between evaluation with and without retraining. Our Noisy Linear Imputation (``lin") also results only in marginal differences between ``Retrain" and ``No-Retrain". We conclude that the retraining step is no longer necessary.}
\label{tab:spearman}
\end{table}
When we apply our Noisy Linear Imputation, we additionally reduce the difference between evaluation with and without retraining. This can be attributed to the reduced distribution shift incurred when using an almost \textit{Minimally Revealing Imputation}. If all pixels were perfectly imputed, the resulting image would not be out-of-distribution.
Since we are interested in the rankings of attribution methods, we again compute Spearman correlation between the rankings obtained with and without retraining and show it in \cref{tab:spearman}. 
The order remains almost always intact between the ``Retrain" with Noisy Linear Imputation and the ``No-Retrain" variant with Noisy Linear Imputation resulting in a rank correlation of \added{$0.84$} in using MoRF and \added{$0.94$} in LeRF. This leads us to the conclusion that ``No-Retrain" and ``Retrain" end up with a highly similar ranking when using Noisy Linear Imputation. Thus, we conclude that the retraining step is not longer justified and can be skipped without significant distortion of the results. \added{Qualitative results are shown in \cref{supp-sec: cifar figures}, cf. \cref{suppfig:cifar-gb-linear}  (CIFAR-10) and \cref{suppfig:food-gb-linear}} (Food-101).

These results allow us to introduce a novel evaluation framework. We refer to the removal with Noisy Linear Imputation and no retraining as ROAD -- Remove and Debias. We showed that ROAD is highly consistent with the compensated results of the ROAR, but comes at an enormous advantage: The retraining step is no longer required. This permits to save a vast amount of computation time. In our experiments, evaluation using the ROAD took only 0.7\,\% of  the resources required for ROAR, as given by the runtimes in \cref{tab:runtimes} obtained on the same hardware (single Nvidia GTX 2080Ti and 8 Cores).

\renewcommand{\wstd}[2]{\makecell{#1\small\\$\pm$#2}}
\begin{table}[h]
    \centering
    \scalebox{0.9}{\begin{tabular}{r cccc}
        \toprule
    \multirow{2}{*}{Strategy}& \multicolumn{2}{c}{Retrain}&\multicolumn{2}{c}{No-Retrain}\\
    \cmidrule(lr){2-3}\cmidrule(lr){4-5}
    & fixed$^\dagger$ & lin  & fixed & lin$^\star$\\
    \midrule
        Time & 3903$\pm$117\,s & 4686$\pm$2\,s & 18.0$\pm$0.1\,s & 33.3$\pm$0.1\,s\\
        Relative & 100\,\% &  120\,\% & 0.5\,\% & 0.9\,\%\\
    \bottomrule
    \end{tabular}}
    \caption{Mean runtime ($5$ runs) for evaluating a single explanation method (IG). $^\dagger$ refers to ROAR, and $\star$ to our ROAD.}
    \label{tab:runtimes}
    \vspace{-0.3cm}
\end{table}

\begin{figure}[th]
\begin{subfigure}[h]{0.48\linewidth}
    \centering
    \includegraphics[width=\linewidth]{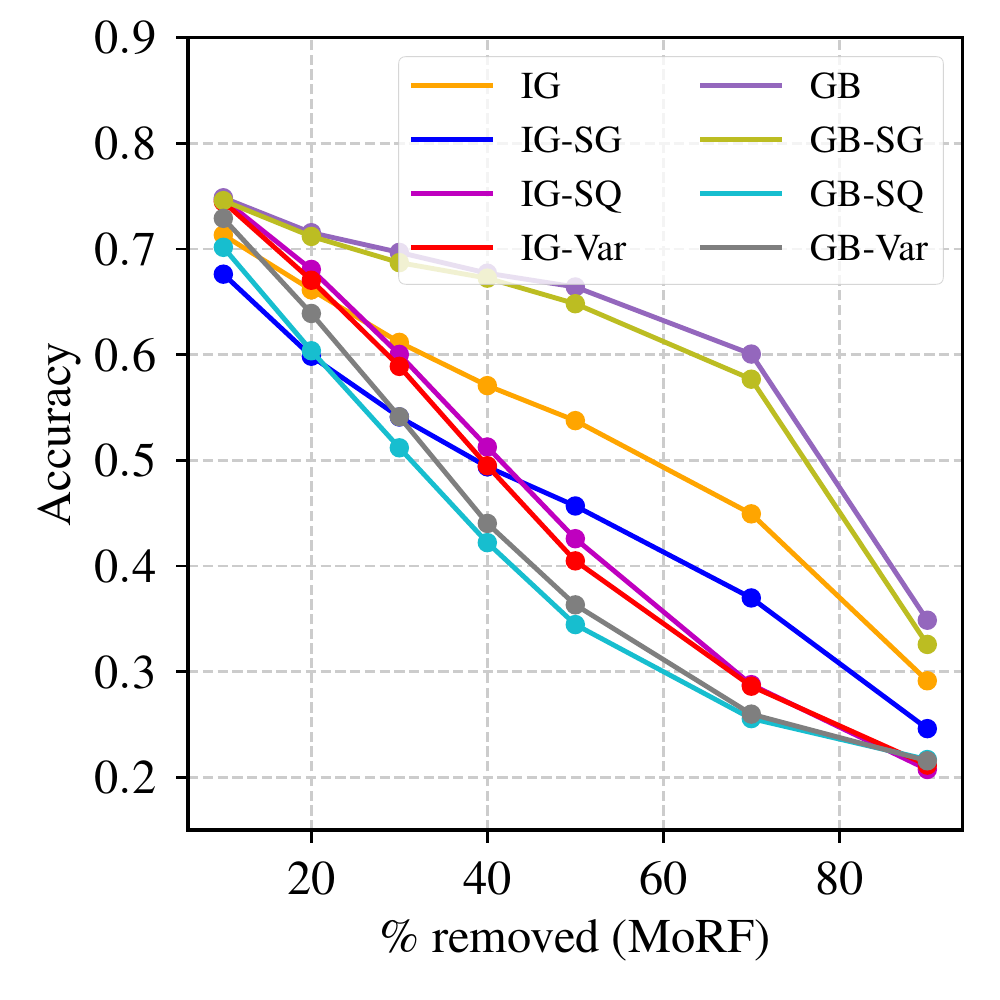}
    \caption{MoRF}
\end{subfigure}
\begin{subfigure}[h]{0.48\linewidth}
    \centering
    \includegraphics[width=\linewidth]{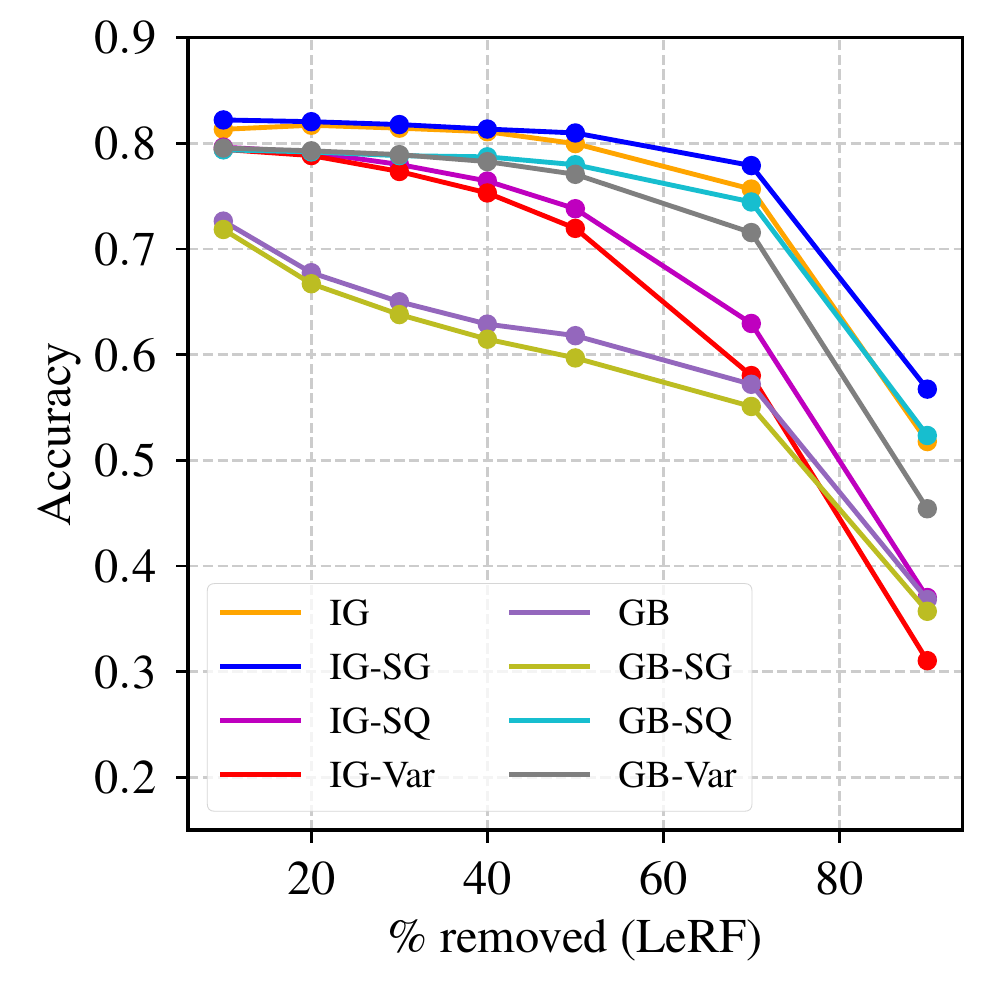}
    \caption{LeRF}
\end{subfigure}
\caption{Evaluation results in MoRF (a) and LeRF (b) using our ROAD framework. \label{fig:8 comparison}}
\vspace{-0.3cm}
\end{figure}

In the end, we illustrate the evaluation results using ROAD among all eight attribution methods in MoRF and LeRF in \cref{fig:8 comparison}. In MoRF, the best ones are IG-SG, GB-SQ, GB-Var and IG, which have lower accuracies in the beginning, whereas they have higher accuracies in LeRF. GB and GB-Var both perform badly in MoRF and LeRF. We see that some inconsistencies still remain, which cannot be compensated by the current imputation. However, the evaluation strategies might also consider different characteristics of an attribution method (e.g., one might be particularly good at identifying irrelevant pixels), which is why perfect agreement might not even be desirable.

\section{Conclusion and Outlook}
We introduced ROAD, an evaluation approach for measuring global fidelity among attribution explanations. ROAD comes with two key advantages over existing methods: (1) it is highly efficient, e.g., permitting a 99\% runtime reduction w.r.t. ROAR, and (2) it circumvents the Class Information Leakage issue, which was thoroughly analyzed in this work. We believe the ROAD framework will be beneficial to the research community because it unifies several methods and is more consistent under varying removal orders. Moreover, it is broadly accessible due to its low resource requirements. ROAD is open-source\footnote{\added{An official implementation is also included in the Quantus framework \cite{hedstrom2022quantus}}}, and can be readily implemented in practical use-cases. Going forward, we plan to investigate more sophisticated imputation models in ROAD as well as other evaluation metrics besides fidelity.

\subsubsection*{Acknowledgements}
We acknowledge the support by the Cluster of Excellence - Machine Learning: New Perspectives for Science, EXC number 2064/1 - Project number 390727645, and the support of the Training Center for Machine Learning (TCML) Tübingen, funded by the German Federal Ministry of Education and Research (BMBF) with grant number 01IS17054, which provided substantial resources for running our large-scale Food-101 experiment.

\typeout{}

\bibliography{main}
\bibliographystyle{icml2022}

\newpage
\appendix
\onecolumn
\section{Additional Theory}
\subsection{Formulation of the MI Bounds for the Binary Case}
\label{sec:formulationbounds}
As we discussed in our main paper, the relationship between Mutual Information (MI) and accuracy is not a function, but comes in form of upper and lower bounds of the obtainable accuracy. If, for example, the binary classification case with equal class priors $p(C=0)=p(C=1)=\frac{1}{2}$ is considered, the following bounds can be derived \citep{Hellman1970perror, Meyen2016masterThesis}:
\begin{equation}
\label{eqn:defmi}
\frac{I(\bx;C) +1}{2} \leq \acc(C|\bx) \leq H_2^{-1}(1-I(\bx;C)),    
\end{equation}
where $H_2^{-1}: [0,1] \rightarrow \left[\frac{1}{2}, 1\right]$ is the inverse of the binary entropy with support $\left[\frac{1}{2}, 1\right]$. For completeness, we restate the proof of this upper bound in \Cref{sec:miandaccproofs}.

\subsection{Reproduction of the proof of the relation between mutual and accuracy in the binary case}
\label{sec:miandaccproofs}
In this section, we reproduce the proofs for the upper and lower bounds of bayesian classifier accuracy given a certain amount of mutual information from the master's thesis by \cite{Meyen2016masterThesis} for completeness. The upper bound given there is tighter than the bounds present in the literature. \\
We consider the following setting ($C$, $\bx$ are random variables):
\begin{itemize}
    \item binary classification problem, $C\in \Omega_C =\{0,1\}$
    \item equal class priors $P(C=0)=\frac{1}{2}, P(C=1)=\frac{1}{2}$
    \item discrete features $\bx$ (which can be the product of multiple random variables)
    \item support set $\Omega_{x}=\supp{\left\{\bx\right\}}$ of \textit{countable} size
\end{itemize}  
We first prove the following Lemma:
\begin{lemma}
Let the assumptions stated above be true. Then, the mutual information is the weighted mean of a function of the conditional accuracies $\acc(C|s)$, where $s\in \Omega_{x}$:
\[
I\left(C; \bx\right) = \sum_{s\in \Omega_{S}} p(s)\left(1-H_2\left[\acc(C|s)\right]\right)
\]\label{lem:micondacc}
\end{lemma}
In this formulation, $p(s)$ is a shorthand for $P(\bx=s)$ and $H_2(p) := -p\log p - (1-p)\log(1-p)$ is the entropy for a binary random variable.\\
\textbf{Proof.}\\
\begin{align}
    I\left(C; \bx\right) &= H(C)-H(C|\bx)\\
    &= \sum_{c\in \Omega_C} p(c) \log \frac{1}{p(c)} - \sum_{s\in \Omega_S} p(s) \sum_{c\in \Omega_C} p(c|s) \log \frac{1}{p(c|s)}\\
    &= \sum_{s\in \Omega_x} p(s) \left[\sum_{c\in \Omega_C} p(c) \log \frac{1}{p(c)} -  \sum_{c\in \Omega_C} p(c|s) \log \frac{1}{p(c|s)}\right]\\
    &= \sum_{s\in \Omega_x} p(s) \left[H(C)-H(C|s)\right]
\end{align}
In our consideration, $\Omega_C =\{0,1\}$ and $P(C=0)=\frac{1}{2}, P(C=1)=\frac{1}{2}$, so $H(C)=1$. Additionally, the bayesian classifier rule yields
\begin{equation}
    acc(C|s)= \left\{\begin{array}{lr}
        P(C=0|s), & \text{for } P(C=1|s) \leq 0.5\\
        P(C=1|s), & \text{for } P(C=1|s) > 0.5
        \end{array}\right. 
\end{equation}
and
\begin{align}
H(C|s) &= -P(C=0|s)\log P(C=0|s) - P(C=1|s) \log P(C=1|s)\\
&= H_2(P(C=0|s)) = H_2(P(C=1|s))\\
&= H_2(acc(C|s)) 
\end{align}
Plugging in the results $H(C)=1$ and $H(C|s)=H_2(\acc(C|s))$, we obtain the proposed lemma.
~$\hfill\square$

For the derivation of upper and lower bounds, Jenssen's inequality is used. $1-H_2(\cdot)$ is a convex function and the $\left\{p(s)\right\}_{s\in\Omega_x}$ are convex multipliers, i.e., they are non-negative and sum up to one. Then,
\begin{align}
1-H_2\left(\acc(C|\bx) \right) & = 1-H_2\left(\sum_{s\in \Omega_{x}} p(s)\acc(C|s) \right)\\
& \leq \sum_{s\in \Omega_{x}} p(s)\left[1-H_2\left(\acc(C|s)\right)\right] = I(\bx;C)
\end{align}
We can restate this equation in terms of accuracy.
\begin{equation}
   H_2\left(\acc(C|\bx) \right) \geq 1-I(C;\bx) 
\end{equation}
Using that $H_2\left(\cdot\right)$ is decreasing monotonically on the interval $\left[\frac{1}{2},1\right]$, so its inverse $H^{-1}_2$ exists, and that $\acc(C|s) \geq 0.5$:
\begin{equation}
    \acc(C|\bx) \leq H_2^{-1}\left(1- I(C; \bx) \right).   
\end{equation}
The inequality sign is flipped again, due to the inverse being monotonically decreasing. Note that the bounds derived for the special case are much tighter than the general ones provided by \citet{Vergara2014reviewMIfeatureSelection} and \citet[Chapter 2.10]{Cover2006}, that are not of any use, because they are even less strict than the trivial bound $\acc(C|\bx) \leq 1$, for the simple case considered here.

For the lower bound, we refer the reader to \citet[eqn. 18]{Hellman1970perror}, where the term $I$ corresponds to $H(C|\bx)=H(C)-I(C;\bx)$ in our notation.
Rewriting the result from \citet{Hellman1970perror} in our notation, we obtain
\begin{equation}
    1-\acc(C|\bx) \leq \frac{H(C)-I(C;\bx)}{2}.  
\end{equation}

Using $H(C)=1$ and rearranging yields
\begin{equation}
    1-\acc(C|\bx) \leq \frac{1-I(C;\bx)}{2}
\end{equation}
and
\begin{equation}
    \acc(C|\bx) \geq \frac{I(C;\bx)+1}{2}.
\end{equation}
~$\hfill\square$

\subsection{Analysis of the LeRF Ordering}
\begin{figure}[t]
    \centering
    \begin{tikzpicture}[
    squarednode/.style={rectangle, draw=black,  thick, minimum size=4mm, text height=.6em,text width=.6em, text centered,text depth=.5ex},
    ]
    
    \newcommand{\cellfc}[0]{darkgray}
    \newcommand{\belowdist}[0]{0.00cm}
    \definecolor{color0}{RGB}{55,162,219}
    \definecolor{color1}{RGB}{255,71,76}
    \definecolor{color2}{RGB}{178,51,120}
    \definecolor{color3}{RGB}{232,187,44}
    \definecolor{color4}{RGB}{25,161,108}
    \definecolor{color5}{RGB}{40,97,205}
    \definecolor{color6}{RGB}{240,158,7}
    \definecolor{color7}{RGB}{180,43,40}
    \definecolor{color8}{RGB}{156,158,114}
    
    \matrix (m1) at (0,0) [matrix of nodes,nodes={squarednode},column sep=-\pgflinewidth, row sep=-\pgflinewidth]{
        |[draw,fill=color0]| a & |[draw,fill=color1]| b & |[draw,fill=color2]| c \\
        |[draw,fill=color3]| d & |[draw,fill=color4]| e & |[draw,fill=color5]| g \\
        |[draw,fill=color6]| g & |[draw,fill=color7]| h & |[draw,fill=color8]| i \\
        };
    \node [above =\belowdist of m1]{$\bm{x}$ (input)};    
    \node (explan)[below =1.5 cm of m1]{\parbox[c]{1.9cm}{\raggedright explainer $\bm{e}$ for model $f$}}; 
    
    \matrix (m2) [below right = 0.3cm and 0.5cm of m1, matrix of nodes,  nodes={squarednode}, column sep=-\pgflinewidth, row sep=-\pgflinewidth]{
    |[draw,fill=black]| \textcolor{white}{a} & |[draw,fill=black]| \textcolor{white}{b}  & |[draw]| c \\
    |[draw]| d & |[draw,fill=black]| \textcolor{white}{e}  & |[draw,fill=black]| \textcolor{white}{g}  \\
    |[draw]| g & |[draw,fill=black]| \textcolor{white}{h}  & |[draw]| i \\
    };
    \node [below =\belowdist of m2]{$\mask$};
    
    \matrix (mxl) [right= 2.4cm of m1, matrix of nodes,  nodes={squarednode}, column sep=-\pgflinewidth, row sep=-\pgflinewidth]{
    |[draw,fill=color0]| a\\
    |[draw,fill=color1]| b\\
    |[draw,fill=color4]| e\\
    |[draw,fill=color5]| g\\
    |[draw,fill=color7]| h\\
    };
    \node [below = \belowdist of mxl]{$\xh$};
    
    \matrix (mxlp) [right = 1.5cm of m2, matrix of nodes,  nodes={squarednode}, column sep=-\pgflinewidth, row sep=-\pgflinewidth]{
        |[draw,fill=color0]| a & |[draw,fill=color1]| b & |[draw,fill=\cellfc]| \textcolor{\cellfc}{2} \\
        |[draw,fill=\cellfc]| \textcolor{\cellfc}{3} & 
        |[draw,fill=color4]| e & |[draw,fill=color5]| g \\
        |[draw,fill=\cellfc]| \textcolor{\cellfc}{6} & |[draw,fill=color7]| h & |[draw,fill=\cellfc]| \textcolor{\cellfc}{8}  \\
    };
    \node [below =\belowdist of mxlp]{$\xhp$ (high importance)};
    
    \node at (4.2, -1.5) {\textcolor{gray}{$\scatterOP_h$}};
    \node at (2.6, -0.3) {\textcolor{gray}{$\mathcal{M}_h$}};
    
    \draw[->, dotted](explan.north) -- (m2.west);
    \draw[->](m1) -- (m2);
    \draw[->](m1) -- (mxl);
    \draw[->](m2) -- (mxl);
    \draw[->](m2) -- (mxlp);
    \draw[->](mxl) -- (mxlp);
    \end{tikzpicture}
    \caption{Analogous analytical model of feature removal in the opposite order (LeRF): The input image $\bm{x}$ is explained by an explanation method that returns a mask $\mask$ indicating important pixels. The remaining, highly important pixels can be extracted from the image using the masking operator $\mathcal{M}_h$ and transformed to a modified variant of the input $\xhp$ via the imputation operator $\scatterOP_h$.\label{fig:functionalsetuplerf}}
\end{figure}
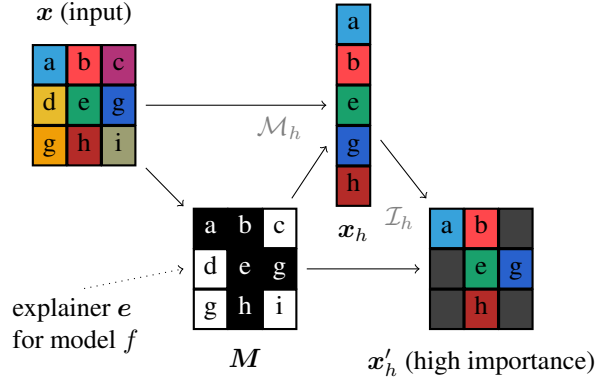

In this section, we analyze the masking impact for the case of the Least Relevant First (LeRF) ordering. We first provide a definition for the operators involved as we did for the Most Relevant First (MoRF) case. In the LeRF setting, the $k$ least important important features per instance are removed. We model the explanation as a choice of features via a binary mask $\mask = \bm{e}\left(f, \bm{x}\right) \in \left\{ 0,1 \right\}^d$, with the corresponding value set to one, if the corresponding feature is among the top-$k$, and to zero otherwise. Furthermore, suppose $\mathcal{M}_{h}: \left\{ 0,1 \right\}^d \times  \mathbb{R}^d \rightarrow \mathbb{R}^{k}$ to be the selection operator for the \textbf{\underline{h}}ighly important dimensions indicated in the mask and $\bm{x}_{h}=\mathcal{M}_{h}\left(\bm{M}, \bm{x}\right)$ to be a vector containing only the remaining, highly important features as shown in \cref{fig:functionalsetuplerf}. We suppose that the features preserve their internal order in $\xh$, i.e., features are ordered ascendingly by their original input indices. 

The LeRF approach with retraining (also called ``Keep and Retrain'', KAR, by \citet{Hooker2019ROAR}) measures the accuracy of a newly trained classifier $f'$ on modified samples $\xhp \coloneqq \scatterOP_{h}\left(\bm{M},\bm{x_h}\right)$, where $\scatterOP_{h}: \left\{ 0,1 \right\}^d\times\mathbb{R}^{k} \rightarrow \mathbb{R}^{d}$ is an imputation operator that redistributes all inputs in the vector $\xh$ to their original positions and sets the remainder to some filling value. This means only the top-$k$ features are kept. For a better evaluation result, the accuracy should increase quickly with increasing $k$, indicating the most influential features are present. Accuracy should not increase much for the high values of $k$, because inserting the low importance features should not have a large effect (equivalently, this means it should not drop much when the least important features are removed). Overall, higher accuracies indicate better attributions in the LeRF setting.

For the LeRF benchmark, the quantity of interest in our analysis will be $I(\xhp; C)$, the class information contained in the filled-in version of the selected high important features. We want to maximize $I(\xhp; C)$ to obtain a good score,
\[
\uparrow I(\xhp; C)~~\Rightarrow~~\uparrow \text{LeRF benchmark}.
\]

As before, we can apply the following, general identity:
\begin{equation}
\label{eqn:maskgeneral2}
   \underbrace{I(\xhp; C)}_\text{Evaluation Outcome} = \underbrace{I(C;\xhp|\mask)}_\text{Feature Info.} +  \underbrace{I(C;\mask)}_\text{Mask Info.} - \underbrace{I(C;\mask|\xhp)}_\text{Mitigator}.
\end{equation}
The interpretation of the terms is analogous to that in our main paper.
\paragraph{Class-Leaking Explanation Map} For the case of the class-leaking map, we again require the imputation operator to be invertible:
\begin{example}
\label{ass:condinvertibleinfilling}
\textit{Invertible Imputation. Let $\scatterOP_h: \left\{ 0,1 \right\}^d \times  \mathbb{R}^{k} \rightarrow \mathbb{R}^{d}$ be the imputation operator that takes the highly important features as an input. We suppose that there are inverse functions $\scatterOP_{h,M}^{-1}$ and $\scatterOP_{h,x}^{-1}$, such that}
\[
\xhp = \scatterOP_h\left(\bm{M}, \xh\right) \Leftrightarrow \mask = \scatterOP_{h,M}^{-1}(\xhp) \wedge \bm{x}_h = \scatterOP_{h,x}^{-1}(\bm{x}_h^\prime).
\]
\end{example}

If, for instance, the pixels removed are set to some reserved value indicating their absence, the infilling operator is invertible. In this case, also the Mitigator $\mic{C}{\mask}{\xhp} = 0$ (see \Cref{sec:biasreduction} for details).
The ``Feature Info" term is constrained to be positive. Thus, the Mask Information has a non-negligible impact on the Evaluation Goal, because a higher Mask term will always increase it.

We can create a another example of a spurious explanation map that shows how evaluation scores are influenced even worse for LeRF: Suppose an explanation map that starts masking out pixels at the top for class zero and at the bottom for class one. Thus, a retrained model will be able to infer the category just from the shape of the masked pixels and obtain the best possible accuracy and thus score in the LeRF setting. However, it does not provide a reasonable attribution for the importance of the features.

\section{GAN Imputation}
\label{sec: gain}
We also use Generative Adversarial Imputation Nets (GAIN) proposed by \citet{yoon2018gain} as an imputation operator. We first train a GAIN model on CIFAR-10. To find the best-performing setup,
we run a hyperparameter selection for the GAIN model. We keep all the default parameters identified by \citet{kachuee2020generative}, but search for the value of \texttt{alpha} ($\alpha$), which can be seen as a weight factor for the reconstruction loss of the non-imputed pixels in the GAN, and the \texttt{hint\_rate} ($hr$) parameter, which provides the Discriminator with hints to balance the difficulty of the tasks. We train the models for 100 epochs which resulted in converged MSEs and Frechet Inception Distances (FIDs). We use MSE to the original pixels to assess the generative quality of the model. \citet{kachuee2020generative} reported low values for both these parameters to perform well, but did not provide the exact values. We extended their value ranges to $\alpha=100$ and performed and exhaustive search. The results for the GAIN models on CIFAR-10 can be seen in \cref{tab:gainhyper}. For the experiments we used the best setup with $\alpha=100$ and $hr=0.01$.
\begin{table}[h]
\centering
\begin{tabular}{r|c|c|c|c}
    & $\alpha$=0.1 & $\alpha$=1  &  $\alpha$=10 & $\alpha$=100 \\
    \hline
    $hr$=0.01 & 0.0131 & 0.0164 & 0.0090 & \textbf{0.0085} \\
    $hr$=0.1 & 0.0113 & 0.0133 & 0.0131 & 0.0101\\
    $hr$=0.3 & 0.0172 & 0.0183 & 0.0151 & 0.0127\\
    $hr$=0.9 & 0.0303 & 0.0484 &0.0379 &0.0088\\
\end{tabular}
\caption{Mean-Squared-Errors for GAIN on CIFAR-10 using different hyperparameter choices. \label{tab:gainhyper}}
\end{table}

\begin{figure}[h]
  \centering
  \begin{minipage}{.23\textwidth}
    \centering
    \begin{subfigure}[h]{.45\textwidth}
        \includegraphics[width=1\textwidth]{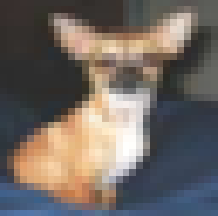}
        \caption{}
        \label{supp-fig:inpaint1}
        \end{subfigure}
    \hfill
    \begin{subfigure}[h]{.45\textwidth}
        \centering
        \includegraphics[width=\textwidth]{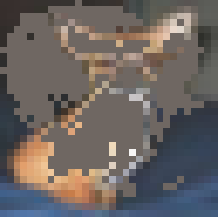}
        \caption{}
        \label{supp-fig:inpaint2}
    \end{subfigure}
    \\
    \begin{subfigure}[h]{.45\textwidth}
        \centering
        \includegraphics[width=\textwidth]{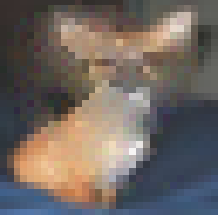}
        \caption{}
        \label{supp-fig:inpaint3}
    \end{subfigure}
    \hfill 
    \begin{subfigure}[h]{.45\textwidth}
        \centering
        \includegraphics[width=\textwidth]{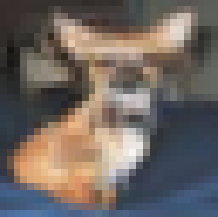}
        \caption{}
        \label{supp-fig:inpaint4}
    \end{subfigure}
  \end{minipage}
  \hspace{15pt}
  \begin{minipage}{.35\textwidth}
    \begin{subfigure}[b]{\textwidth}
        \centering
        \includegraphics[width=\textwidth]{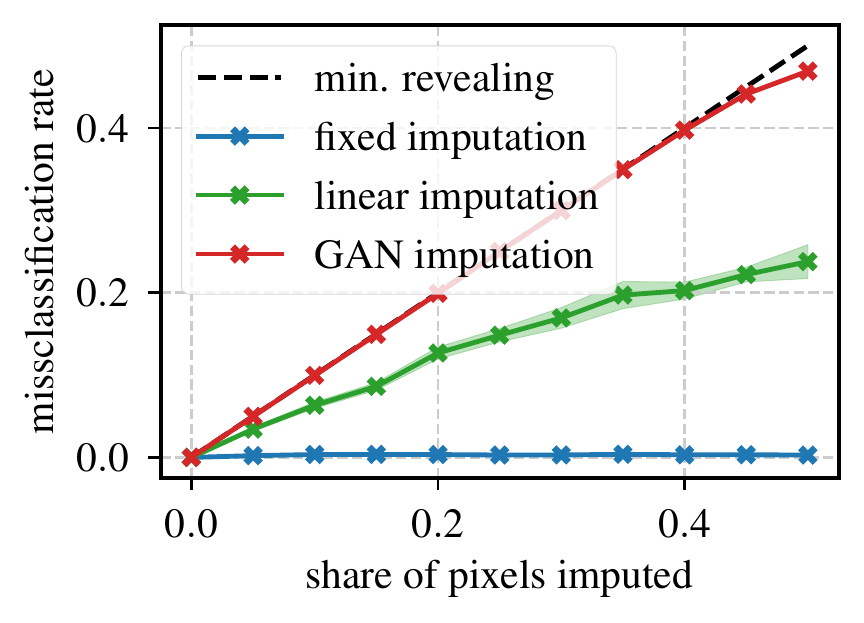}
        \caption{}
        \label{supp-fig:impupredictionacc}
    \end{subfigure}
  \end{minipage}
\caption{The considered imputation operators. When 30\,\% of the original image (a) are removed, they can either be completed by a fixed value (b) or by our proposed Noisy Linear imputation (c) or GAN imputation (d). Training of an imputation predictor (e) shows that it is much harder to tell which pixels are original and which were imputed when using our proposed imputation models, which is closer to the theoretical optimum (black). Hence, Class Information Leakage is reduced by our imputation methods.}
\label{supp-fig: gain imp appendix}
\end{figure}

In \cref{supp-fig: gain imp appendix}, we demonstrate imputation results using three operators for one image (a) from CIFAR-10. Compared to the fixed value imputation (b) and noisy linear imputation (c), GAN imputation (d) yields most natural imputed image. Although it cannot perfectly reconstruct the original image, for example the background is noisy and the body color is different from the original one, it is not easy to deduce the mask from (d). A trained imputation predictor also verifies that GAN imputation is closest to the optimal condition, Minimally Revealing Imputation. 

However, there are drawbacks of the GAN imputation. It may introduce some new ``features" that do not exist in the original sample. For instance the dog in (d) has new patterns on its body. Moreover, it does not give very good results when too many pixels are removed (cf. \cref{subfig:imputation overview}). The GAIN training again requires tuning hyperparameter settings and is highly expensive. Therefore, this model does not allow for the desired improvements (few hyperparameters, efficiency). Compared to GAN, our Noisy Linear imputation does not have these drawbacks. Considering all these factors, we recommend to use Noisy Linear Imputation in the evaluation framework.

\section{Additional Experiments on CIFAR-10}
\label{sec: cifar appendix}
\subsection{Implementation Details}
\label{sec: supp cifar implementation}

\begin{figure}[t]
\centering
   \includegraphics[width=1\linewidth]{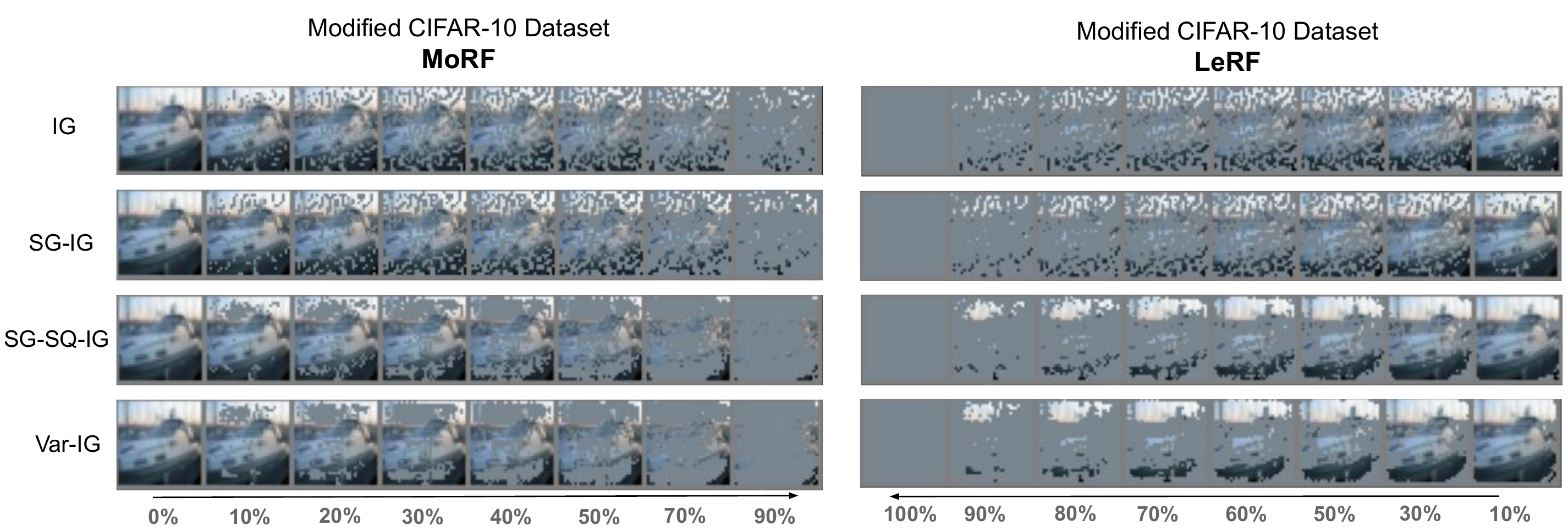}
    \caption{Illustration of modified data set in MoRF/LeRF and fixed value imputation settings. \textbf{Left}: Modifications in the MoRF framework. \textbf{Right}: Modifications in the LeRF framework. \textbf{Top to Bottom}: Modifications using Integrated Gradient (IG)\ \citep{sundararajan2017axiomatic}  and three ensemble variants of IG: SmoothGrad (SG-IG)\ \citep{smilkov2017smoothgrad}, SmoothGrad$^{2}$ (SG-SQ-IG)\ \citep{Hooker2019ROAR}, and VarGrad (Var-IG) \ \citep{adebayo2018sanity}. The percentage of pixels that are removed or kept is given at the bottom.}
    \label{fig:cifar}
\end{figure}

\begin{figure}[h]
    \begin{subfigure}[h]{0.48\textwidth}
        \centering
        \includegraphics[width=0.9\textwidth]{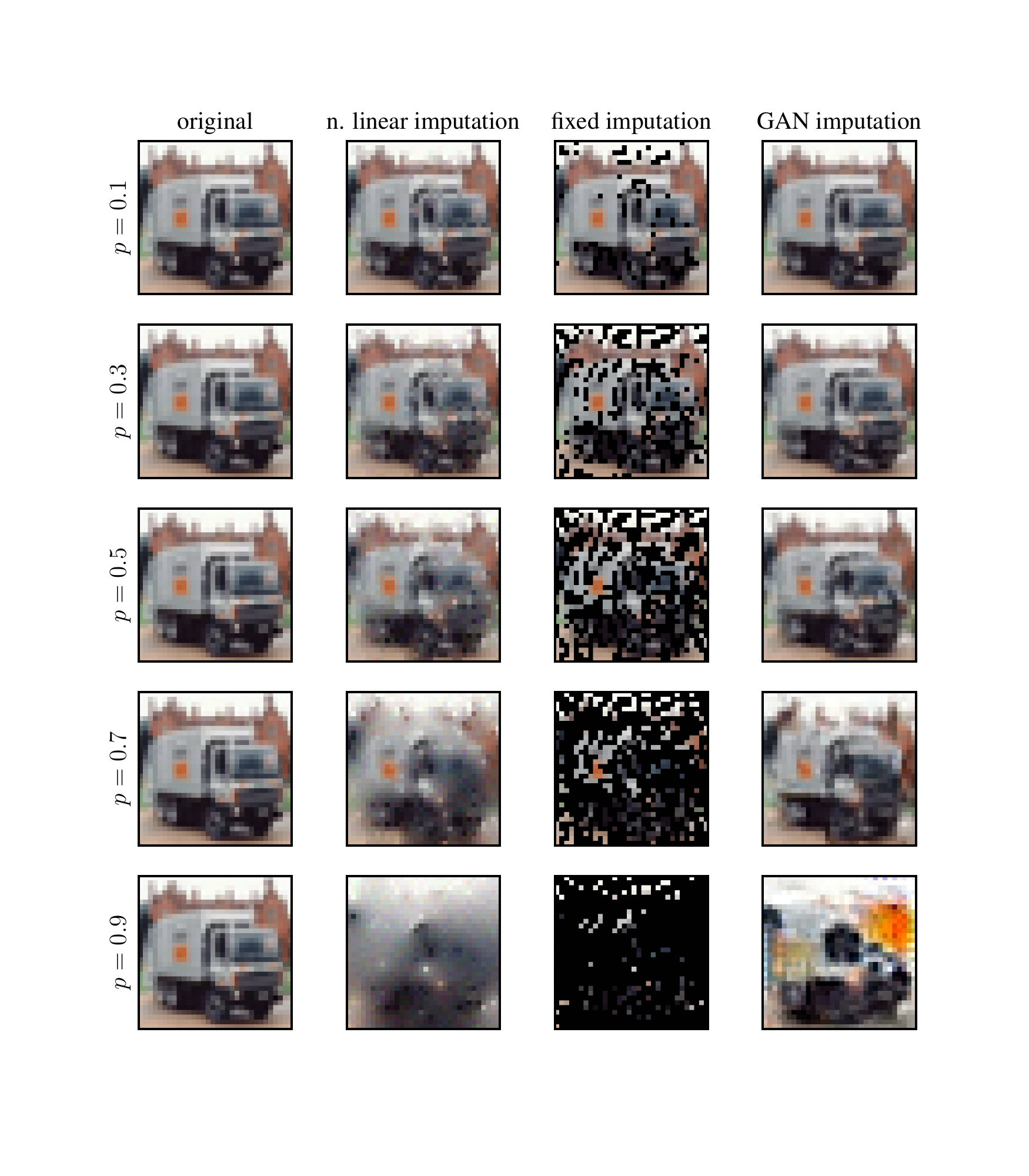}
        \caption{Sample from CIFAR-10}
    \end{subfigure}
    \begin{subfigure}[h]{0.48\textwidth}
        \centering
        \includegraphics[width=0.9\textwidth]{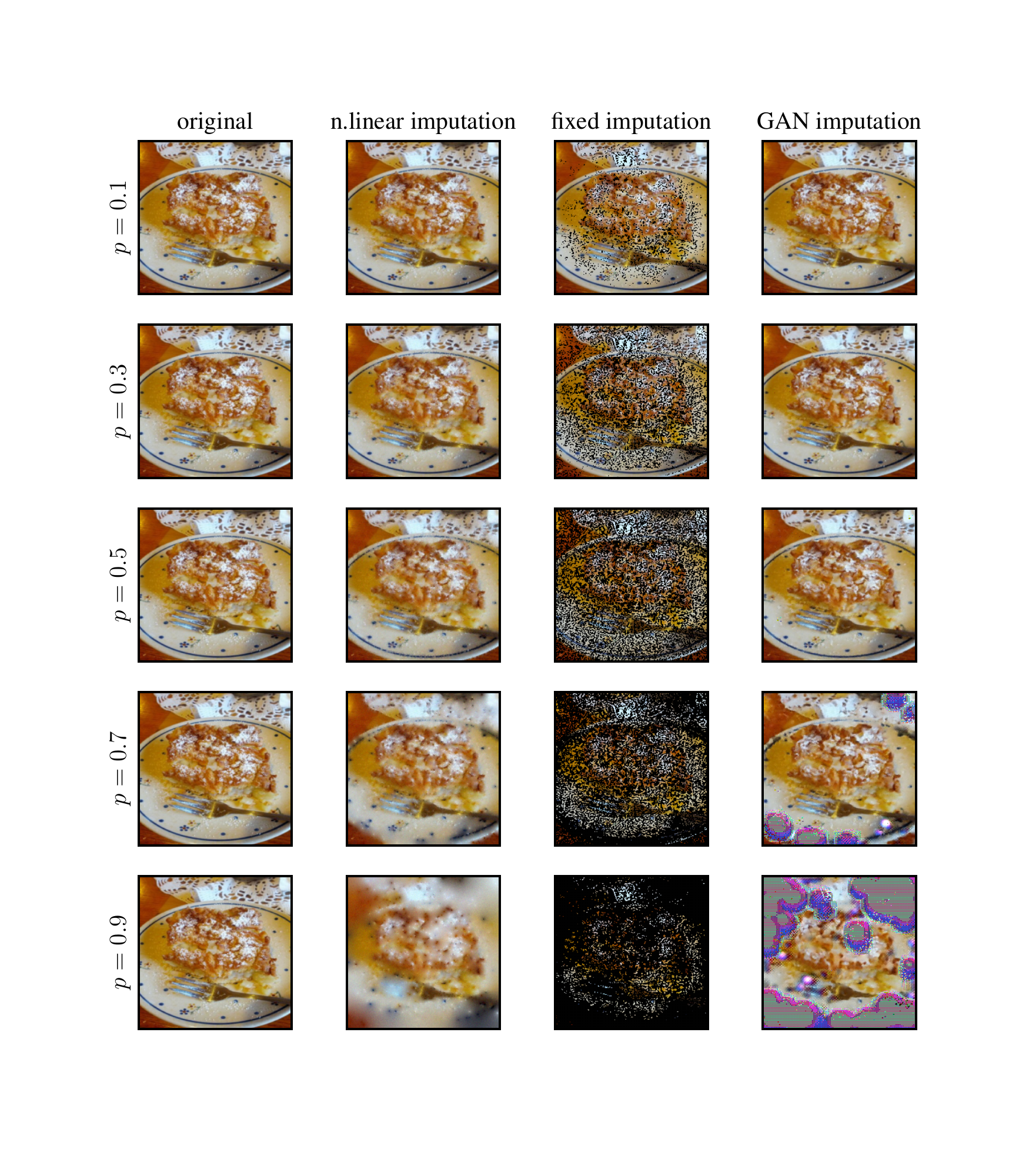}
        \caption{Sample from Food-101}
    \end{subfigure}
     \caption{Sample images from CIFAR-10 and Food-101 imputed with the three methods considered in this work for different percentages. The missing pixels are determined by the IG attribution method (in MoRF order). While the GAN leads to sharper images for the early percentage values, where the linearly imputed samples become more blurry. Artefacts are introduced for high missingness percentages (0.9) in GAN imputation, which may distort the results of the evaluation once again. Therefore, we decide to stick to the Noisy Linear Imputation that operates more stably.}
    \label{subfig:imputation overview}
\end{figure}

In this section, we report implementation details on CIFAR-10 as well as additional results for comparison between fixed value imputation and our \textit{Noisy Linear Imputation}. We also include GAN imputation results. In \cref{subfig:imputation overview}, an overview of using three different imputations with different perturbation percentages are illustrated. 

We train a vanilla ResNet-18 \cite{he2016deep} on CIFAR-10 and compute different explanations using the trained model. The model is trained with the initial learning rate of $0.01$ and the SGD optimizer \cite{sutskever2013importance}. We decrease the learning rate by factor $0.1$ after $25$ and train the model for $40$ epochs on one GPU. The trained model achieves a test set accuracy of 84.5\,\% (comparable to the model in \cite{tomsett2020sanity}). For attributions, we use the same settings as in \citep{Hooker2019ROAR}: As base explanations we implement Integrated Gradient (IG)\ \citep{sundararajan2017axiomatic} and Guided Backprop (GB)\ \citep{springenberg2014striving}. Additionally, we use three ensembling strategies for each: SmoothGrad (SG)\ \citep{smilkov2017smoothgrad}, SmoothGrad$^{2}$ (SG-SQ)\ \citep{Hooker2019ROAR} and VarGrad (Var) \ \citep{adebayo2018sanity}. For each explanation method, we modify the data set using the fraction of pixels $\eta = [0,0.1,0.2,0.3,0.4,0.5,0.7,0.9]$. \Cref{fig:cifar} illustrates the modified images by using four different explanations in the GB-family within MoRF and LeRF orders (fixed mean value imputation is used). 

We use \textbf{$N=5$ runs} and report averaged results for all CIFAR-10 experiments in our paper and indicate the standard errors (which are very small) as an area behind our plots. In \cref{tab:dev ig-sg} and \cref{tab:dev gb-sg}, we show the mean accuracy and its standard deviation at each the fraction of pixels $\eta$ for IG-SG and GB-SG explanations. For other explanations we used, the standard deviation at each $\eta$ in the magnitude of below one percent as well. Mean runtimes (average over 5 runs) for evaluating one explanation method (IG) using all three imputation methods are listed in \cref{subtab:runtimes}. 

\hide{
\begin{table}[h]
\centering
\scalebox{0.8}
{\begin{tabular}{|c|c|c|c|c|c|c|c|c|}
\cline{3-9}
\cline{3-9} \multicolumn{2}{c|}{} & 10  & 20  & 30 & 40 & 50 &  70    & 90   \\ 
\hline
\multirow{3}{*}{\begin{tabular}[c]{c@{}c@{}} Retrain\\ MoRF\end{tabular}} & fixed &  \wstdmy{74.94}{0.57}\ & \wstdmy{75.42}{0.45}   & \wstdmy{75.62}{0.24} & \wstdmy{75.16}{0.50} & \wstdmy{74.95}{0.45} & \wstdmy{73.73}{0.48} & \wstdmy{65.18}{0.85} \\\cline{3-9}
&lin & \wstdmy{69.72}{0.49}     & \wstdmy{68.10}{0.34}  & \wstdmy{67.28}{0.34} & \wstdmy{67.32}{0.22} & \wstdmy{67.52}{0.22}& \wstdmy{66.46}{0.54} & \wstdmy{60.37}{0.51}\\\cline{3-9}
&gan &     &   &  & & &  &  \\\hline

\multirow{3}{*}{\begin{tabular}[c]{c@{}c@{}} No-Retrain\\ MoRF\end{tabular}} & fixed &  \wstdmy{44.06}{0.04} & \wstdmy{29.81}{0.03}  & \wstdmy{21.99}{0.03} & \wstdmy{17.35}{0.02} & \wstdmy{14.67}{0.01}& \wstdmy{11.50}{0.04} & \wstdmy{10.90}{0.03}\\\cline{3-9}
&lin& \wstdmy{67.66}{0.02}     &\wstdmy{59.94}{0.03}  & \wstdmy{54.05}{0.05} & \wstdmy{49.46}{0.04} & \wstdmy{45.63}{0.06} & \wstdmy{36.87}{0.05} & \wstdmy{24.55}{0.04}\\\cline{3-9}
&gan &     &   &  & & &  &  \\\hline

\multirow{3}{*}{\begin{tabular}[c]{c@{}c@{}} Retrain\\ LeRF\end{tabular}} & fixed &  \wstdmy{77.52}{0.26} & \wstdmy{80.14}{0.11}   & \wstdmy{81.08}{0.14} & \wstdmy{81.40}{0.15} & \wstdmy{81.50}{0.07} & \wstdmy{81.85}{0.12}  & \wstdmy{82.01}{0.20}\\\cline{3-9}
&lin & \wstdmy{72.73}{0.20}    & \wstdmy{77.52}{0.23}  &\wstdmy{79.29}{0.41} & \wstdmy{80.57}{0.31} & \wstdmy{80.97}{0.20} & \wstdmy{81.64}{0.26} &  \wstdmy{81.41}{0.13} \\\cline{3-9}
&gan &     &   &  & & &  &  \\\hline

\multirow{3}{*}{\begin{tabular}[c]{c@{}c@{}} No-Retrain\\ LeRF\end{tabular}} & fixed &  \wstdmy{46.76}{0.04} & \wstdmy{51.84}{0.04}   & \wstdmy{53.80}{0.05} & \wstdmy{55.39}{0.06} & \wstdmy{57.59}{0.05} & \wstdmy{64.02}{0.05} & \wstdmy{74.33}{0.03} \\\cline{3-9}
&lin & \wstdmy{56.48}{0.03}     &\wstdmy{72.57}{0.03} & \wstdmy{77.96}{0.02} & \wstdmy{79.54}{0.01} & \wstdmy{81.02}{0.04} & \wstdmy{81.82}{0.02} &  \wstdmy{82.24}{0.01}\\\cline{3-9}
&gan &     &   &  & & &  &  \\\hline
\end{tabular}}
\caption{Mean accuracy at each $\eta$ by using IG-SG in all methods with standard deviations of five individual runs. For LeRF, the accuracy is at (1-$\eta$).}
\label{tab:dev ig-sg}
\end{table}
}

\begin{table}[h]
\centering
\scalebox{0.8}
{\begin{tabular}{|c|c|c|c|c|c|c|c|c|}
\cline{3-9}
\cline{3-9} \multicolumn{2}{c|}{} & 10  & 20  & 30 & 40 & 50 &  70    & 90   \\ 
\hline
\multirow{3}{*}{\begin{tabular}[c]{c@{}c@{}} Retrain\\ MoRF\end{tabular}} & fixed &  \wstdmy{74.94}{0.57}\ & \wstdmy{75.42}{0.45}   & \wstdmy{75.62}{0.24} & \wstdmy{75.16}{0.50} & \wstdmy{74.95}{0.45} & \wstdmy{73.73}{0.48} & \wstdmy{65.18}{0.85} \\\cline{3-9}
&lin & \wstdmy{69.72}{0.49}     & \wstdmy{68.10}{0.34}  & \wstdmy{67.28}{0.34} & \wstdmy{67.32}{0.22} & \wstdmy{67.52}{0.22}& \wstdmy{66.46}{0.54} & \wstdmy{60.37}{0.51}\\\cline{3-9}
&gan &  \wstdmy{74.78}{0.31}  & \wstdmy{73.16}{0.22}  &  \wstdmy{72.02}{0.03} & \wstdmy{71.40}{0.23} & \wstdmy{70.72}{0.30} & \wstdmy{68.44}{0.43} & \wstdmy{59.37}{0.44} \\\hline

\multirow{3}{*}{\begin{tabular}[c]{c@{}c@{}} No-Retrain\\ MoRF\end{tabular}} & fixed &  \wstdmy{44.06}{0.04} & \wstdmy{29.81}{0.03}  & \wstdmy{21.99}{0.03} & \wstdmy{17.35}{0.02} & \wstdmy{14.67}{0.01}& \wstdmy{11.50}{0.04} & \wstdmy{10.90}{0.03}\\\cline{3-9}
&lin& \wstdmy{67.66}{0.02}     &\wstdmy{59.94}{0.03}  & \wstdmy{54.05}{0.05} & \wstdmy{49.46}{0.04} & \wstdmy{45.63}{0.06} & \wstdmy{36.87}{0.05} & \wstdmy{24.55}{0.04}\\\cline{3-9}
&gan & \wstdmy{74.53}{0.04}   & \wstdmy{71.41}{0.04}  & \wstdmy{69.10}{0.06} & \wstdmy{67.55}{0.09} &\wstdmy{66.55}{0.07} & \wstdmy{60.73}{0.12} & \wstdmy{25.46}{0.10} \\\hline

\multirow{3}{*}{\begin{tabular}[c]{c@{}c@{}} Retrain\\ LeRF\end{tabular}} & fixed  & \wstdmy{80.88}{0.14}  & \wstdmy{81.34}{0.15} & \wstdmy{81.41}{0.01} & \wstdmy{81.36}{0.14}  & \wstdmy{81.34}{0.11} & \wstdmy{80.95}{0.01} &  \wstdmy{76.86}{0.34} \\\cline{3-9}
&lin &  \wstdmy{81.41}{0.10} & \wstdmy{81.67}{0.18} & \wstdmy{81.88}{0.16} & \wstdmy{81.56}{0.13} &\wstdmy{81.31}{0.22}  & \wstdmy{79.89}{0.23}  & \wstdmy{72.83}{0.36}  \\\cline{3-9}
&gan &  \wstdmy{81.05}{0.22} & \wstdmy{80.99}{0.15}   & \wstdmy{80.14}{0.16} & \wstdmy{79.25}{0.18}  & \wstdmy{78.24}{0.22} &  \wstdmy{74.92}{0.15} &\wstdmy{68.69}{0.21}  \\\hline

\multirow{3}{*}{\begin{tabular}[c]{c@{}c@{}} No-Retrain\\ LeRF\end{tabular}} & fixed  & \wstdmy{74.34}{0.02} & \wstdmy{69.04}{0.03}  & \wstdmy{64.06}{0.04} & \wstdmy{59.86}{0.03}   & \wstdmy{57.59}{0.03}  & \wstdmy{53.81}{0.06} &  \wstdmy{46.74}{0.02} \\\cline{3-9}
&lin  & \wstdmy{82.20}{0.04}  & \wstdmy{82.04}{0.03} & \wstdmy{81.76}{0.08} & \wstdmy{81.34}{0.06} & \wstdmy{80.97}{0.03}  &\wstdmy{77.89}{0.07} & \wstdmy{56.74}{0.13}     \\\cline{3-9}
&gan   & \wstdmy{80.80}{0.02} &\wstdmy{80.38}{0.03} &\wstdmy{79.90}{0.02} & \wstdmy{78.85}{0.07} & \wstdmy{77.47}{0.08} & \wstdmy{71.14}{0.10}  & \wstdmy{32.96}{0.17}   \\\hline
\end{tabular}}
\caption{Mean accuracy at each $\eta$ by using IG-SG in all methods with standard deviations of five individual runs. For LeRF, the accuracy is at (1-$\eta$).}
\label{tab:dev ig-sg}
\end{table}

\hide{
\begin{table}[h]
\centering
\scalebox{0.8}
{\begin{tabular}{|c|c|c|c|c|c|c|c|c|}
\cline{3-9}
\cline{3-9} \multicolumn{2}{c|}{} & 10  & 20  & 30 & 40 & 50 &  70    & 90   \\ 
\hline
\multirow{3}{*}{\begin{tabular}[c]{c@{}c@{}} Retrain\\ MoRF\end{tabular}} & fixed &  \wstdmy{76.30}{0.43}\ & \wstdmy{75.60}{0.27}   & \wstdmy{74.89}{0.29} & \wstdmy{74.27}{0.29} & \wstdmy{73.37}{0.28} & \wstdmy{72.15}{0.09} & \wstdmy{67.99}{0.24} \\\cline{3-9}
&lin & \wstdmy{72.83}{0.37}     & \wstdmy{71.87}{0.41}  & \wstdmy{71.58}{0.19} & \wstdmy{70.98}{0.15} & \wstdmy{70.47}{0.20}& \wstdmy{67.81}{0.45} & \wstdmy{59.38}{0.46}\\\cline{3-9}
&gan &     &   &  & & &  &  \\\hline

\multirow{3}{*}{\begin{tabular}[c]{c@{}c@{}} No-Retrain\\ MoRF\end{tabular}} & fix &  \wstdmy{73.01}{0.04} & \wstdmy{66.73}{0.04}  & \wstdmy{58.69}{0.09} & \wstdmy{52.47}{0.10} & \wstdmy{48.52}{0.05}& \wstdmy{48.71}{0.04} & \wstdmy{44.39}{0.01}\\\cline{3-9}
&lin & \wstdmy{74.54}{0.03}     &\wstdmy{71.24}{0.04}  & \wstdmy{68.83}{0.02} & \wstdmy{67.21}{0.01} & \wstdmy{64.80}{0.04} & \wstdmy{57.60}{0.08} & \wstdmy{32.98}{0.03}\\\cline{3-9}
&gan &     &   &  & & &  &  \\\hline

\multirow{3}{*}{\begin{tabular}[c]{c@{}c@{}} Retrain\\ LeRF\end{tabular}} & fixed &  \wstdmy{66.97}{0.52} & \wstdmy{70.45}{0.30}   & \wstdmy{71.44}{0.25} & \wstdmy{72.15}{0.15} & \wstdmy{72.72}{0.11} & \wstdmy{73.91}{0.19}  & \wstdmy{75.24}{0.25}\\\cline{3-9}
&lin & \wstdmy{59.88}{0.39}  &\wstdmy{65.42}{0.47} & \wstdmy{67.76}{0.53} & \wstdmy{68.59}{0.19} & \wstdmy{69.42}{0.67} &  \wstdmy{69.90}{0.30} & \wstdmy{72.13}{0.40} \\\cline{3-9}
&gan &     &   &  & & &  &  \\\hline

\multirow{3}{*}{\begin{tabular}[c]{c@{}c@{}} No-Retrain\\ LeRF\end{tabular}} & fixed & \wstdmy{37.12}{0.03}     &\wstdmy{41.63}{0.03} & \wstdmy{42.32}{0.05} & \wstdmy{43.99}{0.07} & \wstdmy{46.96}{0.12} & \wstdmy{57.85}{0.02} &  \wstdmy{69.62}{0.02}\\\cline{3-9}
&lin & \wstdmy{35.87}{0.04}     &\wstdmy{49.21}{0.04} & \wstdmy{55.18}{0.08} & \wstdmy{58.04}{0.03} & \wstdmy{59.75}{0.02} & \wstdmy{63.71}{0.04} &  \wstdmy{71.85}{0.02}\\\cline{3-9}
&gan &     &   &  & & &  &  \\\hline
\end{tabular}}
\caption{Mean accuracy at each $\eta$ by using GB-SG in all methods with standard deviations of five individual runs. For LeRF, the accuracy is at (1-$\eta$).}
\label{tab:dev gb-sg}
\end{table}
}

\begin{table}[h]
\centering
\scalebox{0.8}
{\begin{tabular}{|c|c|c|c|c|c|c|c|c|}
\cline{3-9}
\cline{3-9} \multicolumn{2}{c|}{} & 10  & 20  & 30 & 40 & 50 &  70    & 90   \\ 
\hline
\multirow{3}{*}{\begin{tabular}[c]{c@{}c@{}} Retrain\\ MoRF\end{tabular}} & fixed &  \wstdmy{76.30}{0.43}\ & \wstdmy{75.60}{0.27}   & \wstdmy{74.89}{0.29} & \wstdmy{74.27}{0.29} & \wstdmy{73.37}{0.28} & \wstdmy{72.15}{0.09} & \wstdmy{67.99}{0.24} \\\cline{3-9}
&lin & \wstdmy{72.83}{0.37}     & \wstdmy{71.87}{0.41}  & \wstdmy{71.58}{0.19} & \wstdmy{70.98}{0.15} & \wstdmy{70.47}{0.20}& \wstdmy{67.81}{0.45} & \wstdmy{59.38}{0.46}\\\cline{3-9}
&gan &  \wstdmy{76.64}{0.13} & \wstdmy{75.44}{0.13}  & \wstdmy{74.73}{0.28}  & \wstdmy{73.69}{0.30} & \wstdmy{72.85}{0.34} & \wstdmy{68.97}{0.08}  & \wstdmy{56.81}{0.30} \\\hline

\multirow{3}{*}{\begin{tabular}[c]{c@{}c@{}} No-Retrain\\ MoRF\end{tabular}} & fix &  \wstdmy{73.03}{0.03} & \wstdmy{66.72}{0.03}  & \wstdmy{58.72}{0.07} & \wstdmy{52.51}{0.04} & \wstdmy{48.52}{0.08}& \wstdmy{48.79}{0.06} & \wstdmy{44.43}{0.06}\\\cline{3-9}
&lin & \wstdmy{74.57}{0.08}     &\wstdmy{71.18}{0.06}  & \wstdmy{68.70}{0.08} & \wstdmy{67.24}{0.08} & \wstdmy{64.82}{0.11} & \wstdmy{57.68}{0.06} & \wstdmy{32.59}{0.09}\\\cline{3-9}
&gan & \wstdmy{76.57}{0.03} & \wstdmy{74.70}{0.04} & \wstdmy{72.51}{0.09} & \wstdmy{71.19}{0.07} & \wstdmy{69.64}{0.08} & \wstdmy{60.89}{0.15} &\wstdmy{21.11}{0.16}  \\\hline

\multirow{3}{*}{\begin{tabular}[c]{c@{}c@{}} Retrain\\ LeRF\end{tabular}} & fixed  & \wstdmy{72.39}{0.39}  & \wstdmy{71.76}{0.41} & \wstdmy{71.21}{0.30} & \wstdmy{70.26}{0.50} & \wstdmy{69.83}{0.22}  & \wstdmy{68.32}{0.45} &  \wstdmy{63.29}{0.56} \\\cline{3-9}
&lin & \wstdmy{72.86}{0.24} &  \wstdmy{71.63}{0.27}  & \wstdmy{70.67}{0.42} & \wstdmy{70.08}{0.30} & \wstdmy{69.82}{0.22}  &\wstdmy{68.10}{0.18} & \wstdmy{60.12}{0.34}\\\cline{3-9}
&gan & \wstdmy{75.97}{0.27}   & \wstdmy{74.73}{0.27} & \wstdmy{73.41}{0.24} & \wstdmy{72.74}{0.34}  & \wstdmy{72.20}{0.28}  & \wstdmy{69.89}{0.26}  &  \wstdmy{57.57}{0.24}\\\hline

\multirow{3}{*}{\begin{tabular}[c]{c@{}c@{}} No-Retrain\\ LeRF\end{tabular}} & fixed &  \wstdmy{69.61}{0.04} & \wstdmy{64.90}{0.02}  & \wstdmy{57.88}{0.05} & \wstdmy{51.67}{0.09} & \wstdmy{46.93}{0.06} &\wstdmy{42.40}{0.09} & \wstdmy{37.10}{0.03}  \\\cline{3-9}
&lin &  \wstdmy{71.84}{0.06} & \wstdmy{66.71}{0.08}  & \wstdmy{63.79}{0.05}  & \wstdmy{61.46}{0.09} & \wstdmy{59.69}{0.09}  &\wstdmy{55.09}{0.06}  & \wstdmy{35.72}{0.13}  \\\cline{3-9}
&gan& \wstdmy{75.13}{0.02}  & \wstdmy{72.13}{0.05} & \wstdmy{70.25}{0.05} & \wstdmy{68.56}{0.08} & \wstdmy{67.35}{0.08} & \wstdmy{62.32}{0.13}  & \wstdmy{24.61}{0.19}  \\\hline
\end{tabular}}
\caption{Mean accuracy at each $\eta$ by using GB-SG in all methods with standard deviations of five individual runs. For LeRF, the accuracy is at (1-$\eta$).}
\label{tab:dev gb-sg}
\end{table}

\renewcommand{\wstd}[2]{\makecell{#1\small\\$\pm$#2}}
\begin{table}[h]
    \centering
    \scalebox{0.9}{\begin{tabular}{r cccccc}
        \toprule
    \multirow{2}{*}{Strategy}& \multicolumn{3}{c}{Retrain}&\multicolumn{3}{c}{No-Retrain}\\
    \cmidrule(lr){2-4}\cmidrule(lr){5-7}
    & fixed$^\dagger$ & lin & gan  & fixed & lin$^\star$ & gan\\
    \midrule
        Time & 3903$\pm$117\,s & 4686$\pm$2\,s & 6421$\pm$74\,s& 18.0$\pm$0.1\,s & 33.3$\pm$0.1\,s & 35.0$\pm$0.1\,s\\
        Relative & 100\,\% &  120\,\% & 164\,\% & 0.5\,\% & 0.9\,\% & 0.9\,\%\\
    \bottomrule
    \end{tabular}}
    \caption{Mean runtime ($5$ runs) for evaluating a single explanation method (IG) on three imputation operators. $^\dagger$ refers to ROAR, and $\star$ to our ROAD.}
    \label{subtab:runtimes}
    \vspace{-0.3cm}
\end{table}

\subsection{Correlation Analysis}
In \cref{suptab:spearman}, we show a full view of the Spearman Correlation of rankings between all twelve different evaluation strategies (``Retrain"/``No-Retrain", MoRF/LeRF, and fixed value/Noisy Linear/GAN imputation) used in this paper. In this work, our primary focus was on consistency between the respective Retraining/No-Retraining Methods and the consistency between MoRF/LeRF and we mark the results used in the main paper in bold.
\renewcommand{\wstd}[2]{\makecell{#1\small\\$\pm$#2}}
\begin{table}[h]
\centering
\scalebox{0.8}{\begin{tabular}{|c|c|ccc|ccc|ccc|ccc|}
\cline{3-14}
\multicolumn{2}{c|}{\multirow{3}{*}{}} & \multicolumn{3}{c|}{Retrain} & \multicolumn{3}{c|}{No-Retrain} & \multicolumn{3}{c|}{Retrain} & \multicolumn{3}{c|}{No-Retrain}\\
\multicolumn{2}{c|}{}  & \multicolumn{3}{c|}{MoRF} & \multicolumn{3}{c|}{MoRF} & \multicolumn{3}{c|}{LeRF} & \multicolumn{3}{c|}{LeRF}\\ 

\cline{3-14} \multicolumn{2}{c|}{} & fixed$^\dagger$ & lin & gan  & fixed & lin$^{*}$ & gan &  fixed  & lin & gan  & fixed & lin & gan\\ 
\hline
\multirow{3}{*}{\begin{tabular}[c]{c@{}c@{}c@{}} Retrain\\ MoRF\end{tabular}} & fixed$^\dagger$ &  \wstd{1.00}{0.00} & & &&& &&& &&&\\
&lin & \wstd{0.68}{0.02} & \wstd{1.00}{0.00} & &&& &&& &&&\\
&gan &  \wstd{0.76}{0.01}  & \wstd{0.82}{0.01}& \wstd{1.00}{0.00} &&& &&& &&& \\\hline
\multirow{3}{*}{\begin{tabular}[c]{c@{}c@{}c@{}} No-Retrain\\ MoRF\end{tabular}} & fixed &  \wstd{\textbf{0.15}}{0.01} &  \wstd{0.38}{0.02} & \wstd{0.23}{0.01} & \wstd{1.00}{0.00} && &&& &&&\\
& lin$^{*}$  & \wstd{{0.66}}{0.01} & \wstd{\textbf{0.84}}{0.01}  & \wstd{0.86}{0.01}  & \wstd{0.43}{0.01} & \wstd{1.00}{0.00}& &&& &&& \\
&gan & \wstd{0.65}{0.01} & \wstd{0.62}{0.01} & \wstd{0.84}{0.01} & \wstd{0.14}{0.01}& \wstd{0.78}{0.01} & \wstd{1.00}{0.00} &&& &&&\\\hline
\multirow{3}{*}{\begin{tabular}[c]{c@{}c@{}c@{}} Retrain\\ LeRF\end{tabular}} & fixed &  \wstd{\textbf{-0.01}}{0.01}& \wstd{0.48}{0.02}  & \wstd{0.28}{0.02}  & \wstd{0.66}{0.00} & \wstd{0.47}{0.02} & \wstd{0.13}{0.01} & \wstd{1.00}{0.00} && &&&  \\
&lin & \wstd{0.16}{0.01} & \wstd{\textbf{0.61}}{0.01}  & \wstd{0.34}{0.01}  & \wstd{0.78}{0.01}  & \wstd{0.50}{0.01} &\wstd{0.10}{0.01} & \wstd{0.87}{0.01} & \wstd{1.00}{0.01} & &&& \\
&gan &  \wstd{0.15}{0.01}  & \wstd{0.59}{0.01} & \wstd{0.32}{0.01} & \wstd{0.74}{0.00} & \wstd{0.50}{0.01} & \wstd{0.10}{0.01} & \wstd{0.90}{0.01} & \wstd{0.96}{0.01} & \wstd{1.00}{0.00} &&& \\\hline
\multirow{3}{*}{\begin{tabular}[c]{c@{}c@{}c@{}} No-Retrain\\ LeRF\end{tabular}} & fixed  & \wstd{0.49}{0.01}  &  \wstd{0.44}{0.01}  & \wstd{{0.69}}{0.01} & \wstd{\textbf{0.01}}{0.00} & \wstd{0.60}{0.00} & \wstd{{0.77}}{0.00} & \wstd{\textbf{0.09}}{0.01} & \wstd{0.03}{0.01} & \wstd{-0.03}{0.00} & \wstd{1.00}{0.00} &&\\
& lin &\wstd{0.21}{0.01}  &  \wstd{0.60}{0.01}& \wstd{0.38}{0.01} & \wstd{{0.81}}{0.00} & \wstd{\textbf{0.58}}{0.01} & \wstd{0.22}{0.01} & \wstd{0.85}{0.00} & \wstd{\textbf{0.94}}{0.01} & \wstd{0.91}{0.00} &\wstd{0.10}{0.00} & \wstd{1.00}{0.00} &\\ 
&gan & \wstd{0.05}{0.01} & \wstd{0.47}{0.01} & \wstd{0.17}{0.01} & \wstd{0.69}{0.00} & \wstd{0.36}{0.00} & \wstd{-0.07}{0.01} & \wstd{0.85}{0.00} & \wstd{0.86}{0.01} & \wstd{0.90}{0.01} & \wstd{-0.14}{0.00} & \wstd{0.79}{0.00} & \wstd{1.00}{0.00}\\\hline
\end{tabular}}
\caption{\textbf{CIFAR-10}: Rank Correlations between all evaluation strategies used with standard deviations computed by considering the rankings obtained through five consecutive runs as independent. Results indicated in bold correspond to those reported in the main paper. The ROAR benchmark is marked by $^\dagger$ and our ROAD by $^{*}$.
\label{suptab:spearman}}
\end{table}

\subsection{Extended Figures}
\label{supp-sec: cifar figures}
In this section, we include full qualitative results of using four variants in evaluation strategies (``Retrain"/``No-Retrain", MoRF/LeRF) for three different imputation operators (fixed value/Noisy Linear/GAN imputation). In \cref{suppfig:cifar-ig-fixed}, the full plots of IG-family attribution methods using fixed value imputation are shown, while \cref{suppfig:cifar-gb-fixed} illustrates for the GB-based attribution methods. \cref{suppfig:cifar-ig-linear} and \cref{suppfig:cifar-gb-linear} show the evaluation results when using our Noisy Linear Imputation for IG- and GB-family attribution methods, respectively. From results, we see that using our Noisy Linear Imputation, the consistency between the evaluation rankings conducted in MoRF and LeRF with and without retraining increases, for instance in \cref{suppfig:cifar-ig-linear} compared to \cref{suppfig:cifar-ig-fixed}.

\newcommand{\figuresuppw}[0]{0.8\textwidth}
\begin{figure}[h]
\begin{subfigure}[h]{0.48\textwidth}
    \centering
    \includegraphics[width=\figuresuppw]{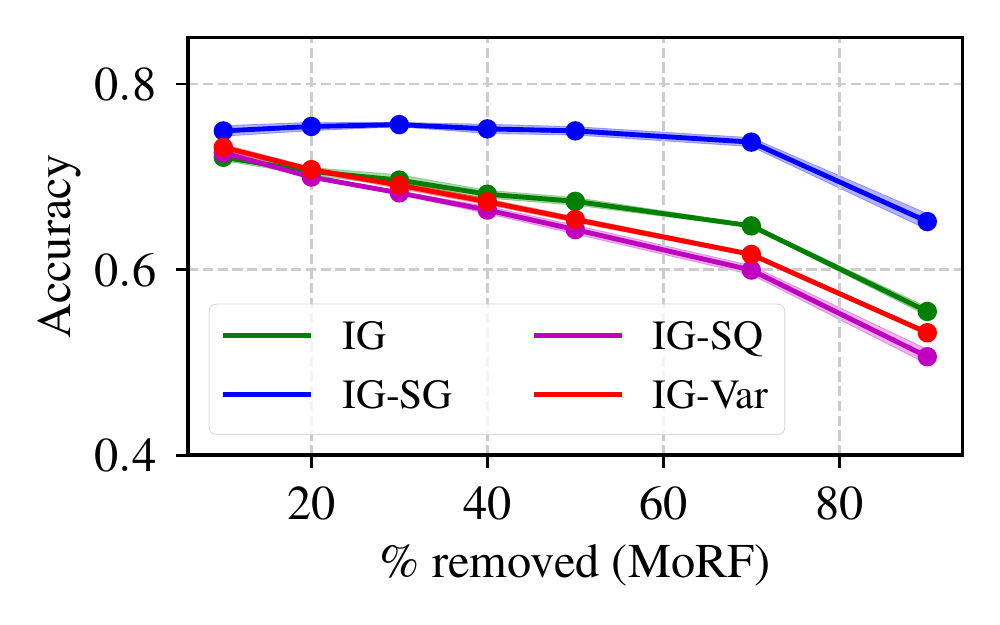}
    \caption{MoRF, Retrain}
\end{subfigure}
\hfill
\begin{subfigure}[h]{0.48\textwidth}
    \centering
    \includegraphics[width=\figuresuppw]{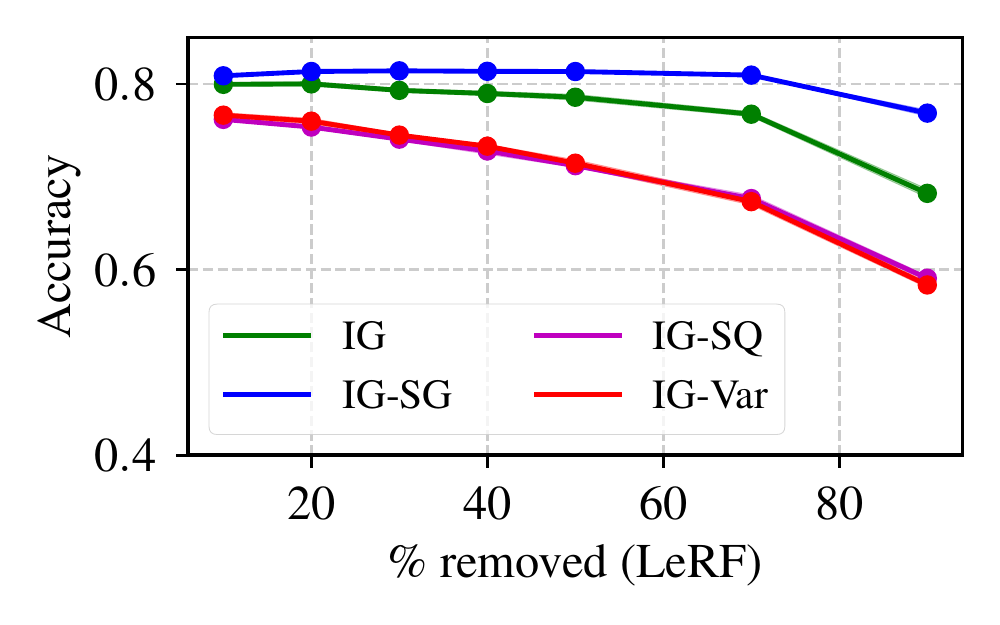}
    \caption{LeRF, Retrain}
\end{subfigure}
\\
\begin{subfigure}[h]{0.48\textwidth}
    \centering
    \includegraphics[width=\figuresuppw]{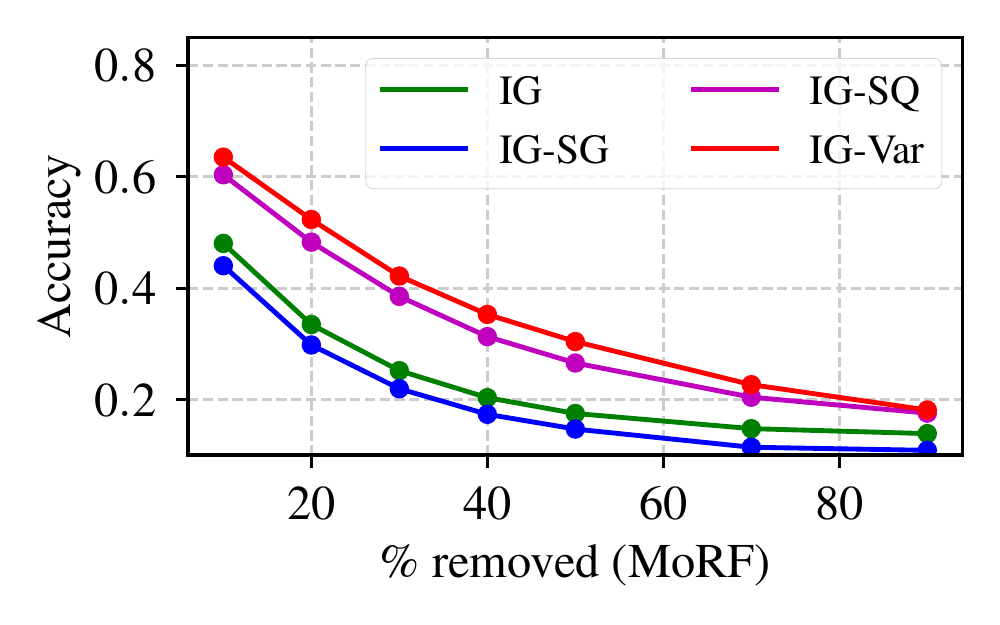}
    \caption{MoRF, No-Retrain}
\end{subfigure}
\hfill
\begin{subfigure}[h]{0.48\textwidth}
    \centering
    \includegraphics[width=\figuresuppw]{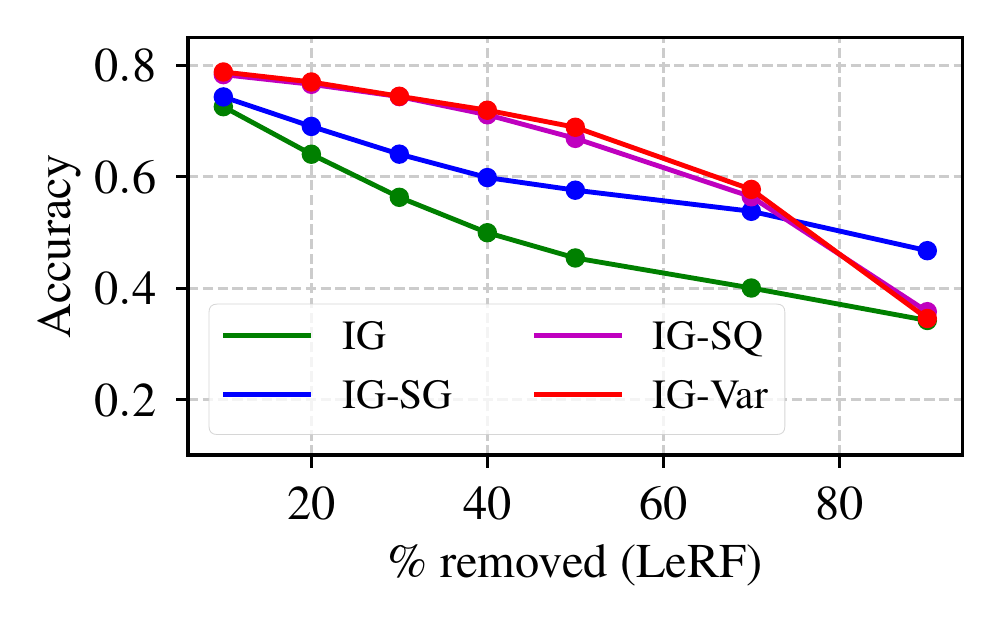}
    \caption{LeRF, No-Retrain}
\end{subfigure}

\caption{Consistency comparison using \textbf{Fixed Value} imputation on \textbf{IG}-based methods on CIFAR-10}
\label{suppfig:cifar-ig-fixed}
\end{figure}

\begin{figure}[h]
\begin{subfigure}[h]{0.48\textwidth}
    \centering
    \includegraphics[width=\figuresuppw]{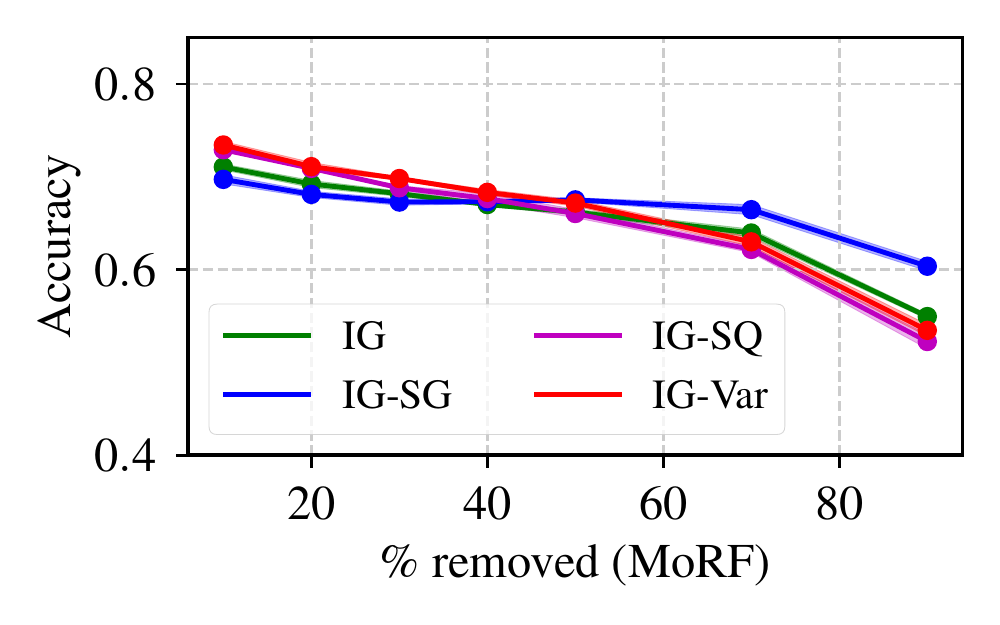}
    \caption{MoRF, Retrain}
\end{subfigure}
\hfill
\begin{subfigure}[h]{0.48\textwidth}
    \centering
    \includegraphics[width=\figuresuppw]{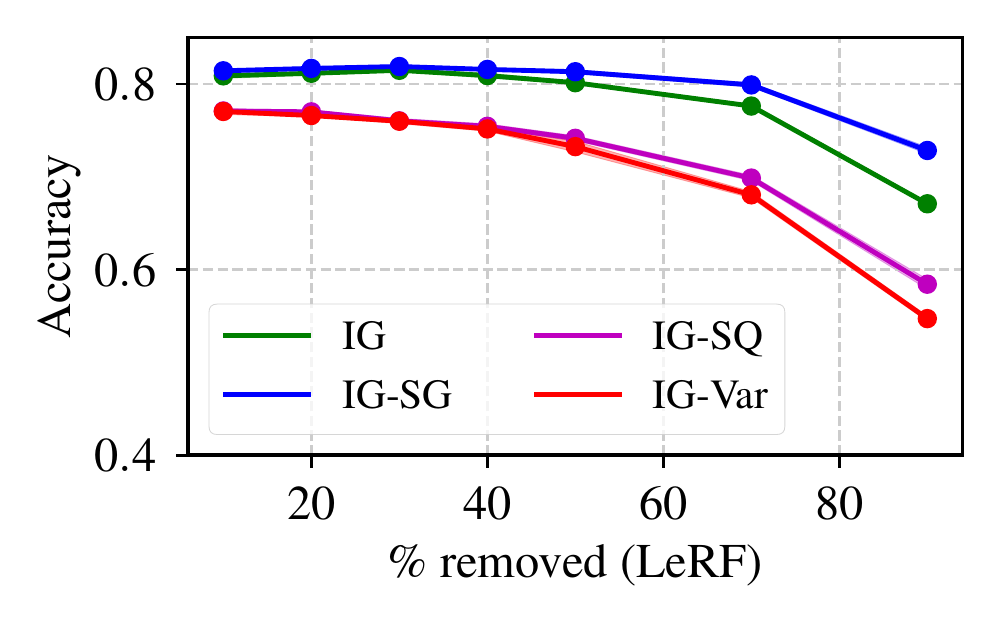}
    \caption{LeRF, Retrain}
\end{subfigure}
\\
\begin{subfigure}[h]{0.48\textwidth}
    \centering
    \includegraphics[width=\figuresuppw]{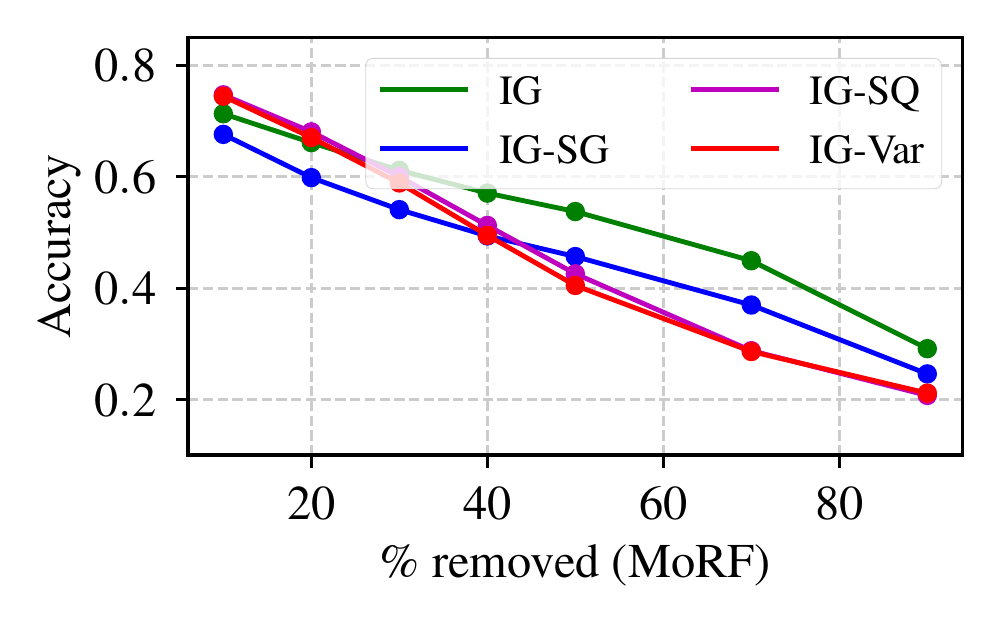}
    \caption{MoRF, No-Retrain}
\end{subfigure}
\hfill
\begin{subfigure}[h]{0.48\textwidth}
    \centering
    \includegraphics[width=\figuresuppw]{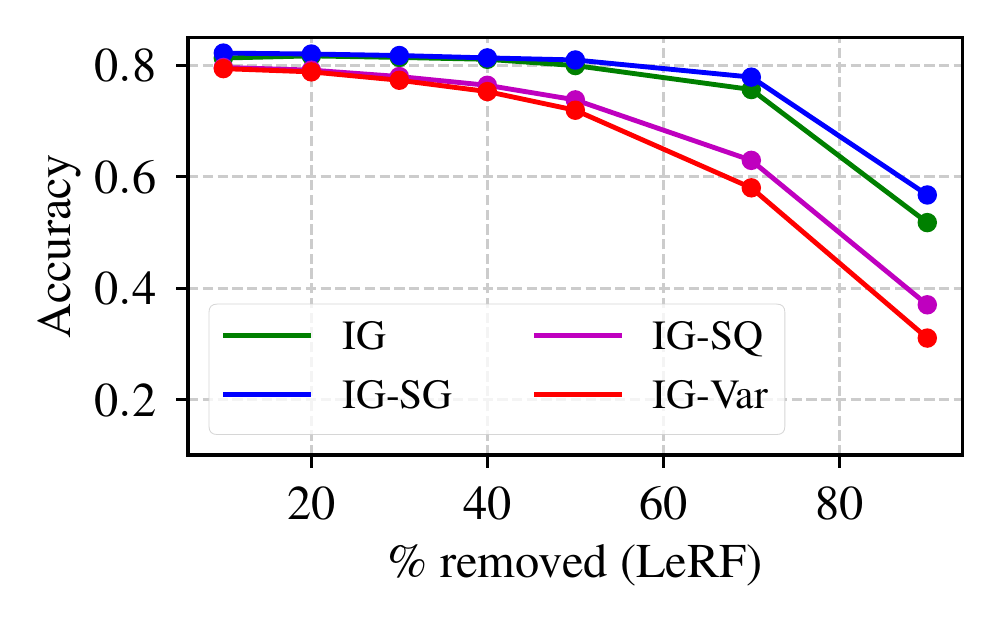}
    \caption{LeRF, No-Retrain}
\end{subfigure}

\caption{Consistency comparison using \textbf{Noisy Linear} imputation on \textbf{IG}-based methods on CIFAR-10}
\label{suppfig:cifar-ig-linear}
\end{figure}

\begin{figure}[h]
\begin{subfigure}[h]{0.48\textwidth}
    \centering
    \includegraphics[width=\figuresuppw]{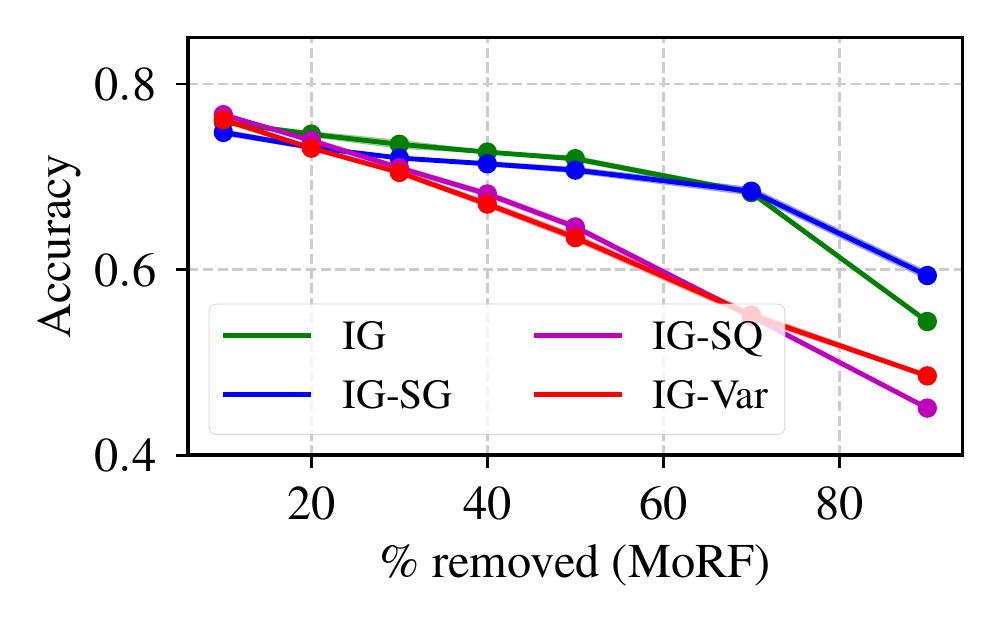}
    \caption{MoRF, Retrain}
\end{subfigure}
\hfill
\begin{subfigure}[h]{0.48\textwidth}
    \centering
    \includegraphics[width=\figuresuppw]{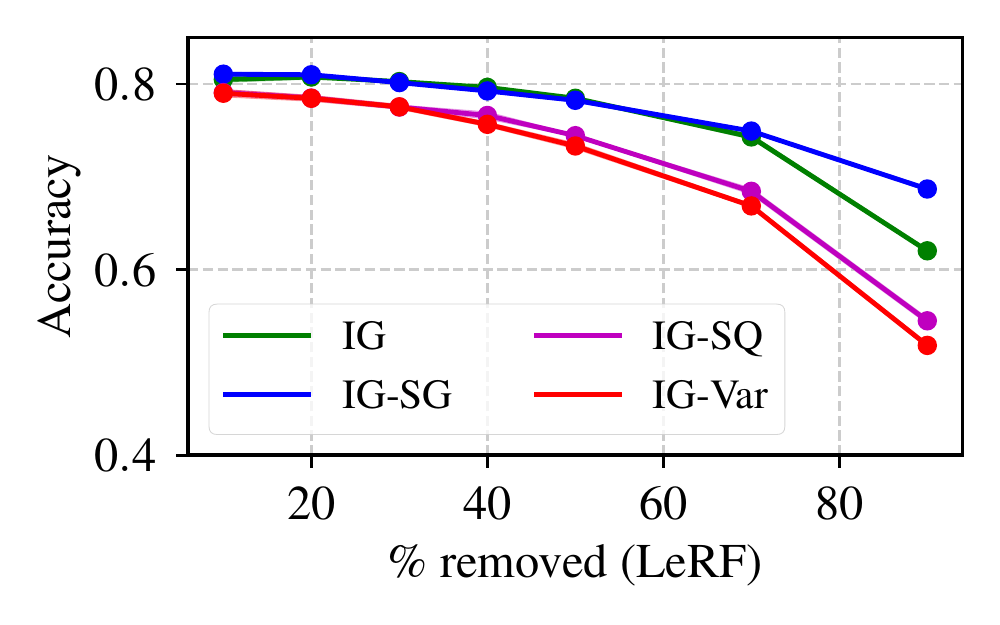}
    \caption{LeRF, Retrain}
\end{subfigure}
\\
\begin{subfigure}[h]{0.48\textwidth}
    \centering
    \includegraphics[width=\figuresuppw]{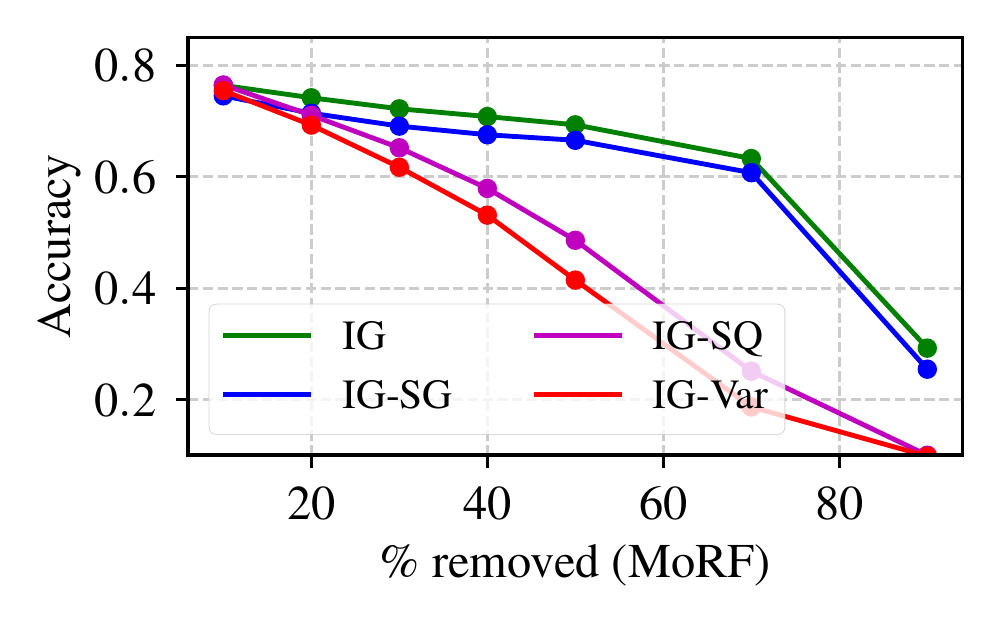}
    \caption{MoRF, No-Retrain}
\end{subfigure}
\hfill
\begin{subfigure}[h]{0.48\textwidth}
    \centering
    \includegraphics[width=\figuresuppw]{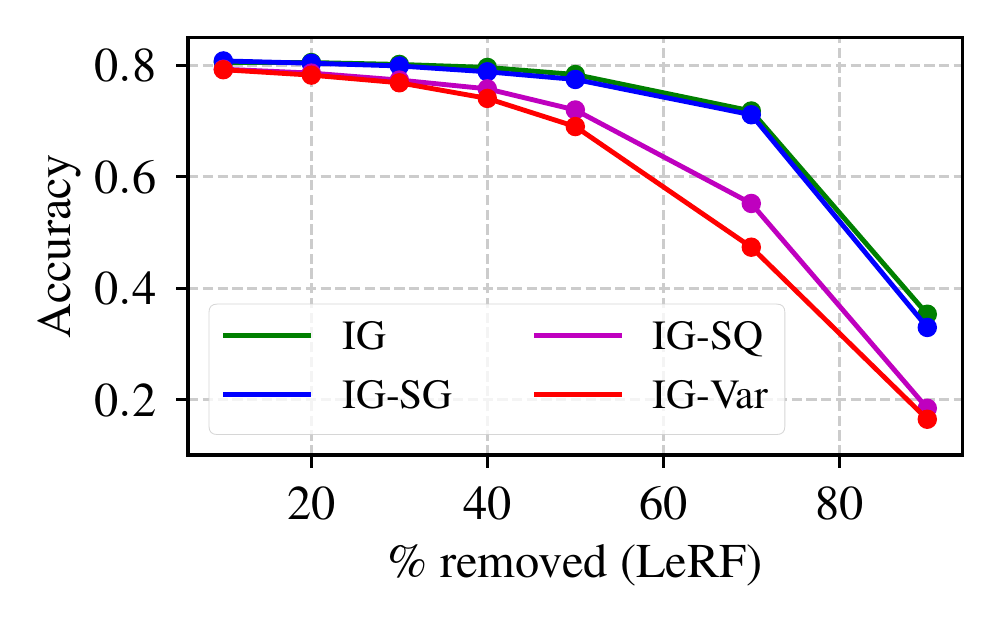}
    \caption{LeRF, No-Retrain}
\end{subfigure}

\caption{Consistency comparison using \textbf{GAN} imputation on \textbf{IG}-based methods on CIFAR-10}
\label{suppfig:cifar-ig-gan}
\end{figure}

\begin{figure}[h]
\begin{subfigure}[h]{0.48\textwidth}
    \centering
    \includegraphics[width=\figuresuppw]{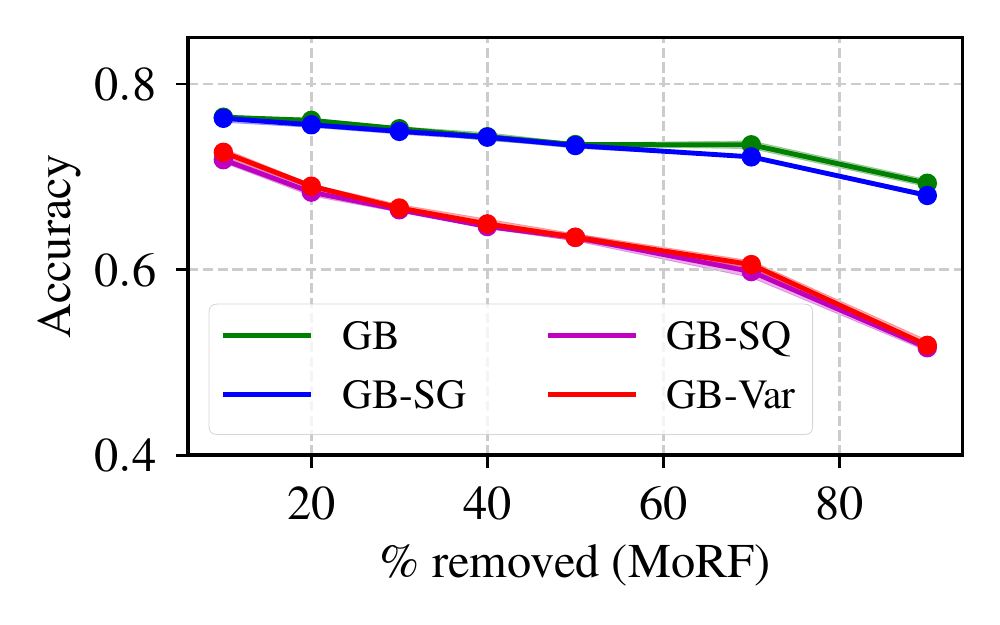}
    \caption{MoRF, Retrain}
\end{subfigure}
\hfill
\begin{subfigure}[h]{0.48\textwidth}
    \centering
    \includegraphics[width=\figuresuppw]{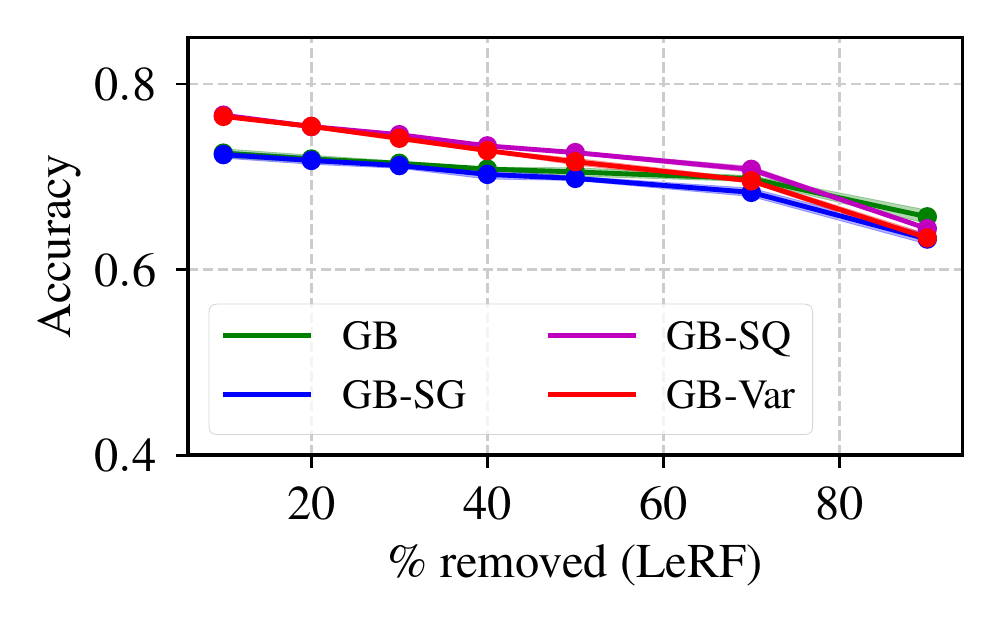}
    \caption{LeRF, Retrain}
\end{subfigure}
\\
\begin{subfigure}[h]{0.48\textwidth}
    \centering
    \includegraphics[width=\figuresuppw]{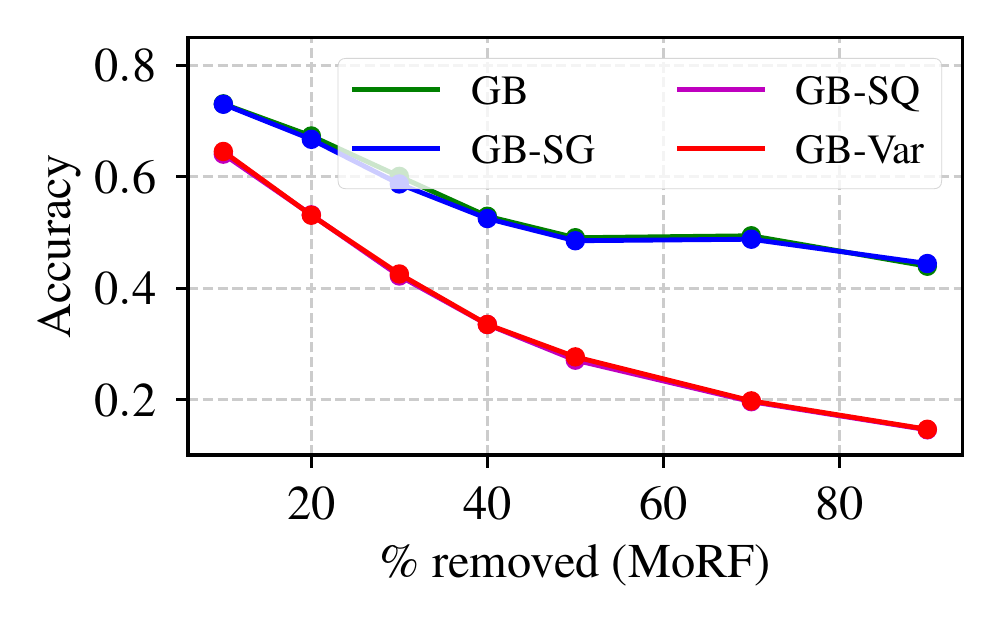}
    \caption{MoRF, No-Retrain}
\end{subfigure}
\hfill
\begin{subfigure}[h]{0.48\textwidth}
    \centering
    \includegraphics[width=\figuresuppw]{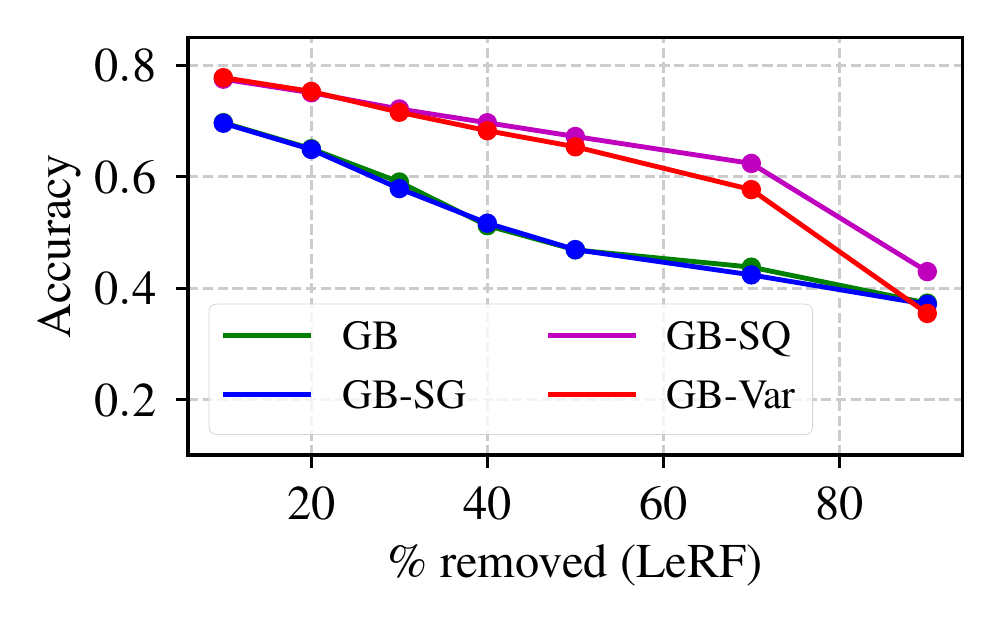}
    \caption{LeRF, No-Retrain}
\end{subfigure}

\caption{Consistency comparison using \textbf{Fixed Value} imputation on \textbf{GB}-based methods on CIFAR-10}
\label{suppfig:cifar-gb-fixed}
\end{figure}

\begin{figure}[h]
\begin{subfigure}[h]{0.48\textwidth}
    \centering
    \includegraphics[width=\figuresuppw]{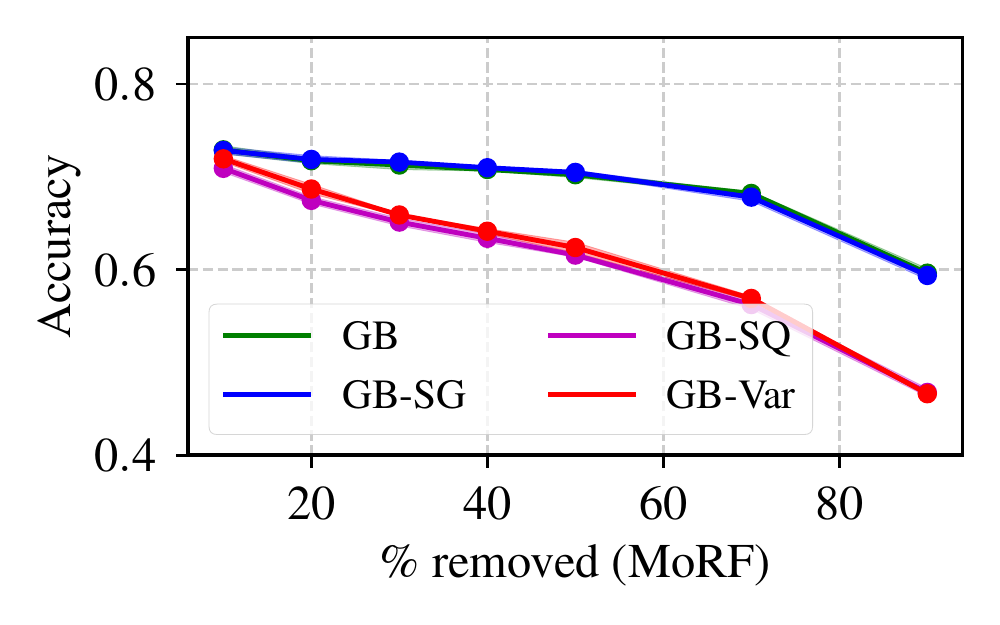}
    \caption{MoRF, Retrain}
\end{subfigure}
\hfill
\begin{subfigure}[h]{0.48\textwidth}
    \centering
    \includegraphics[width=\figuresuppw]{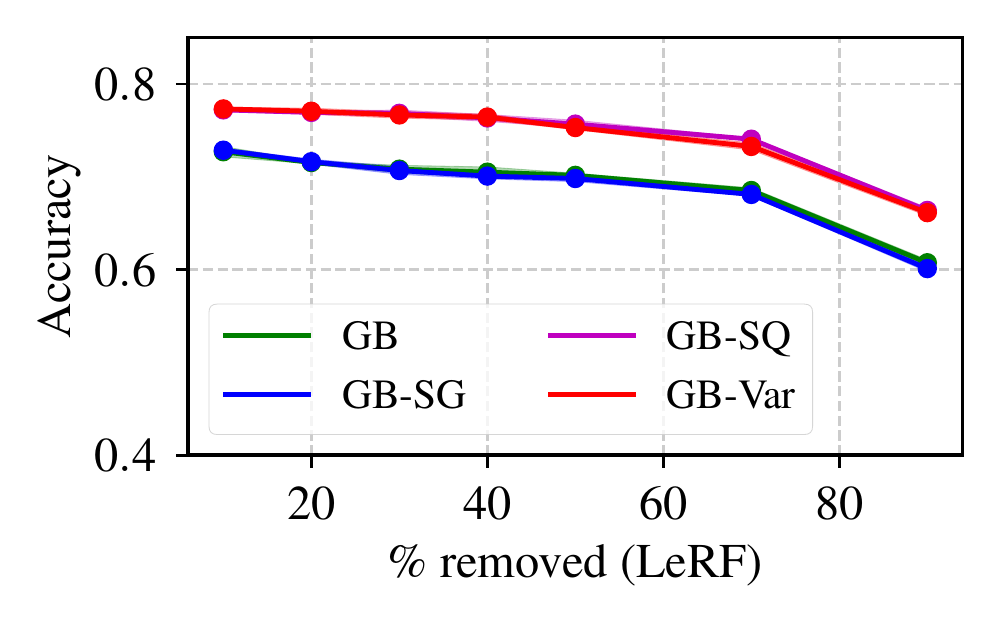}
    \caption{LeRF, Retrain}
\end{subfigure}
\\
\begin{subfigure}[h]{0.48\textwidth}
    \centering
    \includegraphics[width=\figuresuppw]{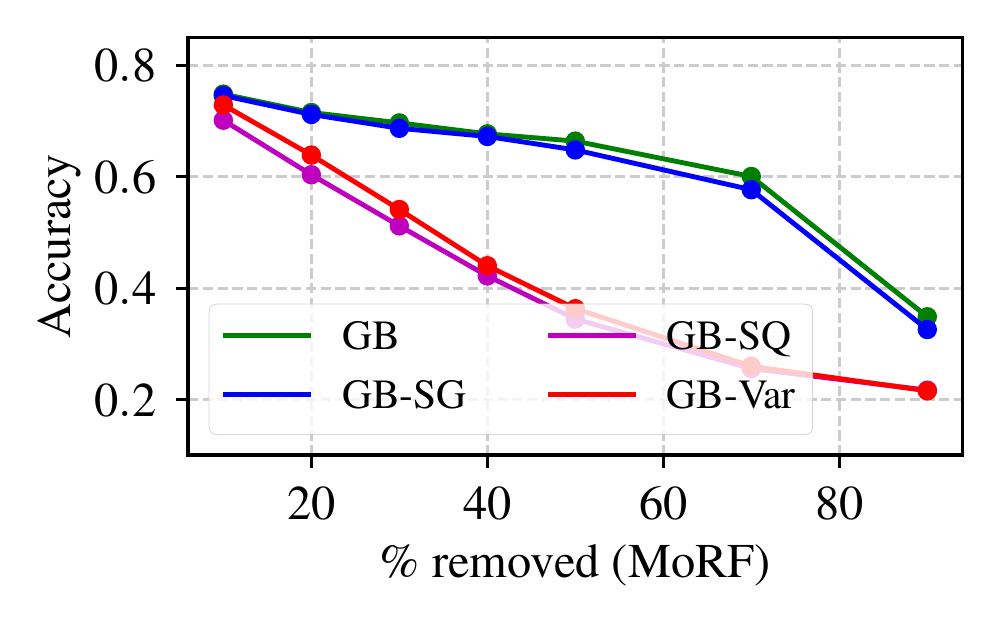}
    \caption{MoRF, No-Retrain}
\end{subfigure}
\hfill
\begin{subfigure}[h]{0.48\textwidth}
    \centering
    \includegraphics[width=\figuresuppw]{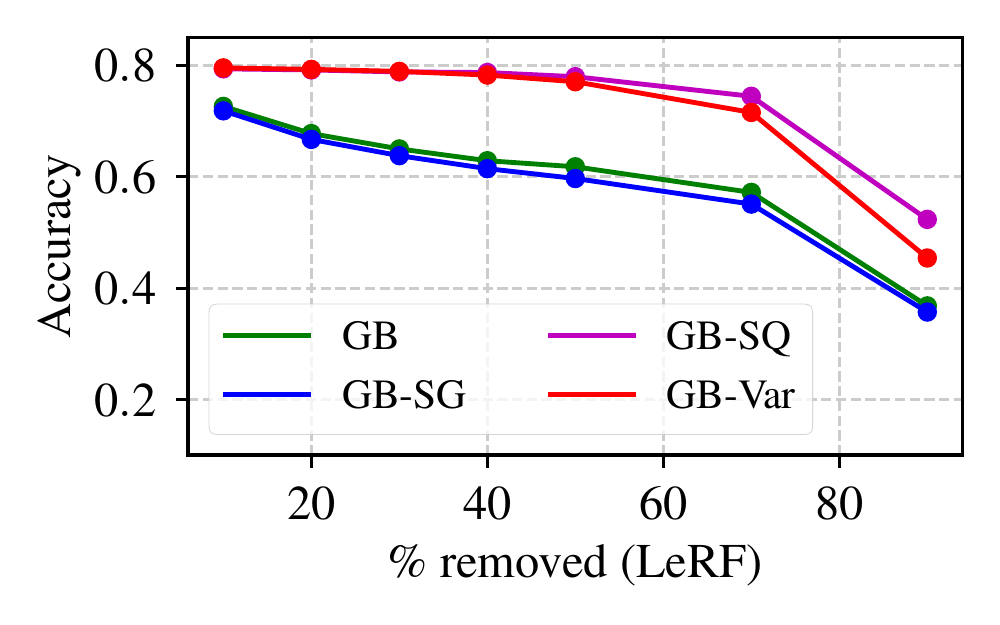}
    \caption{LeRF, No-Retrain}
\end{subfigure}

\caption{Consistency comparison using \textbf{Noisy Linear} imputation on \textbf{GB}-based methods on CIFAR-10}
\label{suppfig:cifar-gb-linear}
\end{figure}

\begin{figure}[h]
\begin{subfigure}[h]{0.48\textwidth}
    \centering
    \includegraphics[width=\figuresuppw]{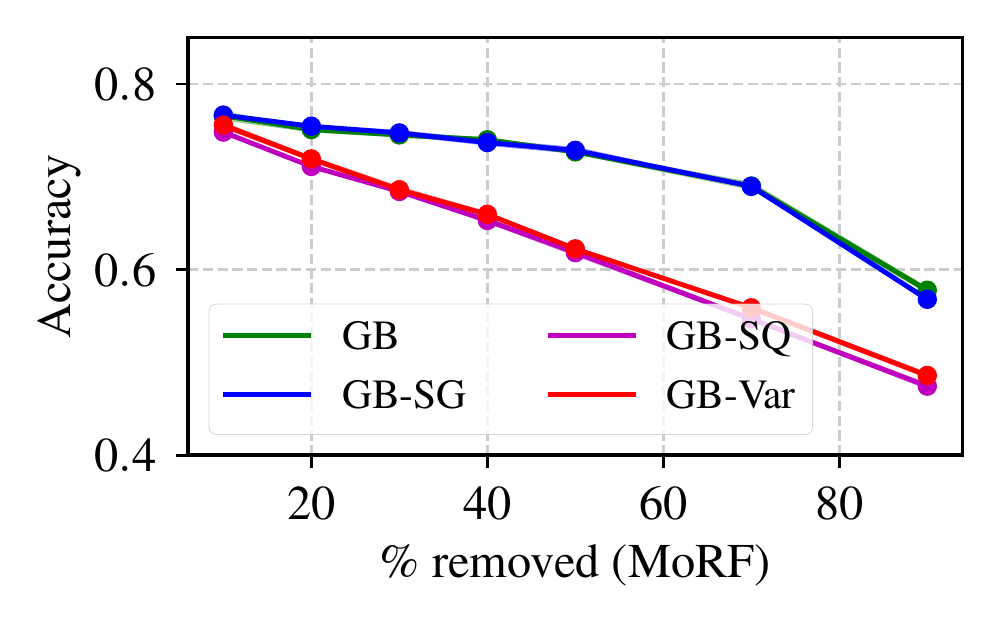}
    \caption{MoRF, Retrain}
\end{subfigure}
\hfill
\begin{subfigure}[h]{0.48\textwidth}
    \centering
    \includegraphics[width=\figuresuppw]{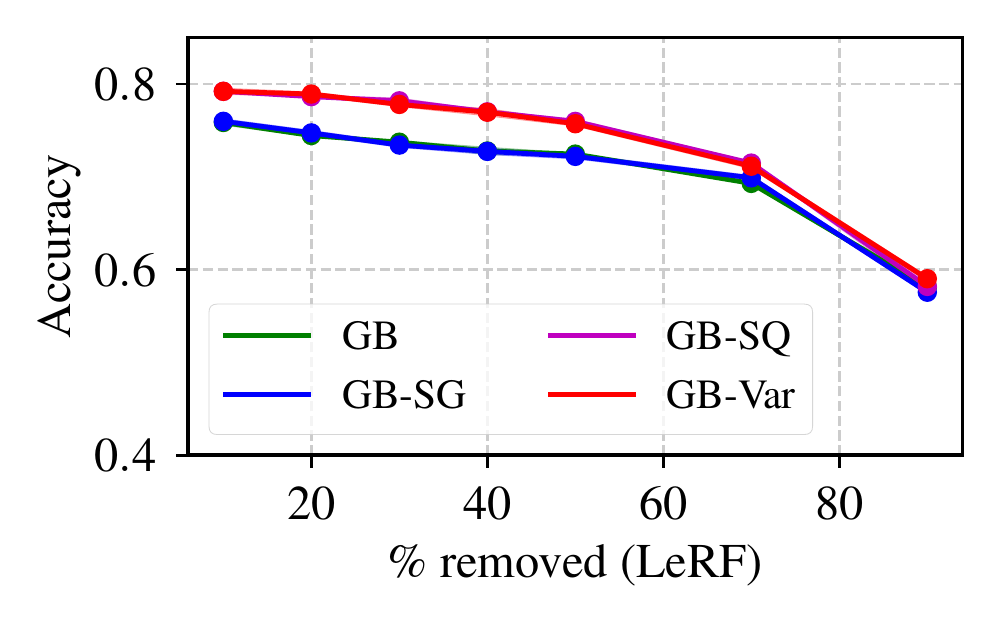}
    \caption{LeRF, Retrain}
\end{subfigure}
\\
\begin{subfigure}[h]{0.48\textwidth}
    \centering
    \includegraphics[width=\figuresuppw]{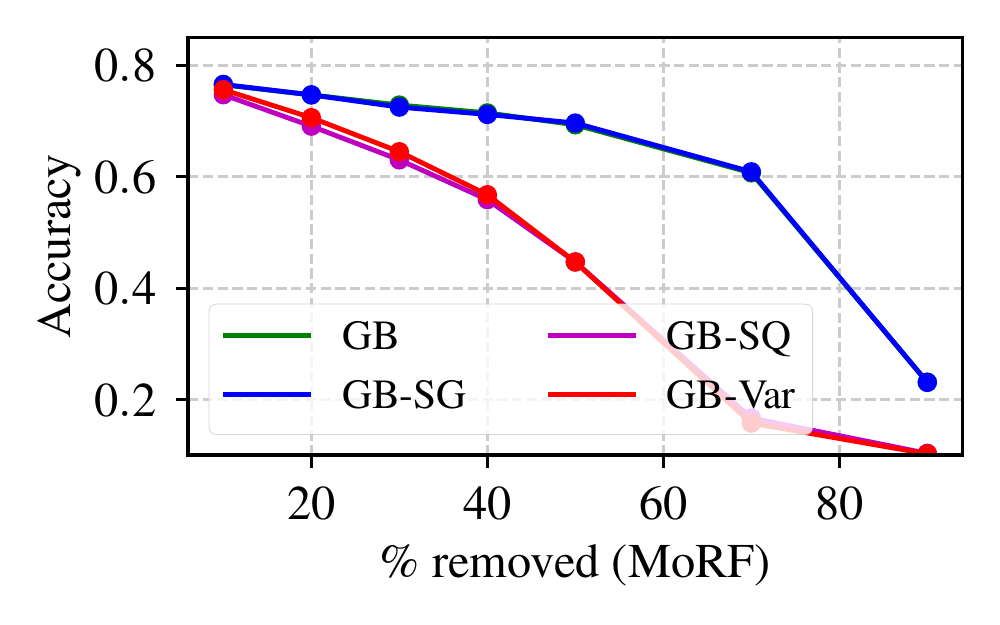}
    \caption{MoRF, No-Retrain}
\end{subfigure}
\hfill
\begin{subfigure}[h]{0.48\textwidth}
    \centering
    \includegraphics[width=\figuresuppw]{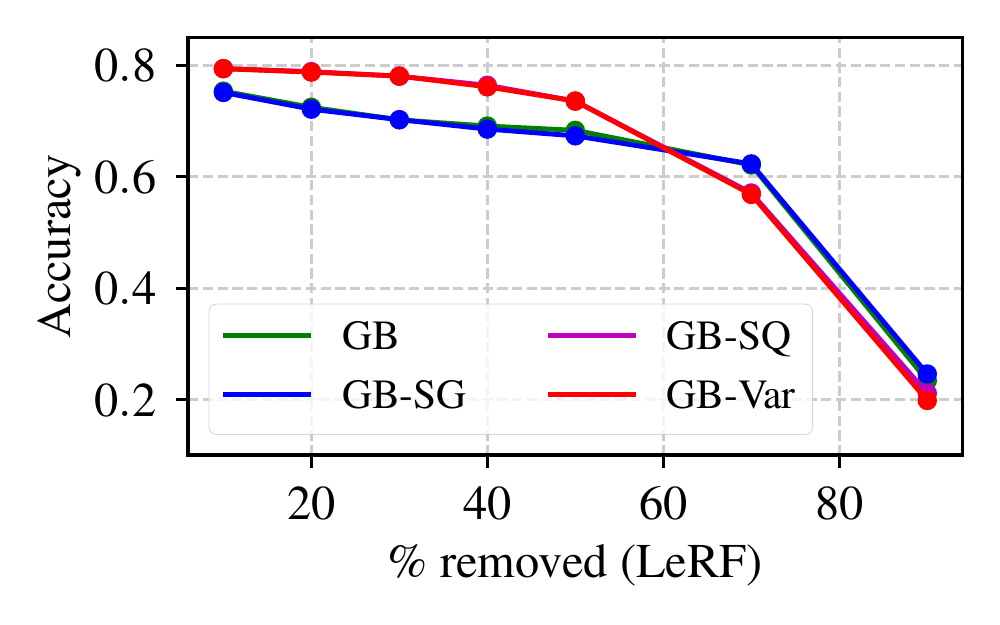}
    \caption{LeRF, No-Retrain}
\end{subfigure}

\caption{Consistency comparison using \textbf{GAN} imputation on \textbf{GB}-based methods on CIFAR-10}
\label{suppfig:cifar-gb-gan}
\end{figure}

\section{Additional Experiments on Food-101}
\label{sec: results food}
\subsection{Implementation Details}
We trained a vanilla ResNet-50 \cite{he2016deep} on Food-101 \cite{bossard2014food}. Concretely, we trained the model using the SGD optimizer. Additionally the model was trained with the initial learning rate of 0.01. The learning rate was reduced by factor of 0.1 after every 10 epochs. In total, we trained 40 epochs with a batch size of 32 and the model achieved the accuracy of 81.67\% on the test set. To run the GAN imputation operator, we first trained a GAIN model on Food-101 as introduced in \cref{sec: gain}. We used the hyper-parameters $\alpha=100$ and $hr=0.1$ and trained the GAIN model with the batch size of 32 for 100 epochs. We computed the eight explanations and run ROAD and ROAR evaluation using the same settings as introduced in \cref{sec: supp cifar implementation} for CIFAR-10. 

\subsection{Correlation Analysis}
In \cref{suptab:food spearman}, we show a full view of Spearman Correlation of rankings given by eight different evaluation strategies (``Retrain"/``No-Retrain", MoRF/LeRF, and fixed/Noisy Linear/GAN imputation) on Food-101. In the table, results marked in bold indicate the consistency of using three imputation operators. We observe that the consistency between the respective Retrain and No-Retrain methods is still very high, which confirms that the efficiency gains reported in the main paper can be realized for larger data sets. Consistency between MoRf/LeRF is improved (over fixed imputation) when using retraining, but decreases slightly when the No-Retraining approach is used. Because the curves are often very close on this dataset (in particular for the No-Retraining setup), small differences might already lead to a change in the ranking and the results are in general noisier than on CIFAR-10. In summary, we observe similar trends, although the consistency gain between MoRF/LeRF in No-Retrain is not as pronounced. Nevertheless, a perfect agreement between MoRF/LeRF might not be desirable. 

\renewcommand{\wstd}[2]{\makecell{#1\small\\$\pm$#2}}
\begin{table}
\centering
\scalebox{0.8}{\begin{tabular}{|c|c|ccc|ccc|ccc|ccc|}
\cline{3-14}
\multicolumn{2}{c|}{\multirow{3}{*}{}} & \multicolumn{3}{c|}{Retrain} & \multicolumn{3}{c|}{No-Retrain} & \multicolumn{3}{c|}{Retrain} & \multicolumn{3}{c|}{No-Retrain}\\
\multicolumn{2}{c|}{}  & \multicolumn{3}{c|}{MoRF} & \multicolumn{3}{c|}{MoRF} & \multicolumn{3}{c|}{LeRF} & \multicolumn{3}{c|}{LeRF}\\ 

\cline{3-14} \multicolumn{2}{c|}{} & fixed$^\dagger$ & lin & gan  & fixed & lin$^{*}$ & gan &  fixed  & lin & gan  & fixed & lin & gan\\ 
\hline
\multirow{3}{*}{\begin{tabular}[c]{c@{}c@{}c@{}} Retrain\\ MoRF\end{tabular}} & fixed$^\dagger$ &  \wstd{1.00}{0.00} & & &&& &&& &&&\\
&lin & \wstd{0.48}{0.03} & \wstd{1.00}{0.00} & &&& &&& &&&\\
&gan &  \wstd{0.50}{0.04}  & \wstd{0.79}{0.03}& \wstd{1.00}{0.00} &&& &&& &&& \\\hline
\multirow{3}{*}{\begin{tabular}[c]{c@{}c@{}c@{}} No-Retrain\\ MoRF\end{tabular}} & fixed &  \wstd{\textbf{0.12}}{0.01} &  \wstd{0.57}{0.02} & \wstd{0.50}{0.01} & \wstd{1.00}{0.00} && &&& &&&\\
& lin$^{*}$  & \wstd{{0.61}}{0.01} & \wstd{\textbf{0.81}}{0.02}  & \wstd{0.67}{0.04}  & \wstd{0.31}{0.01} & \wstd{1.00}{0.00}& &&& &&& \\
&gan & \wstd{0.74}{0.01} & \wstd{0.79}{0.02} & \wstd{\textbf{0.67}}{0.04} & \wstd{0.35}{0.01}& \wstd{0.86}{0.00} & \wstd{1.00}{0.00} &&& &&&\\\hline
\multirow{3}{*}{\begin{tabular}[c]{c@{}c@{}c@{}} Retrain\\ LeRF\end{tabular}} & fixed &  \wstd{\textbf{-0.26}}{0.02}& \wstd{0.41}{0.02}  & \wstd{0.30}{0.02}  & \wstd{0.53}{0.01} & \wstd{0.10}{0.01} & \wstd{0.11}{0.01} & \wstd{1.00}{0.00} && &&&  \\
&lin & \wstd{-0.40}{0.02} & \wstd{\textbf{0.26}}{0.04}  & \wstd{0.19}{0.04}  & \wstd{0.30}{0.03}  & \wstd{-0.05}{0.01} &\wstd{0.09}{0.01} & \wstd{0.83}{0.01} & \wstd{1.00}{0.00} & &&& \\
&gan &  \wstd{-0.18}{0.01}  & \wstd{0.46}{0.04} & \wstd{\textbf{0.32}}{0.04} & \wstd{0.50}{0.03} & \wstd{0.13}{0.02} & \wstd{0.14}{0.03} & \wstd{0.89}{0.02} & \wstd{0.83}{0.01} & \wstd{1.00}{0.00} &&& \\\hline
\multirow{3}{*}{\begin{tabular}[c]{c@{}c@{}c@{}} No-Retrain\\ LeRF\end{tabular}} & fixed  & \wstd{0.79}{0.02}  &  \wstd{0.79}{0.03}  & \wstd{{0.63}}{0.05} & \wstd{\textbf{0.32}}{0.01} & \wstd{0.85}{0.00} & \wstd{{0.89}}{0.00} & \wstd{\textbf{0.02}}{0.01} & \wstd{-0.15}{0.02} & \wstd{0.10}{0.03} & \wstd{1.00}{0.00} &&\\
& lin &\wstd{-0.28}{0.02}  &  \wstd{0.35}{0.02}& \wstd{0.28}{0.04} & \wstd{{0.46}}{0.00} & \wstd{\textbf{-0.03}}{0.00} & \wstd{-0.06}{0.00} & \wstd{0.89}{0.01} & \wstd{\textbf{0.81}}{0.02} & \wstd{0.87}{0.01} &\wstd{-0.11}{0.00} & \wstd{1.00}{0.00} &\\ 
&gan & \wstd{-0.45}{0.02} & \wstd{-0.08}{0.03} & \wstd{-0.04}{0.04} & \wstd{0.23}{0.00} & \wstd{-0.37}{0.00} & \wstd{\textbf{-0.44}}{0.00} & \wstd{0.58}{0.01} & \wstd{0.61}{0.01} & \wstd{\textbf{0.54}}{0.00} & \wstd{-0.41}{0.00} & \wstd{0.70}{0.00} & \wstd{1.00}{0.00}\\\hline
\end{tabular}}
\caption{\textbf{Food-10}: Rank Correlations between all evaluation strategies used with standard deviations computed by considering the rankings obtained through five consecutive runs as independent. The ROAR benchmark is marked by $^\dagger$ and our ROAD by $^{*}$. Bold results highlight the consistency between Retrain and No-Retrain (still very high) as well as MoRF and LeRF evaluation strategies using different imputation operators (fair increase when using Noisy Linear and GAN imputations instead of fixed imputation in ``Retrain", decrease in ``No-Retrain").
\label{suptab:food spearman}}
\end{table}

\subsection{Extended Figures}
\begin{figure}[]
\begin{subfigure}[h]{0.4\textwidth}
    \centering
    \includegraphics[width=\textwidth]{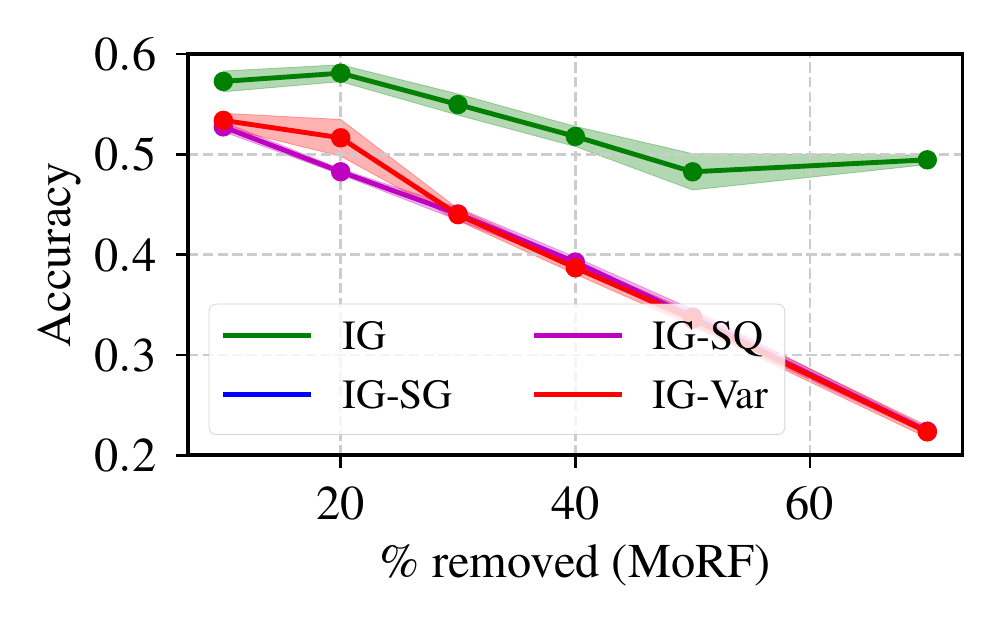}
    \caption{MoRF, Retrain}
\end{subfigure}
\hfill
\begin{subfigure}[h]{0.4\textwidth}
    \centering
    \includegraphics[width=\textwidth]{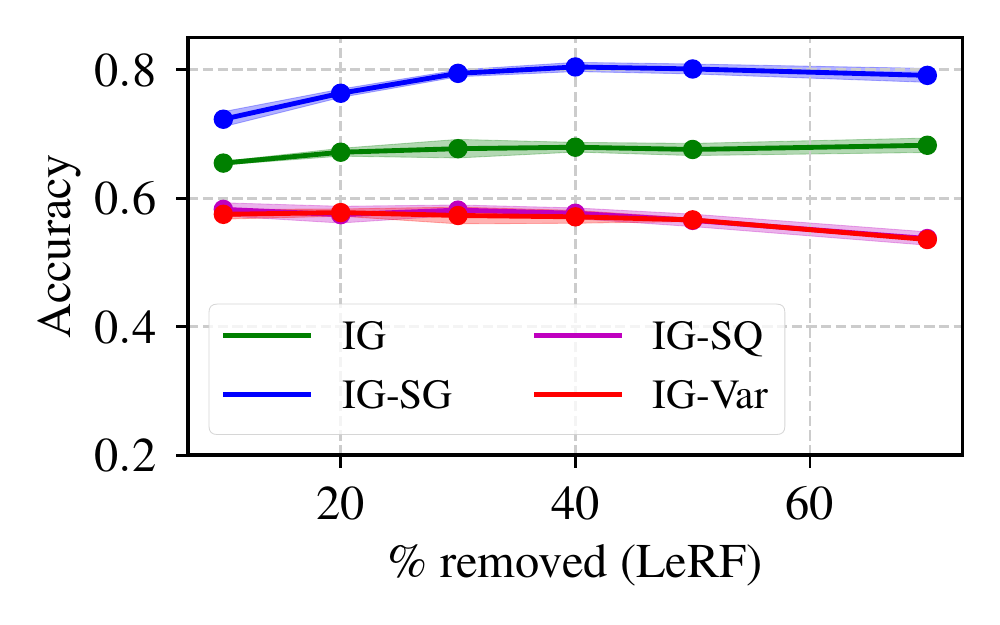}
    \caption{LeRF, Retrain}
\end{subfigure}
\\
\begin{subfigure}[h]{0.4\textwidth}
    \centering
    \includegraphics[width=\textwidth]{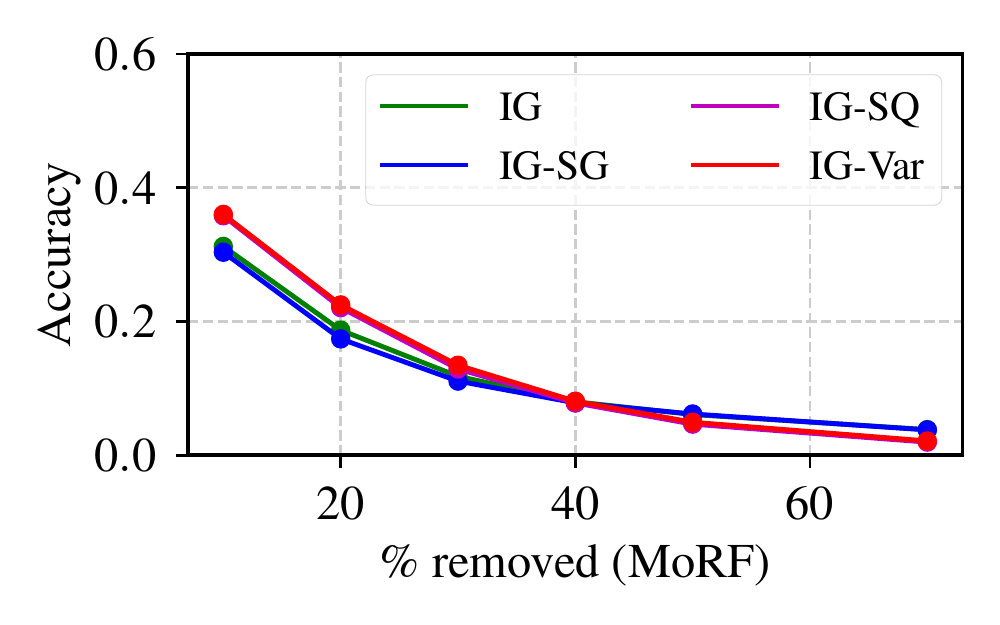}
    \caption{MoRF, No-Retrain}
\end{subfigure}
\hfill
\begin{subfigure}[h]{0.4\textwidth}
    \centering
    \includegraphics[width=\textwidth]{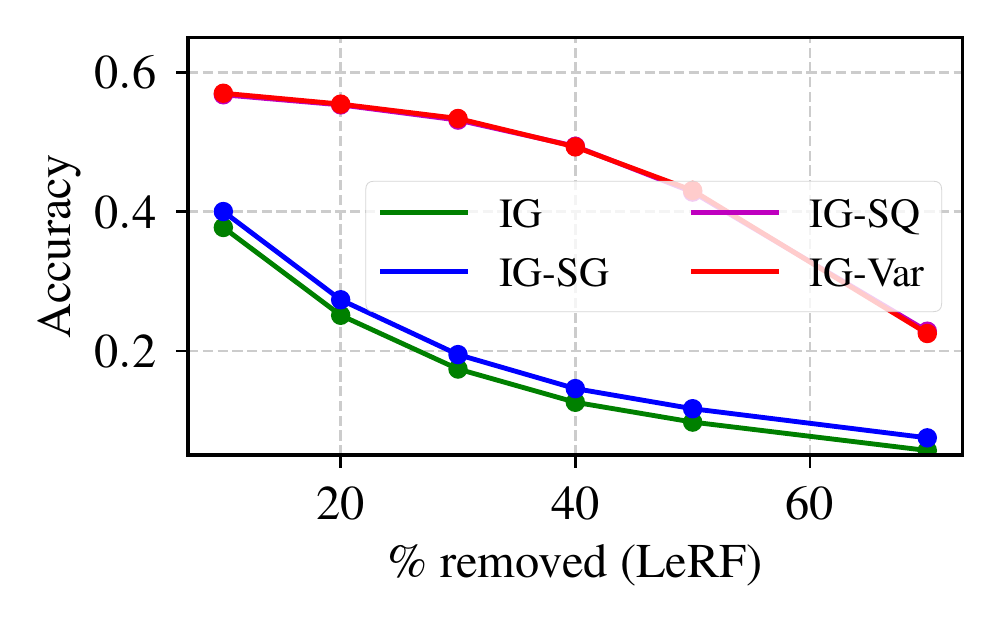}
    \caption{LeRF, No-Retrain}
\end{subfigure}

\caption{Consistency comparison using \textbf{Fixed Value} imputation on \textbf{IG}-based methods on Food-101.}
\label{suppfig:food-ig-fixed}
\end{figure}

\begin{figure}[]
\begin{subfigure}[h]{0.4\textwidth}
    \centering
    \includegraphics[width=\textwidth]{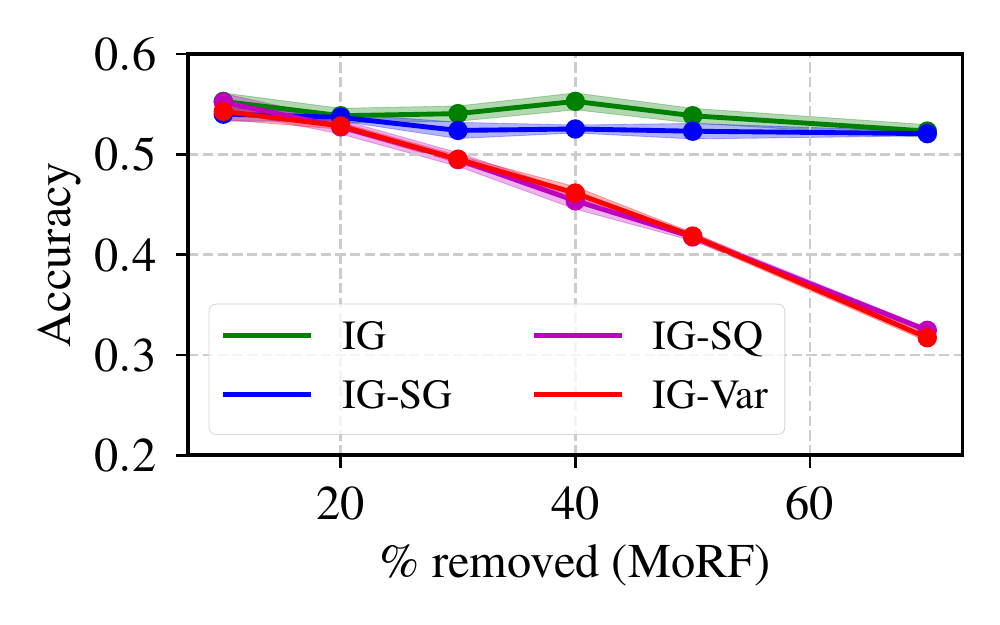}
    \caption{MoRF, Retrain}
\end{subfigure}
\hfill
\begin{subfigure}[h]{0.4\textwidth}
    \centering
    \includegraphics[width=\textwidth]{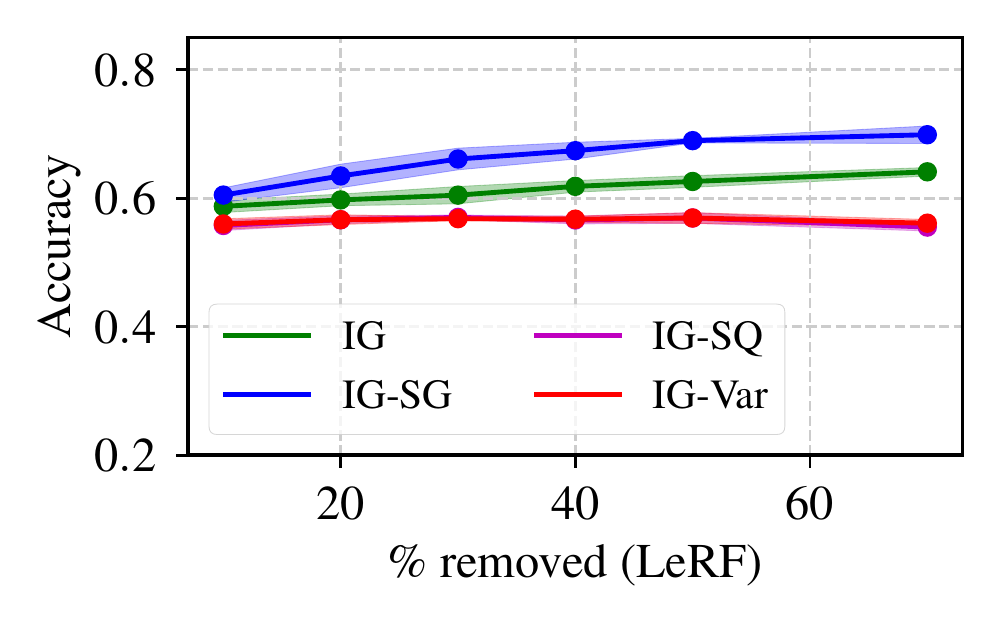}
    \caption{LeRF, Retrain}
\end{subfigure}
\\
\begin{subfigure}[h]{0.4\textwidth}
    \centering
    \includegraphics[width=\textwidth]{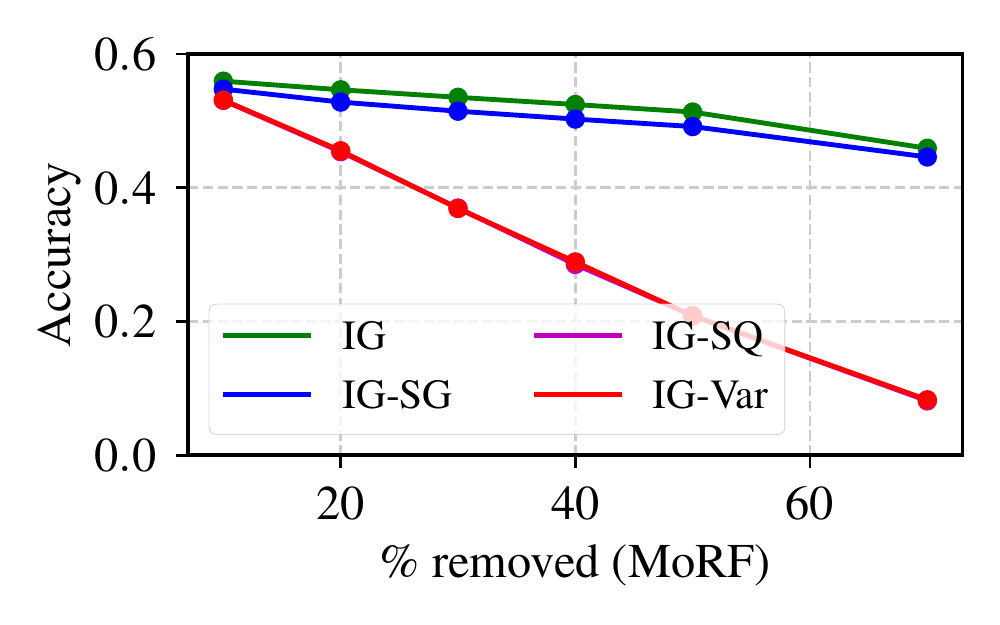}
    \caption{MoRF, No-Retrain}
\end{subfigure}
\hfill
\begin{subfigure}[h]{0.4\textwidth}
    \centering
    \includegraphics[width=\textwidth]{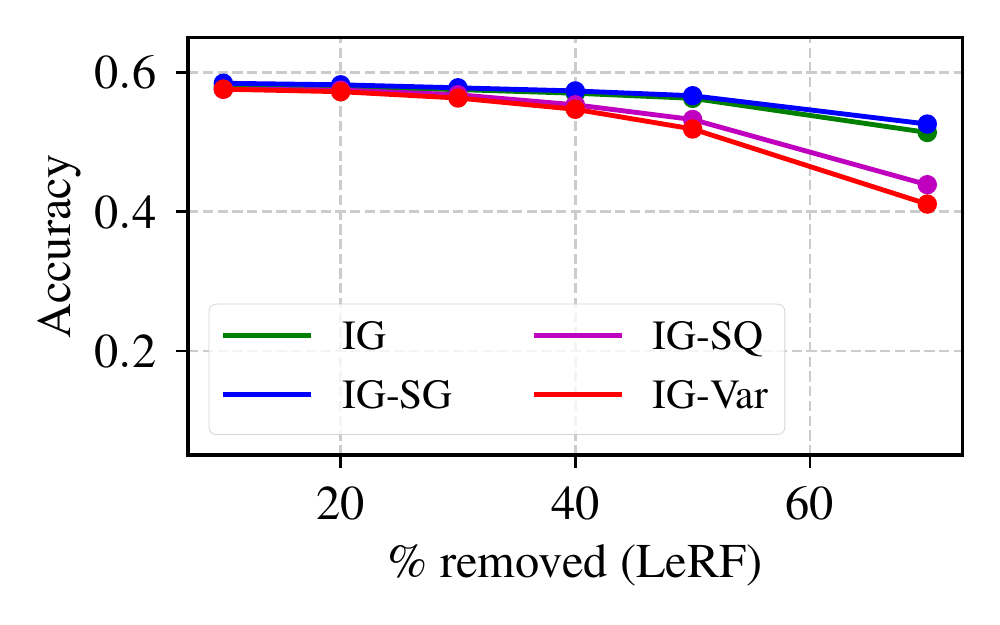}
    \caption{LeRF, No-Retrain}
\end{subfigure}

\caption{Consistency comparison using \textbf{Noisy Linear} imputation on \textbf{IG}-based methods on Food-101.}
\label{suppfig:food-ig-linear}
\end{figure}

\begin{figure}[]
\begin{subfigure}[h]{0.4\textwidth}
    \centering
    \includegraphics[width=\textwidth]{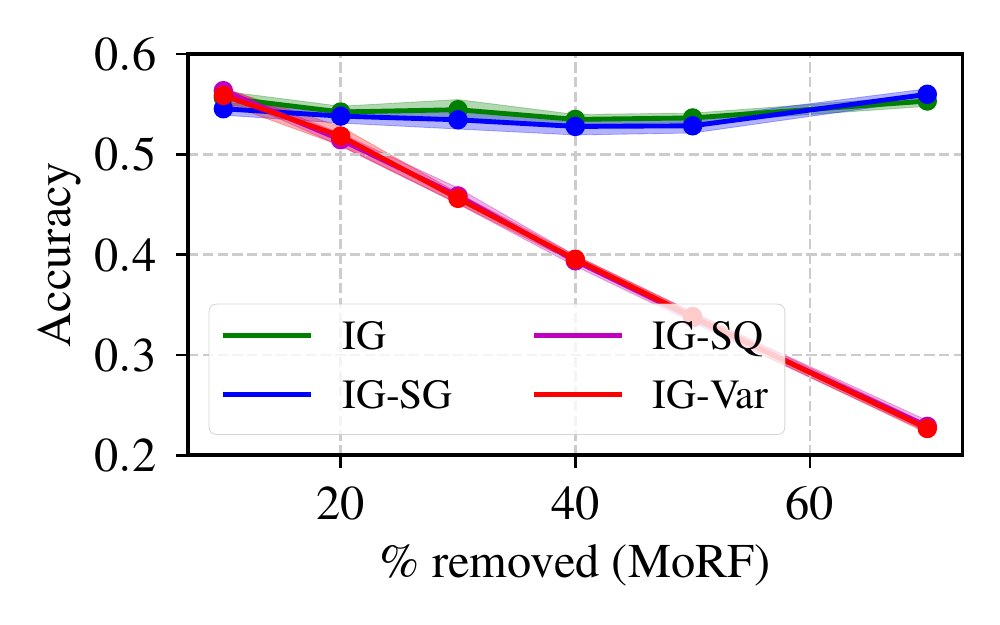}
    \caption{MoRF, Retrain}
\end{subfigure}
\hfill
\begin{subfigure}[h]{0.4\textwidth}
    \centering
    \includegraphics[width=\textwidth]{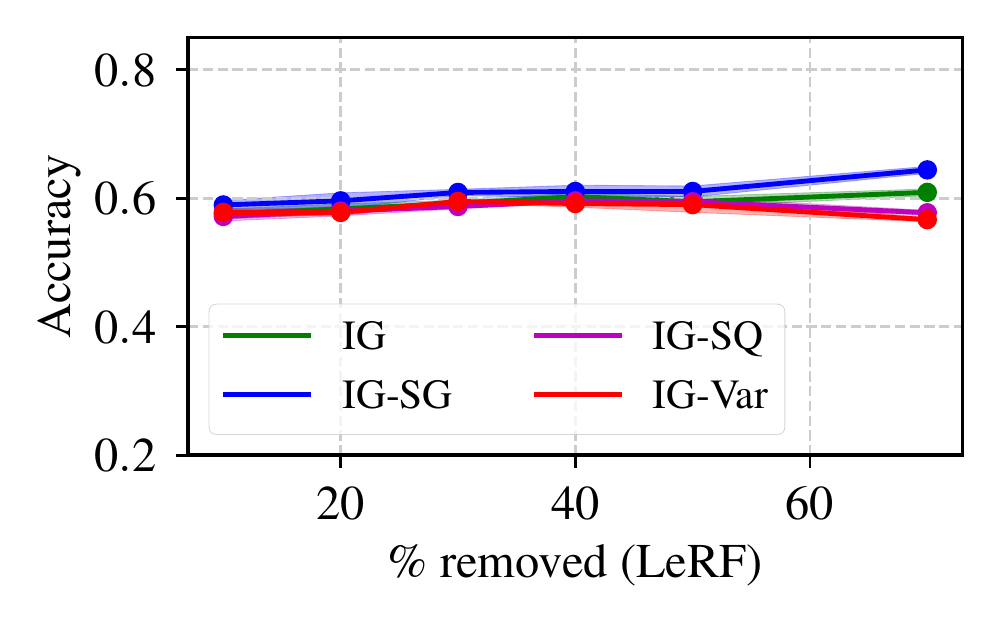}
    \caption{LeRF, Retrain}
\end{subfigure}
\\
\begin{subfigure}[h]{0.4\textwidth}
    \centering
    \includegraphics[width=\textwidth]{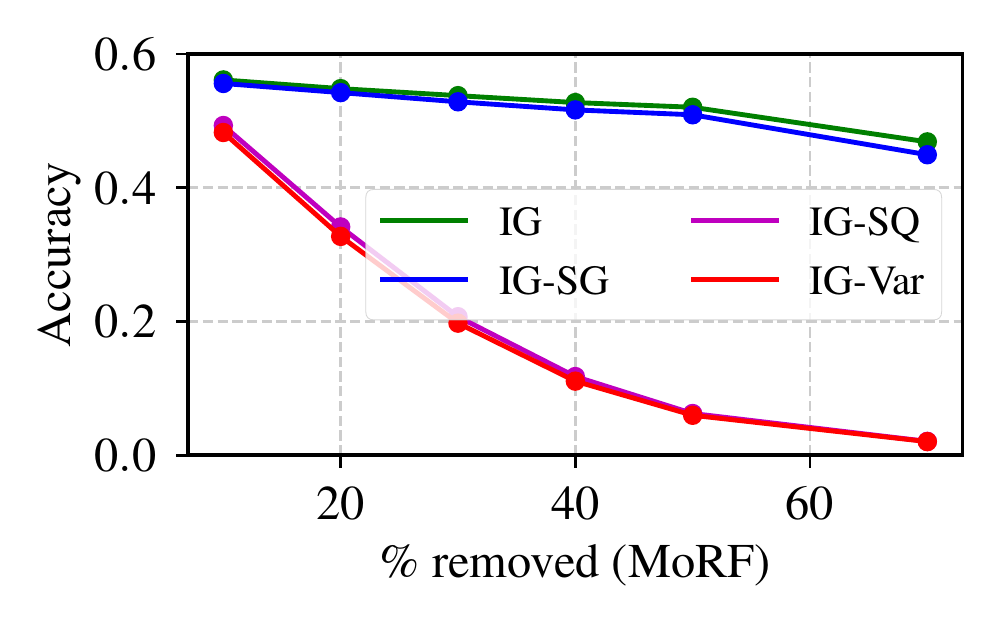}
    \caption{MoRF, No-Retrain}
\end{subfigure}
\hfill
\begin{subfigure}[h]{0.4\textwidth}
    \centering
    \includegraphics[width=\textwidth]{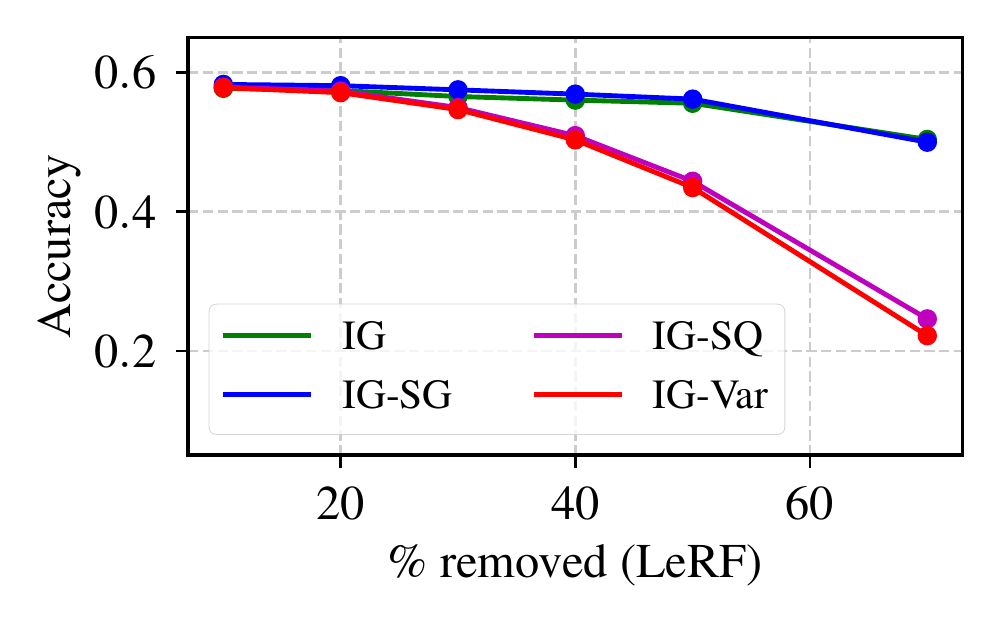}
    \caption{LeRF, No-Retrain}
\end{subfigure}

\caption{Consistency comparison using \textbf{GAN} imputation on \textbf{IG}-based methods on Food-101.}
\label{suppfig:food-ig-gan}
\end{figure}

\begin{figure}[]
\begin{subfigure}[h]{0.48\textwidth}
    \centering
    \includegraphics[width=\figuresuppw]{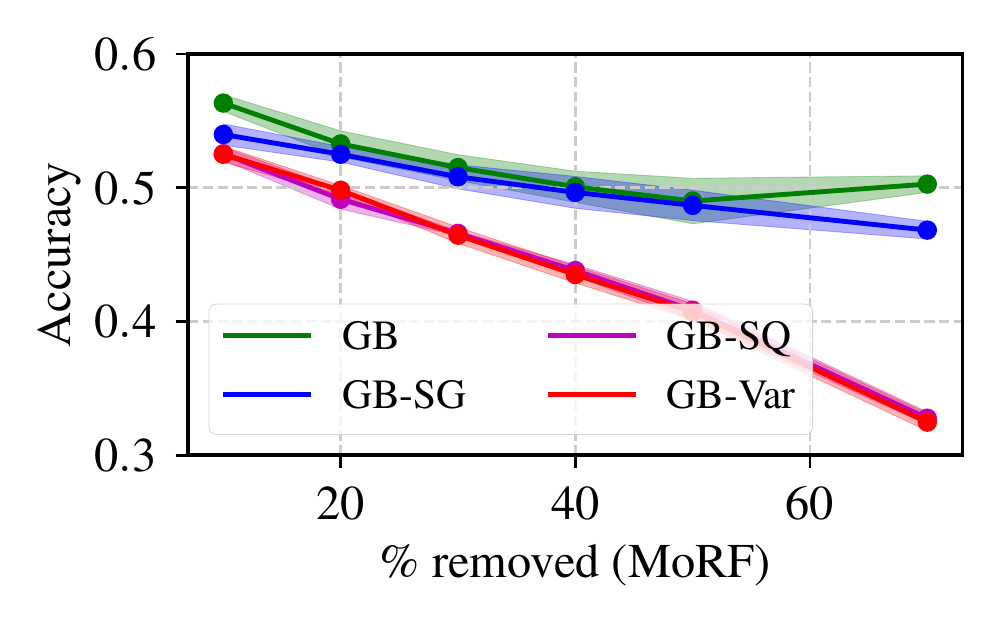}
    \caption{MoRF, Retrain}
\end{subfigure}
\hfill
\begin{subfigure}[h]{0.48\textwidth}
    \centering
    \includegraphics[width=\figuresuppw]{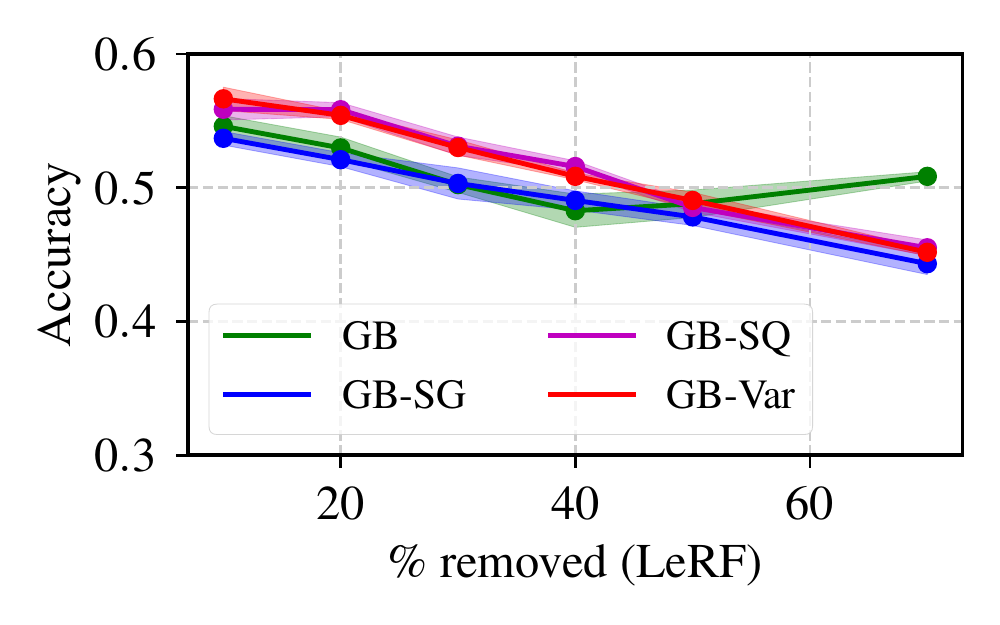}
    \caption{LeRF, Retrain}
\end{subfigure}
\\
\begin{subfigure}[h]{0.48\textwidth}
    \centering
    \includegraphics[width=\figuresuppw]{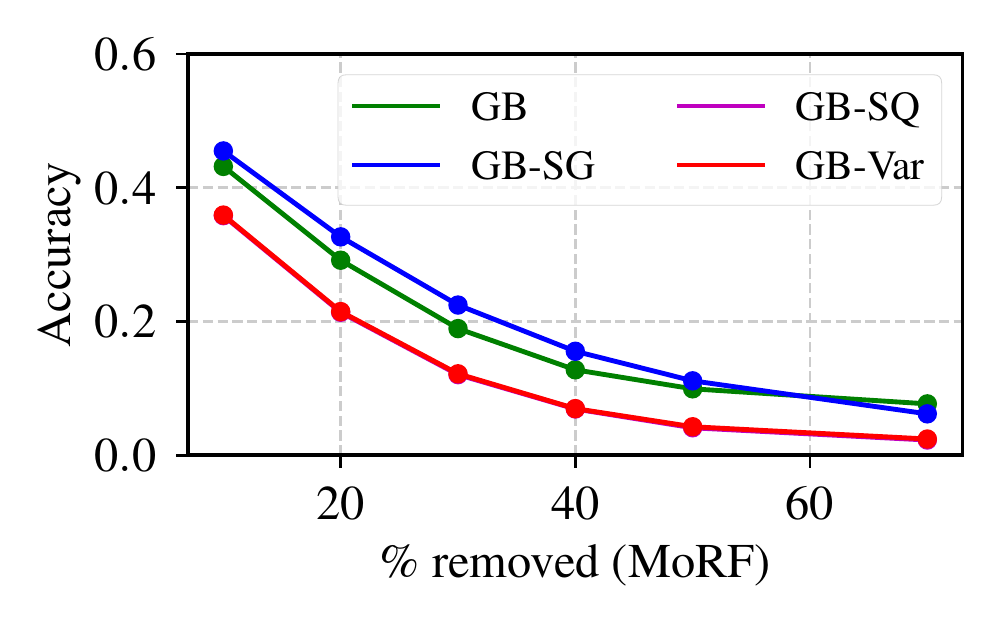}
    \caption{MoRF, No-Retrain}
\end{subfigure}
\hfill
\begin{subfigure}[h]{0.48\textwidth}
    \centering
    \includegraphics[width=\figuresuppw]{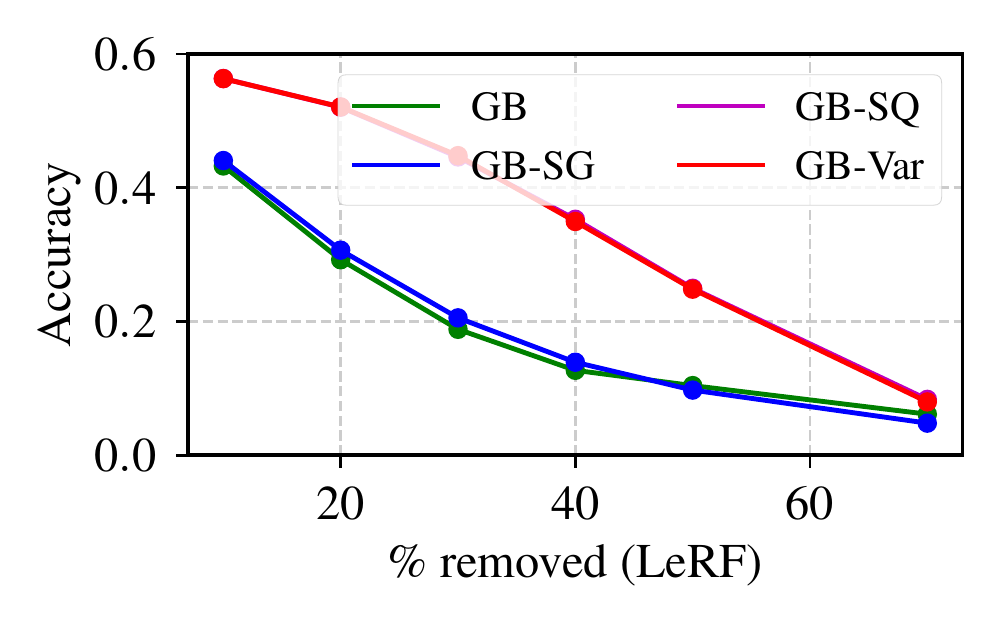}
    \caption{LeRF, No-Retrain}
\end{subfigure}

\caption{Consistency comparison using \textbf{Fixed Value} imputation on \textbf{GB}-based methods on Food-101.}
\label{suppfig:food-gb-fixed}
\end{figure}

\begin{figure}[]
\begin{subfigure}[h]{0.48\textwidth}
    \centering
    \includegraphics[width=\figuresuppw]{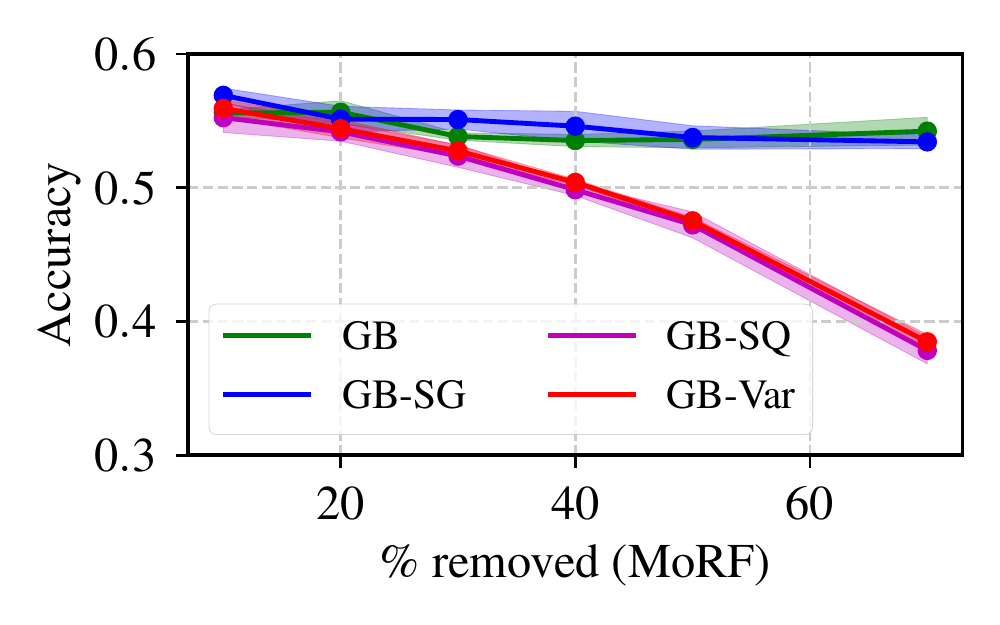}
    \caption{MoRF, Retrain}
\end{subfigure}
\hfill
\begin{subfigure}[h]{0.48\textwidth}
    \centering
    \includegraphics[width=\figuresuppw]{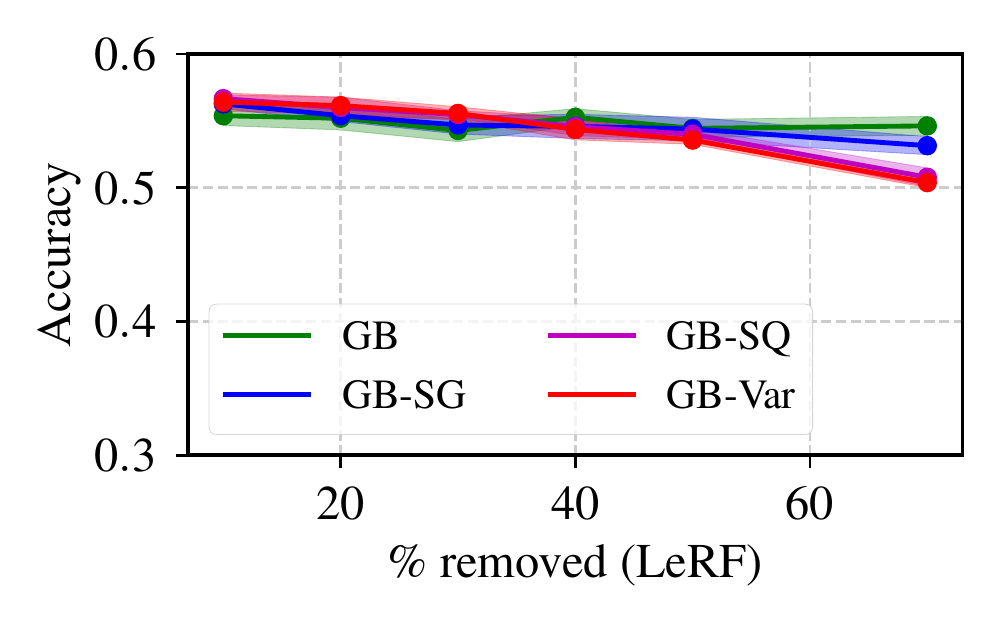}
    \caption{LeRF, Retrain}
\end{subfigure}
\\
\begin{subfigure}[h]{0.48\textwidth}
    \centering
    \includegraphics[width=\figuresuppw]{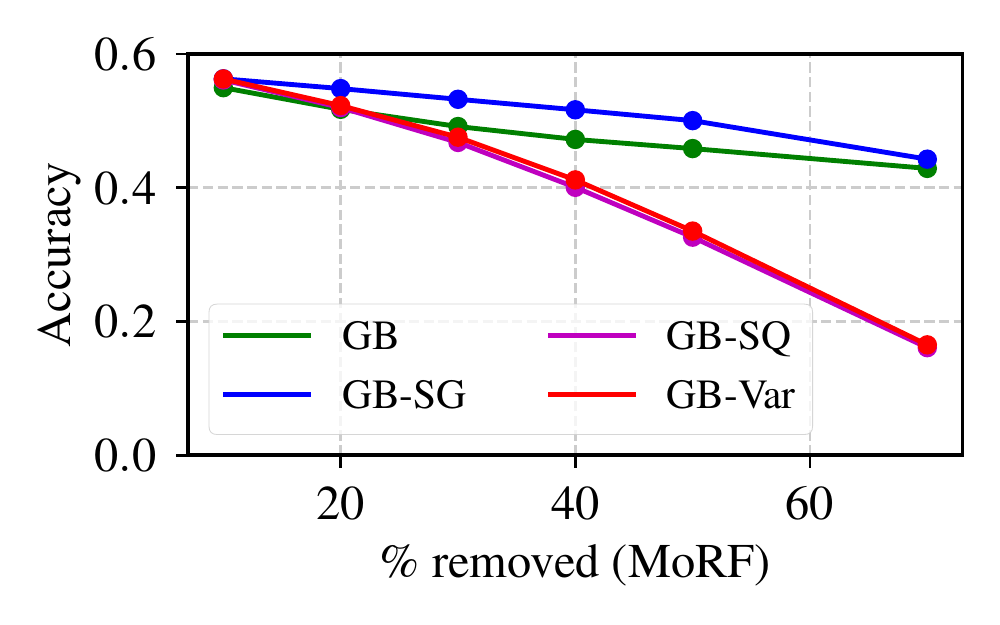}
    \caption{MoRF, No-Retrain}
\end{subfigure}
\hfill
\begin{subfigure}[h]{0.48\textwidth}
    \centering
    \includegraphics[width=\figuresuppw]{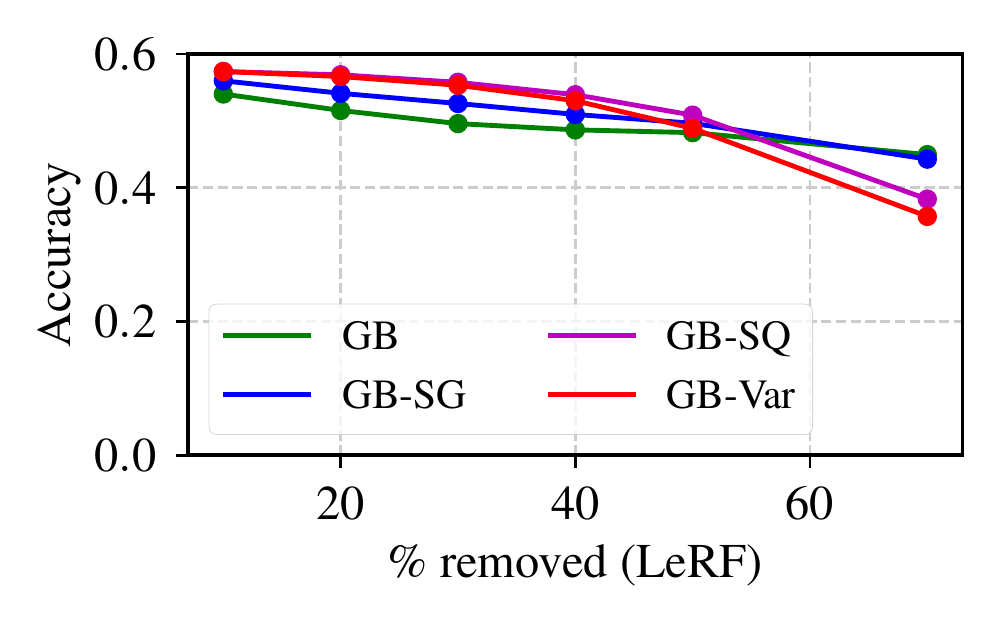}
    \caption{LeRF, No-Retrain}
\end{subfigure}

\caption{Consistency comparison using \textbf{Noisy Linear} imputation on \textbf{GB}-based methods on Food-101.}
\label{suppfig:food-gb-linear}
\end{figure}

\begin{figure}[]
\begin{subfigure}[h]{0.4\textwidth}
    \centering
    \includegraphics[width=\textwidth]{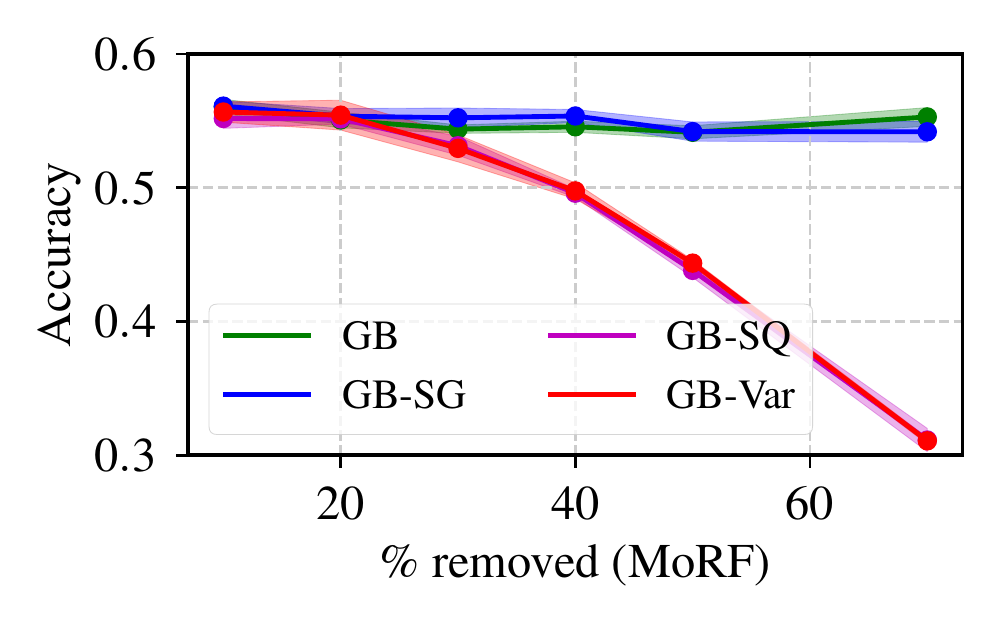}
    \caption{MoRF, Retrain}
\end{subfigure}
\hfill
\begin{subfigure}[h]{0.4\textwidth}
    \centering
    \includegraphics[width=\textwidth]{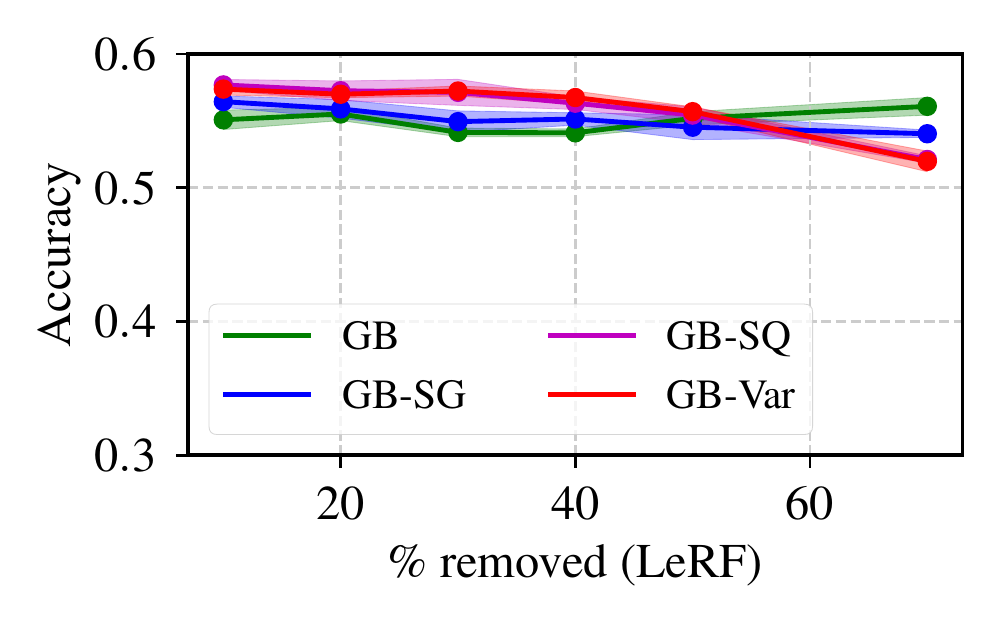}
    \caption{LeRF, Retrain}
\end{subfigure}
\\
\begin{subfigure}[h]{0.4\textwidth}
    \centering
    \includegraphics[width=\textwidth]{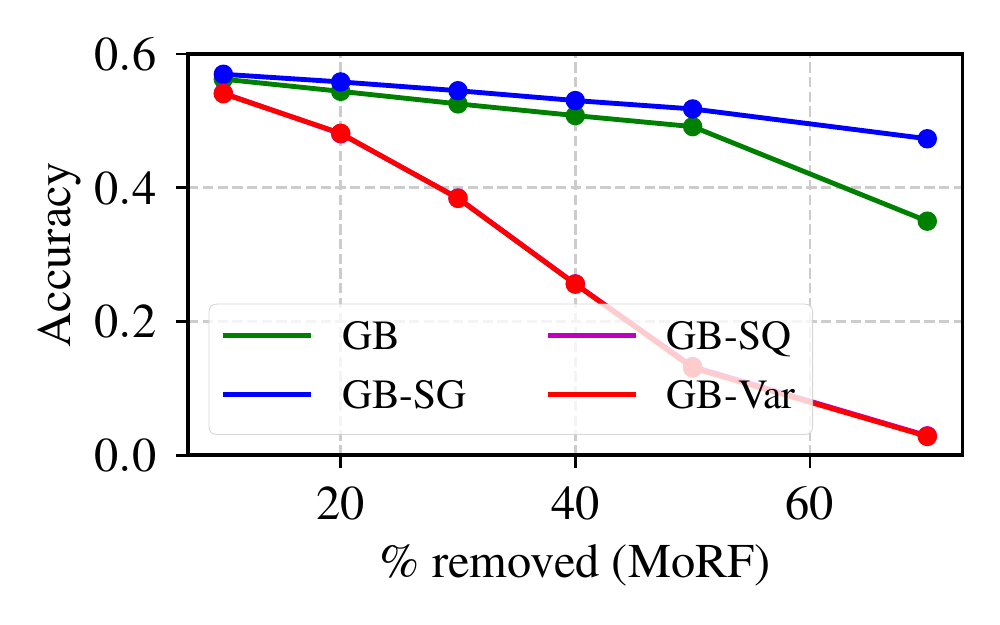}
    \caption{MoRF, No-Retrain}
\end{subfigure}
\hfill
\begin{subfigure}[h]{0.4\textwidth}
    \centering
    \includegraphics[width=\textwidth]{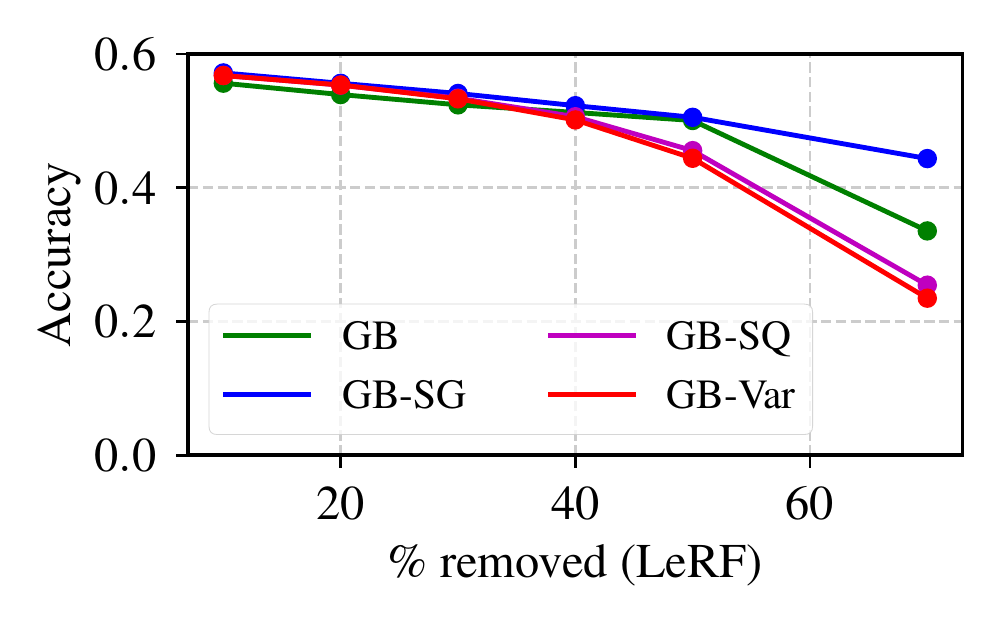}
    \caption{LeRF, No-Retrain}
\end{subfigure}

\caption{Consistency comparison using \textbf{GAN} imputation on \textbf{GB}-based methods on Food-101.}
\label{suppfig:food-gb-gan}
\end{figure}

Full qualitative results of using four variants in evaluation strategies (``Retrain"/``No-Retrain", MoRF/LeRF) for three different imputation operators (fixed value/Noisy Linear/GAN imputation) are listed from \cref{suppfig:food-ig-fixed} to \cref{suppfig:food-gb-gan}. 
\cref{suppfig:food-ig-linear} and \cref{suppfig:food-gb-linear} show the evaluation results when using our Noisy Linear Imputation for IG- and GB-family attribution methods, respectively. From results, we see that using our Noisy Linear Imputation, the consistency between the evaluation results using ``Retrain" and ``No-Retrain" are more consistent compared to using the fixed value imputation. Therefore, retraining can be safely skipped by using our Noisy Linear Imputation.


\end{document}


%

%

\onecolumn
\aistatstitle{Evaluating Feature Attribution -- An Information-Theoretic Perspective: Supplementary Material}
\section{ON THE CONNECTION BETWEEN MUTUAL INFORMATION AND ACCURACY}
\begin{figure}[b]
  \centering
  \includegraphics[width=0.3\textwidth]{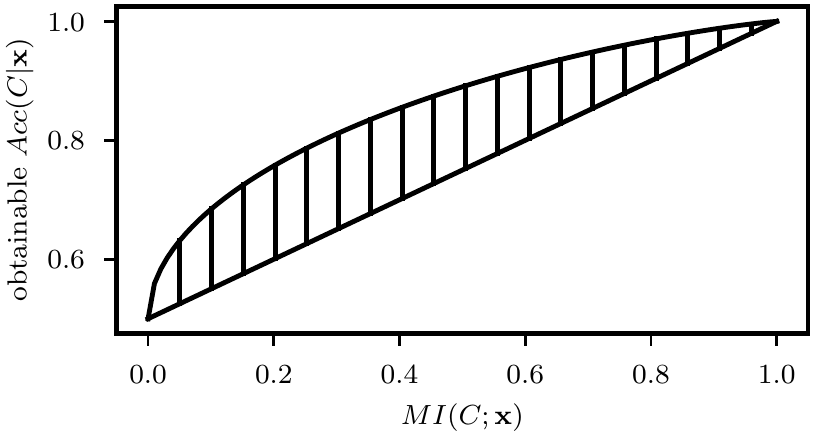}
  \caption{Relation between MI and obtainable optimal accuracy for the two-class problem with equal class priors. The knowledge of the MI $I(\bx;C)$ implies strong bounds for the obtainable accuracy. Figure adapted from \cite{Meyen2016masterThesis}. The lower bound is linear.\label{fig:accandmi}
  }
\end{figure}

It is well-known that the classification performance of an optimal classifier in the Bayesian sense (assigning the class with the highest posterior) is dependent on the MI between features and labels \citep{Hellman1970perror, Vergara2014reviewMIfeatureSelection, Meyen2016masterThesis}. Nevertheless, the relationship is not a function, but comes in form of upper and lower bounds of the obtainable accuracy. If, for example, the binary classification case with equal class priors $p(C=0)=p(C=1)=\frac{1}{2}$ is considered, the following bounds can be derived \citep{Hellman1970perror, Meyen2016masterThesis}:
\begin{equation}
\label{eqn:defmi}
\frac{I(\bx;C) +1}{2} \leq \acc(C|\bx) \leq H_2^{-1}(1-I(\bx;C)),    
\end{equation}
where $H_2^{-1}: [0,1] \rightarrow \left[\frac{1}{2}, 1\right]$ is the inverse of the binary entropy with support $\left[\frac{1}{2}, 1\right]$. For completeness, we restate the proof of this upper bound in \Cref{sec:miandaccproofs}.

For the simple two-class problem, the bounds are shown in Fig. \ref{fig:accandmi}. They impose strong limits on the optimal classification performance, if the mutual information $I(\bx;C)$ is known. For the retraining approaches, accuracy is the measure of interest, so we argue that their performance will also be highly determined by the amount of mutual information present between selected features and the class labels. This allows for an analysis of the approaches on an information-theoretic level, using mutual information. We also observe that the lower bound of the attainable accuracy is linear, which we will use for our geometric debiasing strategy in a later section.

\subsection{Reproduction of the proof of the relation between mutual and accuracy in the binary case}
\label{sec:miandaccproofs}
In this section, we reproduce the proofs for the upper and lower bounds of bayesian classifier accuracy given a certain amount of mutual information from the master's thesis by \cite{Meyen2016masterThesis} for completeness. The upper bound given there is tighter than the bounds present in the literature. \\
We consider the following setting ($C$, $\bx$ are random variables):
\begin{itemize}
    \item binary classification problem, $C\in \Omega_C =\{0,1\}$
    \item equal class priors $P(C=0)=\frac{1}{2}, P(C=1)=\frac{1}{2}$
    \item discrete features $\bx$ (which can be the product of multiple random variables)
    \item support set $\Omega_{x}=\supp{\left\{\bx\right\}}$ of \textit{countable} size
\end{itemize}  
We first prove the following Lemma:
\begin{lemma}
Let the assumptions stated above be true. Then, the mutual information is the weighted mean of a function of the conditional accuracies $\acc(C|s)$, where $s\in \Omega_{x}$:
\[
I\left(C; \bx\right) = \sum_{s\in \Omega_{S}} p(s)\left(1-H_2\left[\acc(C|s)\right]\right)
\]\label{lem:micondacc}
\end{lemma}
In this formulation, $p(s)$ is a shorthand for $P(\bx=s)$ and $H_2(p) := -p\log p - (1-p)\log(1-p)$ is the entropy for a binary random variable.\\
\textbf{Proof.}\\
\begin{align}
    I\left(C; \bx\right) &= H(C)-H(C|\bx)\\
    &= \sum_{c\in \Omega_C} p(c) \log \frac{1}{p(c)} - \sum_{s\in \Omega_S} p(s) \sum_{c\in \Omega_C} p(c|s) \log \frac{1}{p(c|s)}\\
    &= \sum_{s\in \Omega_x} p(s) \left[\sum_{c\in \Omega_C} p(c) \log \frac{1}{p(c)} -  \sum_{c\in \Omega_C} p(c|s) \log \frac{1}{p(c|s)}\right]\\
    &= \sum_{s\in \Omega_x} p(s) \left[H(C)-H(C|s)\right]
\end{align}
In our consideration, $\Omega_C =\{0,1\}$ and $P(C=0)=\frac{1}{2}, P(C=1)=\frac{1}{2}$, so $H(C)=1$. Additionally, the bayesian classifier rule yields
\begin{equation}
    acc(C|s)= \left\{\begin{array}{lr}
        P(C=0|s), & \text{for } P(C=1|s) \leq 0.5\\
        P(C=1|s), & \text{for } P(C=1|s) > 0.5
        \end{array}\right. 
\end{equation}
and
\begin{align}
H(C|s) &= -P(C=0|s)\log P(C=0|s) - P(C=1|s) \log P(C=1|s)\\
&= H_2(P(C=0|s)) = H_2(P(C=1|s))\\
&= H_2(acc(C|s)) 
\end{align}
Plugging in the results $H(C)=1$ and $H(C|s)=H_2(\acc(C|s))$, we obtain the proposed lemma.
~$\hfill\square$

For the derivation of upper and lower bounds, Jenssen's inequality is used. $1-H_2(\cdot)$ is a convex function and the $\left\{p(s)\right\}_{s\in\Omega_x}$ are convex multipliers, i.e., they are non-negative and sum up to one. Then,
\begin{align}
1-H_2\left(\acc(C|\bx) \right) & = 1-H_2\left(\sum_{s\in \Omega_{x}} p(s)\acc(C|s) \right)\\
& \leq \sum_{s\in \Omega_{x}} p(s)\left[1-H_2\left(\acc(C|s)\right)\right] = I(\bx;C)
\end{align}
We can restate this equation in terms of accuracy.
\begin{equation}
   H_2\left(\acc(C|\bx) \right) \geq 1-I(C;\bx) 
\end{equation}
Using that $H_2\left(\cdot\right)$ is decreasing monotonically on the interval $\left[\frac{1}{2},1\right]$, so its inverse $H^{-1}_2$ exists, and that $\acc(C|s) \geq 0.5$:
\begin{equation}
    \acc(C|\bx) \leq H_2^{-1}\left(1- I(C; \bx) \right).   
\end{equation}
The inequality sign is flipped again, due to the inverse being monotonically decreasing. Note that the bounds derived for the special case are much tighter than the general ones provided by \citet{Vergara2014reviewMIfeatureSelection} and \citet[Chapter 2.10]{Cover2006}, that are not of any use, because they are even less strict than the trivial bound $\acc(C|\bx) \leq 1$, for the simple case considered here.

For the lower bound, we refer the reader to \citet[eqn. 18]{Hellman1970perror}, where the term $I$ corresponds to $H(C|\bx)=H(C)-I(C;\bx)$ in our notation.
Rewriting the result from \citet{Hellman1970perror} in our notation, we obtain
\begin{equation}
    1-\acc(C|\bx) \leq \frac{H(C)-I(C;\bx)}{2}.  
\end{equation}

Using $H(C)=1$ and rearranging yields
\begin{equation}
    1-\acc(C|\bx) \leq \frac{1-I(C;\bx)}{2}
\end{equation}
and
\begin{equation}
    \acc(C|\bx) \geq \frac{I(C;\bx)+1}{2}.
\end{equation}
~$\hfill\square$

\section{ANALYSIS OF LeRF}
\subsection{Masking Bias}
\begin{figure}[b]
    \centering
    \begin{tikzpicture}[
    squarednode/.style={rectangle, draw=black,  thick, minimum size=4mm},
    ]
    
    \newcommand{\cellfc}[0]{darkgray}
    \newcommand{\belowdist}[0]{0.00cm}
    \definecolor{color0}{RGB}{55,162,219}
    \definecolor{color1}{RGB}{255,71,76}
    \definecolor{color2}{RGB}{178,51,120}
    \definecolor{color3}{RGB}{232,187,44}
    \definecolor{color4}{RGB}{25,161,108}
    \definecolor{color5}{RGB}{40,97,205}
    \definecolor{color6}{RGB}{240,158,7}
    \definecolor{color7}{RGB}{180,43,40}
    \definecolor{color8}{RGB}{156,158,114}
    
    \matrix (m1) at (0,0) [matrix of nodes,nodes={squarednode},column sep=-\pgflinewidth, row sep=-\pgflinewidth]{
        |[draw,fill=color0]| 0 & |[draw,fill=color1]| 1 & |[draw,fill=color2]| 2 \\
        |[draw,fill=color3]| 3 & |[draw,fill=color4]| 4 & |[draw,fill=color5]| 5 \\
        |[draw,fill=color6]| 6 & |[draw,fill=color7]| 7 & |[draw,fill=color8]| 8 \\
        };
    \node [above =\belowdist of m1]{$\bm{x}$ (input)};    
    \node (explan)[below =1.5 cm of m1]{\parbox[c]{1.9cm}{\raggedright explainer $\bm{e}$ for model $f$}}; 
    
    \matrix (m2) [below right = 0.3cm and 0.5cm of m1, matrix of nodes,  nodes={squarednode}, column sep=-\pgflinewidth, row sep=-\pgflinewidth]{
    |[draw,fill=black]| \textcolor{white}{0} & |[draw,fill=black]| \textcolor{white}{1}  & |[draw]| 2 \\
    |[draw]| 3 & |[draw,fill=black]| \textcolor{white}{4}  & |[draw,fill=black]| \textcolor{white}{5}  \\
    |[draw]| 6 & |[draw,fill=black]| \textcolor{white}{7}  & |[draw]| 8 \\
    };
    \node [below =\belowdist of m2]{$\mask$};
    
    \matrix (mxl) [right= 2.4cm of m1, matrix of nodes,  nodes={squarednode}, column sep=-\pgflinewidth, row sep=-\pgflinewidth]{
    |[draw,fill=color0]| 0\\
    |[draw,fill=color1]| 1\\
    |[draw,fill=color4]| 4\\
    |[draw,fill=color5]| 5\\
    |[draw,fill=color7]| 7\\
    };
    \node [below = \belowdist of mxl]{$\xh$};
    
    \matrix (mxlp) [right = 1.5cm of m2, matrix of nodes,  nodes={squarednode}, column sep=-\pgflinewidth, row sep=-\pgflinewidth]{
        |[draw,fill=color0]| 0 & |[draw,fill=color1]| 1 & |[draw,fill=\cellfc]| \textcolor{\cellfc}{2} \\
        |[draw,fill=\cellfc]| \textcolor{\cellfc}{3} & 
        |[draw,fill=color4]| 4 & |[draw,fill=color5]| 5 \\
        |[draw,fill=\cellfc]| \textcolor{\cellfc}{6} & |[draw,fill=color7]| 7 & |[draw,fill=\cellfc]| \textcolor{\cellfc}{8}  \\
    };
    \node [below =\belowdist of mxlp]{$\xhp$ (high importance)};
    
    \node at (4, -1.5) {\textcolor{gray}{$\scatterOP_h$}};
    \node at (2.6, -0.3) {\textcolor{gray}{$\mathcal{M}_h$}};
    
    \draw[->, dotted](explan.north) -- (m2.west);
    \draw[->](m1) -- (m2);
    \draw[->](m1) -- (mxl);
    \draw[->](m2) -- (mxl);
    \draw[->](m2) -- (mxlp);
    \draw[->](mxl) -- (mxlp);
    \end{tikzpicture}
    \caption{Analogous analytical model of feature removal in the opposite order (LeRF): The input image $\bm{x}$ is explained by an explanation method that returns a mask $\mask$ indicating important pixels. The remaining, highly important pixels can be extracted from the image using the masking operator $\mathcal{M}_h$ and transformed to a modified variant of the input $\xhp$ via the imputation operator $\scatterOP_h$.\label{fig:functionalsetuplerf}}
\end{figure}

In this section, we analyze the masking bias for the case of the Least Relevant First (LeRF) ordering. We first provide a definition for the operators involved as we did for the Most Relevant First (MoRF) case. In the LeRF setting, the $k$ least important important features per instance are removed. We model the explanation as a choice of features via a binary mask $\mask = \bm{e}\left(f, \bm{x}\right) \in \left\{ 0,1 \right\}^d$, with the corresponding value set to one, if the corresponding feature is among the top-$k$, and to zero otherwise. Furthermore, suppose $\mathcal{M}_{h}: \left\{ 0,1 \right\}^d \times  \mathbb{R}^d \rightarrow \mathbb{R}^{k}$ to be the selection operator for the \textbf{\underline{h}}ighly important dimensions indicated in the mask and $\bm{x}_{h}=\mathcal{M}_{h}\left(\bm{M}, \bm{x}\right)$ to be a vector containing only the remaining, highly important features as shown in \cref{fig:functionalsetuplerf}. We suppose that the features preserve their internal order in $\xh$, i.e., features are ordered ascendingly by their original input indices. 

The LeRF approach with retraining (also called ``Keep and Retrain'', KAR, by \citet{Hooker2019ROAR}) measures the accuracy of a newly trained classifier $f'$ on modified samples $\xhp \coloneqq \scatterOP_{h}\left(\bm{M},\bm{x_h}\right)$, where $\scatterOP_{h}: \left\{ 0,1 \right\}^d\times\mathbb{R}^{k} \rightarrow \mathbb{R}^{d}$ is an imputation operator that redistributes all inputs in the vector $\xh$ to their original positions and sets the remainder to some filling value. This means only the top-$k$ features are kept. For a better evaluation result, the accuracy should increase quickly with increasing $k$, indicating the most influential features are present. Accuracy should not increase much for the high values  of $k$ because removing the low importance features should not have a large effect. Overall, higher accuracies indicate better explanations in the LeRF setting.

For the LeRF benchmark, the quantity of interest in our analysis will be $I(\xhp; C)$, the class information contained in the filled-in version of the selected high important features. We want to maximize $I(\xhp; C)$ to obtain a good score,
\[
\uparrow I(\xhp; C)~~\Rightarrow~~\uparrow \text{LeRF benchmark}.
\]

As before, we can apply the following, general identity:
\begin{equation}
\label{eqn:maskgeneral}
   \underbrace{I(\xhp; C)}_\text{Evaluation Goal} = \underbrace{I(C;\xhp|\mask)}_\text{Feature Info.} +  \underbrace{I(C;\mask)}_\text{Mask Info.} - \underbrace{I(C;\mask|\xhp)}_\text{Mitigator}.
\end{equation}
The interpretation of the terms is analogous to that in our main paper.
\paragraph{Class-Leaking Explanation Map} For the case of the class-leaking map, we again require the imputation operator to be invertible:
\begin{example}
\label{ass:condinvertibleinfilling}
\textit{Invertible Imputation. Let $\scatterOP_h: \left\{ 0,1 \right\}^d \times  \mathbb{R}^{k} \rightarrow \mathbb{R}^{d}$ be the imputation operator that takes the highly important features as an input. We suppose that there are inverse functions $\scatterOP_{h,M}^{-1}$ and $\scatterOP_{h,x}^{-1}$, such that}
\[
\xhp = \scatterOP_h\left(\bm{M}, \xh\right) \Leftrightarrow \mask = \scatterOP_{h,M}^{-1}(\xhp) \wedge \bm{x}_h = \scatterOP_{h,x}^{-1}(\bm{x}_h^\prime).
\]
\end{example}

If, for instance, the pixels removed are set to some reserved value indicating their absence, the infilling operator is invertible. In this case, also the Mitigator $\mic{C}{\mask}{\xhp} = 0$ (see \Cref{sec:invimputationdetails} for details) 
The ``Feature Info" term is constrained to be positive. Thus, the Mask Information has a non-negligible impact on the Evaluation Goal, because a higher Mask term will always increase it.

We can create a another example of a spurious explanation map that shows how evaluation scores are influenced even worse for LeRF: Suppose an explanation map that starts masking out pixels at the top for class zero and at the bottom for class one. Thus, a retrained model will be able to infer the category just from the shape of the masked pixels and obtain the best possible accuracy and thus score in the LeRF setting. However, it does not provide a reasonable attribution for the importance of the features.

\subsection{Geometric Bias}
In this section, we give details on the analysis of the geometric biases in the LeRF benchmark. As before, we  consider the mask to have a negligible influence on the problem and
\begin{equation}
    \mi{C}{\xhp} \approx \mim{C}{\xhp} = \mim{C}{\xh}.
\end{equation}
For the LeRF Benchmark, we introduce the opposite condition that reflects the share of information uniquely contained in the low-importance features:
\begin{condition} Ratio of class information is lower in low-importance features. Suppose
\[
\frac{\micm{\xl}{C}{\xh}}{\mim{\xl}{C}} \leq \frac{H\left(\xl\middle|\xh\right)_M}{H\left(\xl\right)_M}.
\]
\end{condition}
Usually, image features exhibit high dependency. Therefore, through high correlations some parts of the information $H(\bm{x}_l)$ will be given away through observation of $\bm{x}_h$. The right hand side $\betl \coloneqq \frac{H\left(\bm{x}_l\middle|\bm{x_h}\right)_M}{H\left(\bm{x}_h\right)_M}< 1$  provides the share of information that uniquely contained in $\bm{x}_l$ and not in  $\bm{x}_h$. Now we argue, that the share of unique class information left in $\bm{x}_l$ after observation of $\bm{x}_h$ is at lower or equal to the share of the total information. This is reasonable, since the class information should be hidden in the features deemed highly important to a large extent.

Using the condition, we can perform the following derivation:
\begin{align}
\begin{split}
    \mim{\xh}{C} &= \mim{\xh,\xl}{C} - \micm{\xl}{C}{\xh}\\
    &\leq \mim{\xl,\xh}{C} - \frac{\entcm{\xl}{\xh}}{\entm{\xl}}\mim{\xl}{C}\\
    &\coloneqq \mim{\bm{x}}{C} - \beta_l \mim{\xl}{C}.
\end{split}
\end{align}
The identity $\mim{\bm{x}}{C}= \mim{\xl,\xh}{C}$ holds because, given $\mask$, there is an invertible transform between $(\xh,\xl)$ and $\bm{x}$, so equality holds due to the data processing inequality.

The first term gives the total class-information in the features given the mask that provides their order. Having treated the effect of the mask in the previous section, we consider $\mim{\bm{x}}{C}$ to be a constant given by the data set and explanation method. As expected, we subtract some information, that is contained in the removed, low-ranked features $\xl$. Its effect is weighted by the term $\betl$, that gives the share of new information left in $\xh$ after observing $\xl$. Intuitively, if the share is $1$, the high importance features are independent of of $\xl$ and the information is irreversibly lost. However, if there is a strong dependency which is common in images, $\beta \approx 0$  and even if the $\mim{\xl}{C}$ is high, there is no effect, because the information is redundant and can still be reproduced from the remaining features. 

We conclude that for a good score in the LeRF benchmark, high dependence between $\xl,\xh$ is desirable.

\section{DETAILED DERIVATIONS}
\subsection{Invertible Imputation}
\label{sec:invimputationdetails}
We again present the proof that the Mitigator equals 0 when using invertible infilling in Sec.\ 4.1. Let $\scatterOP_{l,M}^{-1}$ be the perfectly invertible, mask-revealing operation function.
\begin{align}
\begin{split}
         I(C;M|\bm{x}_l^\prime) &=  H(M|\bm{x}_l^\prime) - H(M|\bm{x}_l^\prime,C) \\
          &=H(\scatterOP_{l,M}^{-1}(\bm{x}_l^\prime)|\bm{x}_l^\prime) - H(\scatterOP_{l,M}^{-1}(\bm{x}_l^\prime)|\bm{x}_l^\prime, C) \\
          &= 0-0 \\ 
          &= 0
\end{split}
\end{align}
Because the output quantity is entirely determined by the condition, the uncertainty about the outcome and thus the entropy is $H=0$.
This proof also holds when taking the most important features as an input and changing $\xlp$ to $\xhp$ and $\scatterOP_{l,M}^{-1}(\cdot)$ to $\scatterOP_{h,M}^{-1}(\cdot)$, respectively.

\section{IMPLEMENTATION DETAILS}
\subsection{Linear Imputation}
In this paragraph, we provide details on the linear imputation model used. We use a vectorized implementation to set up the sparse system and use the $\texttt{scipy.sparse}$ package for our implementation. We set up one equation per pixel and channel. This can result in a large number of equations, which fortunately only exhibit dependencies to few neighboring pixels. In practice, we observe a run time which is almost linear in the number of pixels masked out. At the boundaries we use the same weights as for pixels that lie in the interior of the image. However, we take the weight of the unknown value to be the sum of the existing neighbor's weights. In an example, we would have
\begin{equation}
    \frac{5}{12}\bx_{0,0}=\frac{1}{6}\bx_{1,0} + \frac{1}{6}\bx_{0,1} + \frac{1}{12}\bx_{1,1}.
\end{equation}
We add gaussian noise of $0.01\times \text{img\_range}$ (where range is the difference between the highest pixel values and the lowest) to the solution of the equation system.
\subsection{Geometric Debiasing}
\begin{figure} 
\captionsetup[subfigure]{justification=centering}
    \centering
 \begin{subfigure}[b]{0.58\textwidth}
    \centering 
    \includegraphics[width=\textwidth]{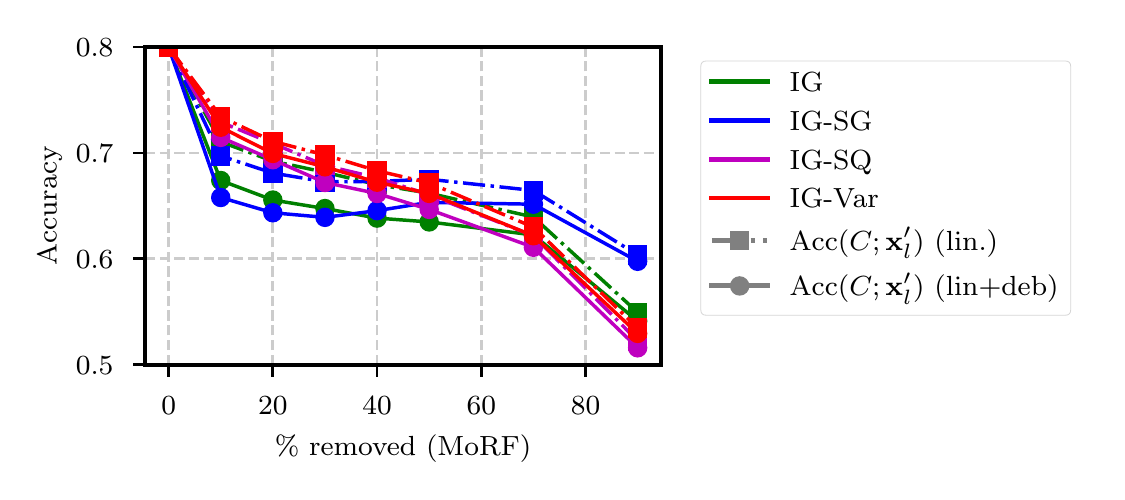}
    \label{fig:igleak}
    \caption{Debiased (solid) vs. linearly imputed (dotted) curves for IG methods}
     \end{subfigure}
    \hfill
  \begin{subfigure}[b]{0.35\textwidth}
  \centering
     \includegraphics[width=\textwidth]{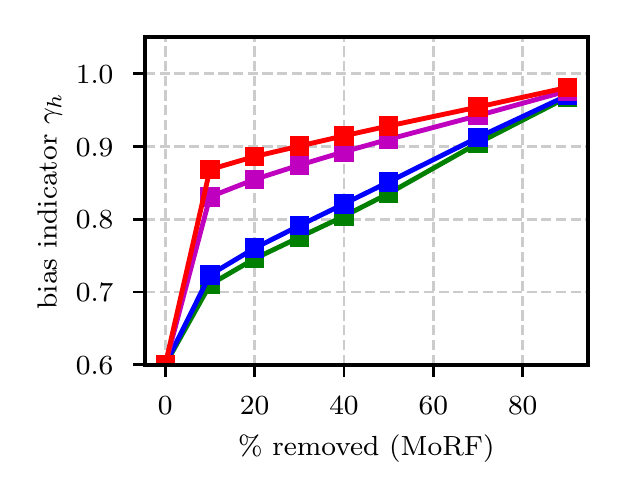}
     \caption{Computed bias indicator $\gamh$ for the IG methods}
    \label{fig:gbleak}
  \end{subfigure}
  
   \begin{subfigure}[b]{0.58\textwidth}
    \centering 
    \includegraphics[width=\textwidth]{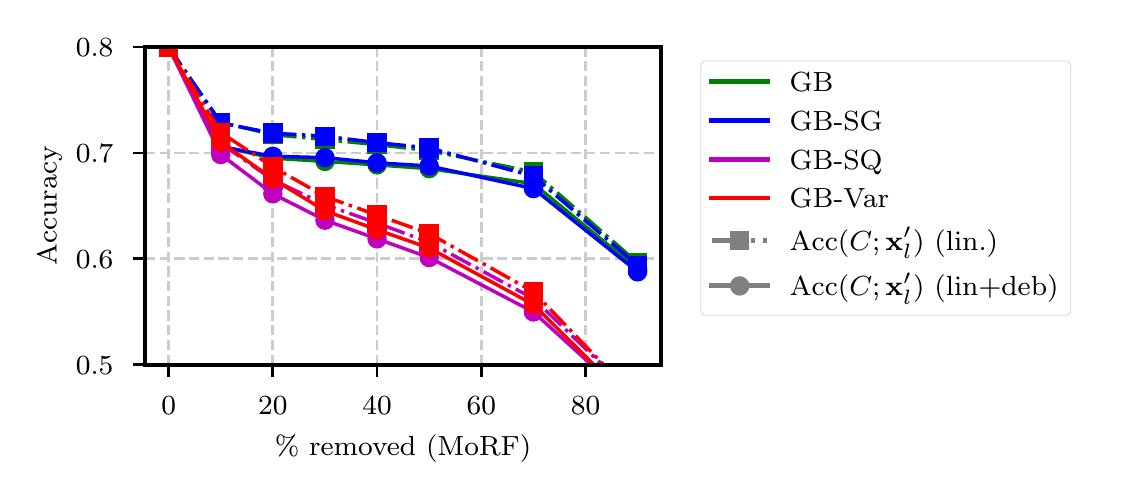}
    \label{fig:igleakinfill}
    \caption{Debiased (solid) vs. linearly imputed (dotted) curves for GB methods}
     \end{subfigure}
    \hfill
  \begin{subfigure}[b]{0.35\textwidth}
  \centering
     \includegraphics[width=\textwidth]{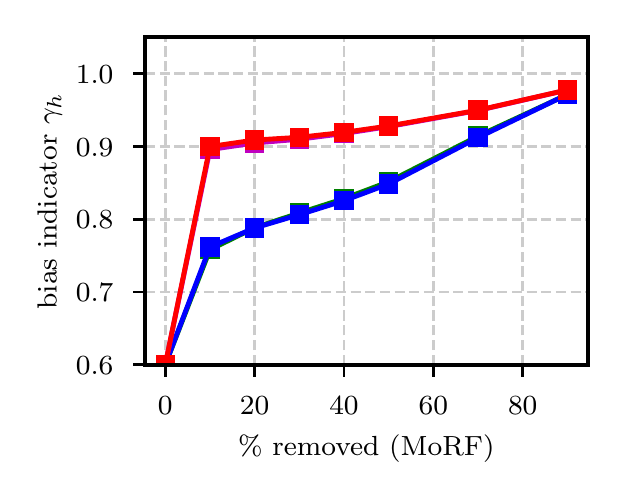}
    \label{fig:gbleakinfill}
      \caption{Computed bias indicator $\gamh$ for the GB methods}
  \end{subfigure}
  \caption{Debiasing on CIFAR-10 (MoRF, retraining setting shown). When the bias indicator $\gamh$ is close to $1$, no compensation is required. However, if it is lower, the debiasing model scales the drop from the baseline accuracy which results in an adapted curve. Debiasing results in particularly high changes for the IG methods.}
  \label{fig:debiascifar}
\end{figure}

Our simple geometric debiasing model can be described by the following equation for the MoRF setting:
\begin{equation}
\text{Acc}^\prime\coloneqq
\acc\left(\bm{x};C\right)- \frac{\acc\left(\bm{x};C\right) - \acc\left(\xlp;C\right)}{\gamma_h}.
\end{equation}
Conversely, in the LeRF setting we have
\begin{equation}
\text{Acc}^\prime\coloneqq
\acc\left(\bm{x};C\right)- \frac{\acc\left(\bm{x};C\right) - \acc\left(\xhp;C\right)}{\gamma_l}.
\end{equation}
We call $\text{Acc}^\prime$ the debiased accuracy. $\acc\left(\bm{x};C\right)$ is the accuracy of the network using the full set of images. In our experiment on CIFAR-10, $\acc\left(\bm{x};C\right) = 0.8$ (note that to not distort our results, we use no pretraining and no data augmentation of any kind, which leads to models with higher accuracy). The debiasing models leverage the accuracy drop by the inverse of the bias indicator, which was the result of our derivation. In \Cref{fig:debiascifar}, we show the debiased curves for the attribution methods that we used on the CIFAR-10 data set.

\section{EXPERIMENTAL DETAILS}
\subsection{Toy Experiment}
\begin{figure} 
 \begin{subfigure}[b]{0.2\textwidth}
    \centering 
    \scalebox{0.55}{\clipbox{0.0cm 0.1cm 0.4cm 0.1cm}{\import{figures/toy}{ToyOrderings.pgf}}}
    \label{fig:toyorderings}
     \end{subfigure}
    \hfill
  \begin{subfigure}[b]{.3\textwidth}
  \centering
    \scalebox{0.55}{\includegraphics[]{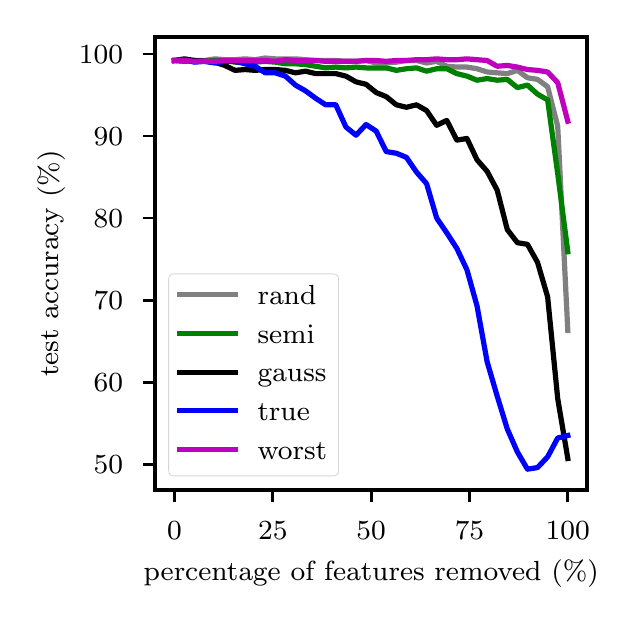}}
    \label{fig:toy_roar}
  \end{subfigure}
   \begin{subfigure}[b]{.3\textwidth}
    \scalebox{0.55}{\includegraphics[]{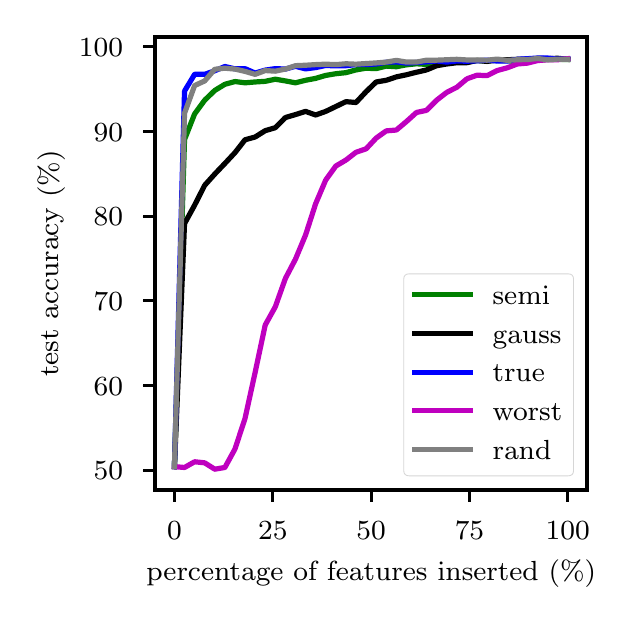}}
     \label{fig:toy_kar}
     \end{subfigure}
  \caption{\textbf{Left}: Artificially crafted, fixed feature importance scores for the synthetic data set. \textbf{Middle}: Results of retraining benchmarks on toy experiment using MoRF. \textbf{Right}: Results of retraining benchmarks on toy experiment using LeRF. While, after the ``true'' ordering, ``gauss'' performs best in MoRF, it is ranked directly after ``worst'' for the LeRF setting. The order of the attribution strategies is flipped.}
  \label{fig:debiastoy}
\end{figure}

In this section, we introduce the toy experiment that we conducted to investigate the geometric bias in additional details.
As outlined in the main paper, we sample 2D gray-scale images from a Gaussian Process (GP) distribution that introduces spatial correlation. We use a mixture model with two classes and class-conditional mean vectors $\mu_0, \mu_1$ but keep the same covariance matrix $\bm{\Sigma}$ over both classes (squared exponential kernel on distance matrix). Our data distribution is given by $p(\bm{x}) = \mathcal{N}\left(\bm{x}; \bm{\mu}_0, \bm{\Sigma} \right)p(C=0) + \mathcal{N}\left(\bm{x}; \bm{\mu}_1, \bm{\Sigma} \right)p(C=1)$. We use a squared exponential kernel with a kernel width of 20\,\% of the total image width. We use a resolution of 28$\times$28 pixels which is the same as the MNIST data set. The mean vectors are 2d-sinusoidal function. Please refer to our code for the full generation routine. Samples are shown in the main paper.
In addition to the three feature attribution methods in the paper, we consider the ``true'' ordering that corresponds to the true importance and the ``worst'' that provides the inverted feature order. They are shown again on the left of \cref{fig:debiastoy}. We expect the three hand crafted feature maps in the first row (``rand'', ``semi'', and ``gauss'') to be almost equally uninformative. 

\begin{figure} 
 \begin{subfigure}[b]{0.5\textwidth}
    \centering 
    \includegraphics[width=0.9\textwidth]{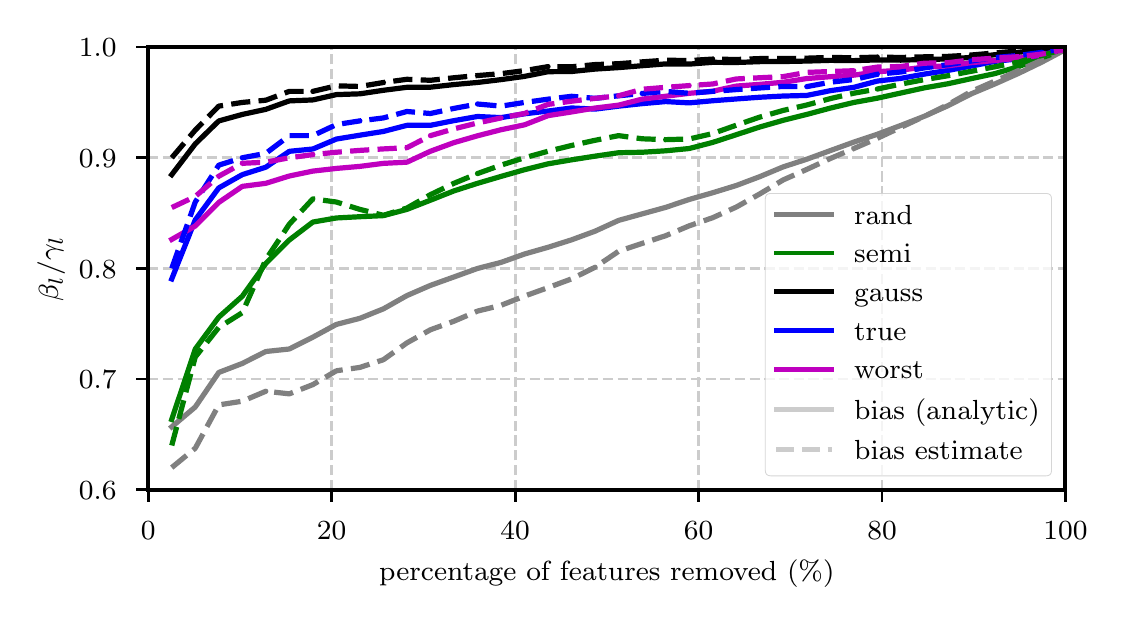}
    \label{fig:roarbiastoy}
     \end{subfigure}
    \hfill
  \begin{subfigure}[b]{0.5\textwidth}
  \centering
     \includegraphics[width=0.9\textwidth]{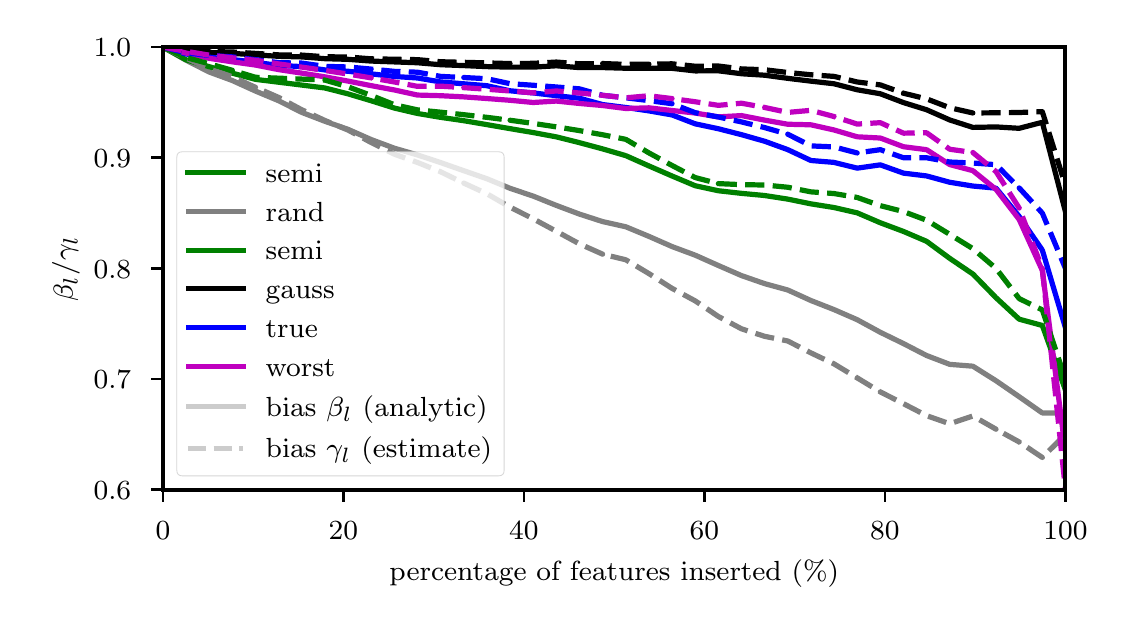}
    \label{fig:karbiastoy}
  \end{subfigure}
  \caption{\textbf{Left}: Bias indicators $\gamh$ and actual bias terms $\betah$ computed for the orderings in the toy experiment in the MoRF setting. \textbf{Right}: Bias indicators $\gamh$ and actual bias terms $\betah$ computed for the orderings in the toy experiment for the LeRF scheme. We observe that the bias indicator closely follows the analytical solution and that the indicator values show different behavior, in particular, when only few pixels are missing.}
  \label{fig:toybiasind}
\end{figure}

The results of MoRF and LeRF evaluation are given the center and the right of \cref{fig:debiastoy}.
Each result is the mean of accuracies obtained by $15$ independently trained logistic regression classifiers provided by the \texttt{scikit-learn} package. Additionally, we computed the bias terms and the indicator which we show in our main paper. Recall that for LeRF, high accuracy is desirable, whereas MoRF it should be low. Both strategies succeed in identifying the best and the worst-case, though only by a tiny margin. Nevertheless, the ranking of three maps in between (``rand'', ``semi'', ``gauss'') is reversed for LeRF and MoRF. This is possibly due to their very different bias terms, that we computed and show in \cref{fig:toybiasind}. This figure also highlights the correspondence of our indicators to the actual value of the bias term, which can be analytically computed for this experiment via the covariance matrix of the Gaussian Process that we used. 

\subsection{CIFAR-10 Experiments}
\begin{figure*}[t]
\centering
   \includegraphics[width=1\linewidth]{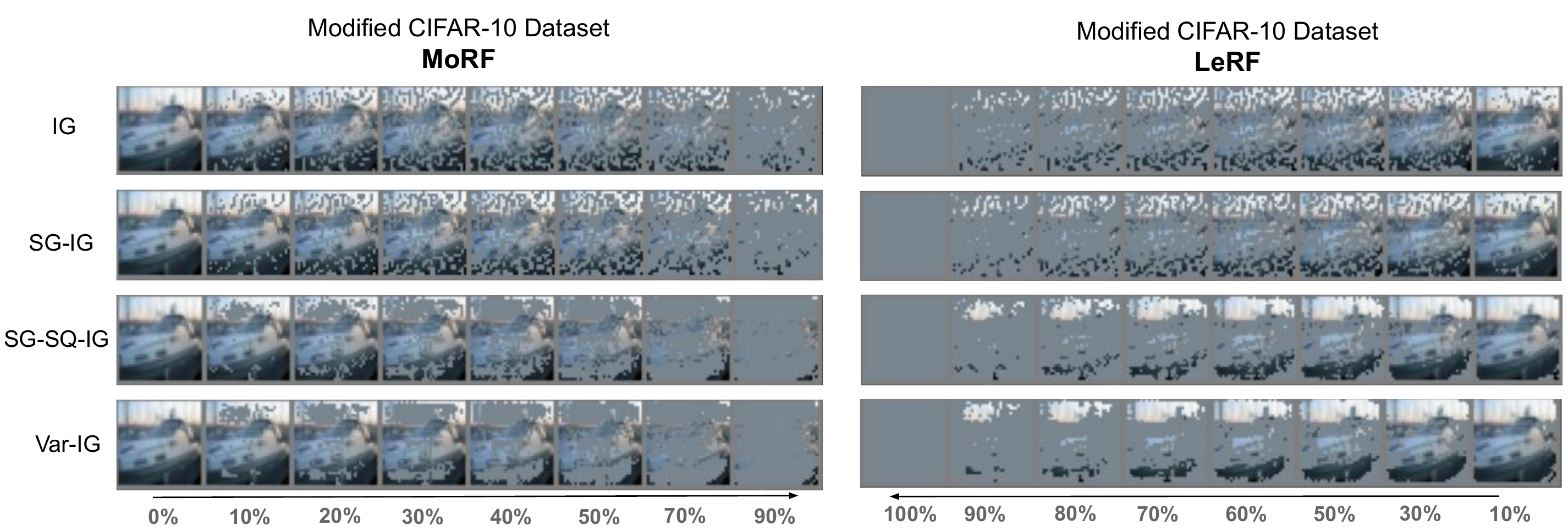}
    \caption{Illustration of modified data set in MoRF/LeRF and fixed value imputation settings. \textbf{Left}: Modifications in the MoRF framework. \textbf{Right}: Modifications in the LeRF framework. \textbf{Top to Bottom}: Modifications using Integrated Gradient (IG)\ \citep{sundararajan2017axiomatic}  and three ensemble variants of IG: SmoothGrad (SG-IG)\ \citep{smilkov2017smoothgrad}, SmoothGrad$^{2}$ (SG-SQ-IG)\ \citep{Hooker2019ROAR}, and VarGrad (Var-IG) \ \citep{adebayo2018sanity}. The percentage of pixels that are removed or kept is given at the bottom.}
    \label{fig:cifar}
    \vspace{-15pt}
\end{figure*}

We train a vanilla ResNet-18 \citet{he2016deep} on CIFAR-10 and compute different explanations using the trained model. The model is trained with the initial learning rate of $0.01$ and the SGD optimizer \citet{sutskever2013importance}. We decrease the learning rate by factor $0.1$ after $25$ and train the model for $40$ epochs on one (Nvidia RTX 3090) GPU. The trained model achieves a test set accuracy of 84.5\,\% (comparable to the model in \citet{tomsett2020sanity}). For attributions, we use the same settings as in \citep{Hooker2019ROAR}: As base explanations we implement Integrated Gradient (IG)\ \citep{sundararajan2017axiomatic} and Guided Backprop (GB)\ \citep{springenberg2014striving}. Additionally, we use three ensembling strategies for each: SmoothGrad (SG)\ \citep{smilkov2017smoothgrad}, SmoothGrad$^{2}$ (SG-SQ)\ \citep{Hooker2019ROAR} and VarGrad (Var) \ \citep{adebayo2018sanity}. For each explanation method, we modify the data set using the fraction of pixels $\eta = [0,0.1,0.2,0.3,0.4,0.5,0.7,0.9]$. \Cref{fig:cifar} illustrates the modified images by using four different explanations in the GB-family within MoRF and LeRF orders (fixed mean value imputation is used). 

We use \textbf{$N=5$ runs} and report averaged results for all CIFAR-10 experiments in our paper and indicate the standard errors (which are very small) as an area behind our plots. In \cref{tab:dev ig-sg} and \cref{tab:dev gb-sg}, we show the mean accuracy and its standard deviation at each the fraction of pixels $\eta$ for IG-SG and GB-SG explanations. For other explanations we used, the standard deviation at each $\eta$ in the magnitude of below one percent as well.

\begin{table}[h]
\centering
\scalebox{0.8}
{\begin{tabular}{|c|c|c|c|c|c|c|c|c|}
\cline{3-9}
\cline{3-9} \multicolumn{2}{c|}{} & 10  & 20  & 30 & 40 & 50 &  70    & 90   \\ 
\hline
\multirow{2}{*}{\begin{tabular}[c]{c@{}c@{}} Retrain\\ MoRF\end{tabular}} & fix &  \wstdmy{74.94}{0.57}\ & \wstdmy{75.42}{0.45}   & \wstdmy{75.62}{0.24} & \wstdmy{75.16}{0.50} & \wstdmy{74.95}{0.45} & \wstdmy{73.73}{0.48} & \wstdmy{65.18}{0.85} \\\cline{3-9}
&lin$^{*}$ & \wstdmy{69.72}{0.49}     & \wstdmy{68.10}{0.34}  & \wstdmy{67.28}{0.34} & \wstdmy{67.32}{0.22} & \wstdmy{67.52}{0.22}& \wstdmy{66.46}{0.54} & \wstdmy{60.37}{0.51}\\\hline
\multirow{2}{*}{\begin{tabular}[c]{c@{}c@{}} Non-Retrain\\ MoRF\end{tabular}} & fix &  \wstdmy{44.06}{0.04} & \wstdmy{29.81}{0.03}  & \wstdmy{21.99}{0.03} & \wstdmy{17.35}{0.02} & \wstdmy{14.67}{0.01}& \wstdmy{11.50}{0.04} & \wstdmy{10.90}{0.03}\\\cline{3-9}
&lin$^{*}$ & \wstdmy{67.66}{0.02}     &\wstdmy{59.94}{0.03}  & \wstdmy{54.05}{0.05} & \wstdmy{49.46}{0.04} & \wstdmy{45.63}{0.06} & \wstdmy{36.87}{0.05} & \wstdmy{24.55}{0.04}\\\hline
\multirow{2}{*}{\begin{tabular}[c]{c@{}c@{}} Retrain\\ LeRF\end{tabular}} & fix &  \wstdmy{77.52}{0.26} & \wstdmy{80.14}{0.11}   & \wstdmy{81.08}{0.14} & \wstdmy{81.40}{0.15} & \wstdmy{81.50}{0.07} & \wstdmy{81.85}{0.12}  & \wstdmy{82.01}{0.20}\\\cline{3-9}
&lin$^{*}$ & \wstdmy{72.73}{0.20}    & \wstdmy{77.52}{0.23}  &\wstdmy{79.29}{0.41} & \wstdmy{80.57}{0.31} & \wstdmy{80.97}{0.20} & \wstdmy{81.64}{0.26} &  \wstdmy{81.41}{0.13} \\\hline

\multirow{2}{*}{\begin{tabular}[c]{c@{}c@{}} Non-Retrain\\ LeRF\end{tabular}} & fix &  \wstdmy{46.76}{0.04} & \wstdmy{51.84}{0.04}   & \wstdmy{53.80}{0.05} & \wstdmy{55.39}{0.06} & \wstdmy{57.59}{0.05} & \wstdmy{64.02}{0.05} & \wstdmy{74.33}{0.03} \\\cline{3-9}
&lin$^{*}$ & \wstdmy{56.48}{0.03}     &\wstdmy{72.57}{0.03} & \wstdmy{77.96}{0.02} & \wstdmy{79.54}{0.01} & \wstdmy{81.02}{0.04} & \wstdmy{81.82}{0.02} &  \wstdmy{82.24}{0.01}\\ \hline
\end{tabular}}
\caption{Mean accuracy at each $\eta$ by using IG-SG in all methods with standard deviations of five individual runs. For LeRF, the accuracy is at (1-$\eta$).}
\label{tab:dev ig-sg}
\end{table}

\begin{table}[h]
\centering
\scalebox{0.8}
{\begin{tabular}{|c|c|c|c|c|c|c|c|c|}
\cline{3-9}
\cline{3-9} \multicolumn{2}{c|}{} & 10  & 20  & 30 & 40 & 50 &  70    & 90   \\ 
\hline
\multirow{2}{*}{\begin{tabular}[c]{c@{}c@{}} Retrain\\ MoRF\end{tabular}} & fix &  \wstdmy{76.30}{0.43}\ & \wstdmy{75.60}{0.27}   & \wstdmy{74.89}{0.29} & \wstdmy{74.27}{0.29} & \wstdmy{73.37}{0.28} & \wstdmy{72.15}{0.09} & \wstdmy{67.99}{0.24} \\\cline{3-9}
&lin$^{*}$ & \wstdmy{72.83}{0.37}     & \wstdmy{71.87}{0.41}  & \wstdmy{71.58}{0.19} & \wstdmy{70.98}{0.15} & \wstdmy{70.47}{0.20}& \wstdmy{67.81}{0.45} & \wstdmy{59.38}{0.46}\\\hline

\multirow{2}{*}{\begin{tabular}[c]{c@{}c@{}} Non-Retrain\\ MoRF\end{tabular}} & fix &  \wstdmy{73.01}{0.04} & \wstdmy{66.73}{0.04}  & \wstdmy{58.69}{0.09} & \wstdmy{52.47}{0.10} & \wstdmy{48.52}{0.05}& \wstdmy{48.71}{0.04} & \wstdmy{44.39}{0.01}\\\cline{3-9}
&lin$^{*}$ & \wstdmy{74.54}{0.03}     &\wstdmy{71.24}{0.04}  & \wstdmy{68.83}{0.02} & \wstdmy{67.21}{0.01} & \wstdmy{64.80}{0.04} & \wstdmy{57.60}{0.08} & \wstdmy{32.98}{0.03}\\\hline

\multirow{2}{*}{\begin{tabular}[c]{c@{}c@{}} Retrain\\ LeRF\end{tabular}} & fix &  \wstdmy{66.97}{0.52} & \wstdmy{70.45}{0.30}   & \wstdmy{71.44}{0.25} & \wstdmy{72.15}{0.15} & \wstdmy{72.72}{0.11} & \wstdmy{73.91}{0.19}  & \wstdmy{75.24}{0.25}\\\cline{3-9}
&lin$^{*}$  & \wstdmy{59.88}{0.39}  &\wstdmy{65.42}{0.47} & \wstdmy{67.76}{0.53} & \wstdmy{68.59}{0.19} & \wstdmy{69.42}{0.67} &  \wstdmy{69.90}{0.30} & \wstdmy{72.13}{0.40} \\\hline

\multirow{2}{*}{\begin{tabular}[c]{c@{}c@{}} Non-Retrain\\ LeRF\end{tabular}} & fix & \wstdmy{37.12}{0.03}     &\wstdmy{41.63}{0.03} & \wstdmy{42.32}{0.05} & \wstdmy{43.99}{0.07} & \wstdmy{46.96}{0.12} & \wstdmy{57.85}{0.02} &  \wstdmy{69.62}{0.02}\\\cline{3-9}

&lin$^{*}$ & \wstdmy{35.87}{0.04}     &\wstdmy{49.21}{0.04} & \wstdmy{55.18}{0.08} & \wstdmy{58.04}{0.03} & \wstdmy{59.75}{0.02} & \wstdmy{63.71}{0.04} &  \wstdmy{71.85}{0.02}\\ \hline
\end{tabular}}
\caption{Mean accuracy at each $\eta$ by using GB-SG in all methods with standard deviations of five individual runs. For LeRF, the accuracy is at (1-$\eta$).}
\label{tab:dev gb-sg}
\end{table}

\subsection{Masking}
\begin{figure} 
\captionsetup[subfigure]{justification=centering}
    \centering
 \begin{subfigure}[b]{0.45\textwidth}
    \centering 
    \includegraphics[width=\textwidth]{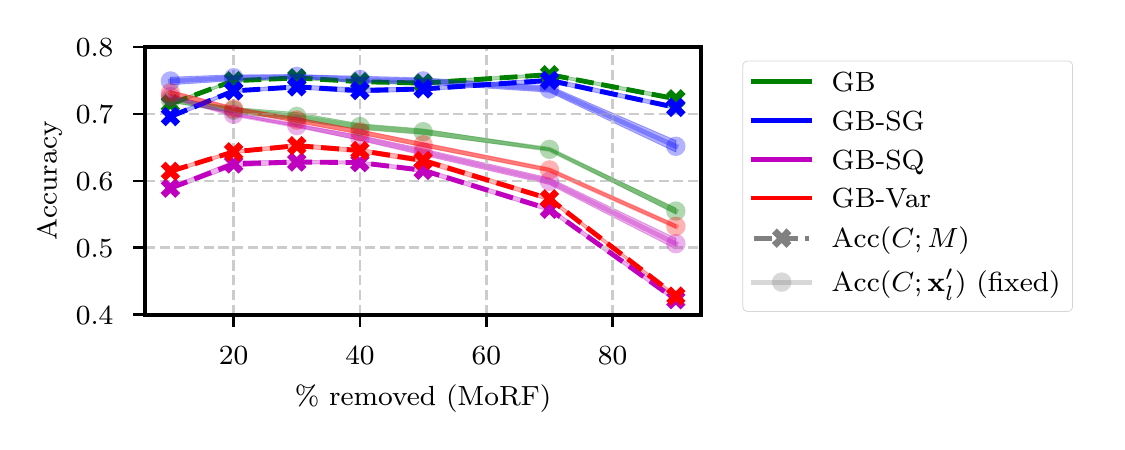}
    \label{fig:igleak}
    \caption{Mask Information and MoRF scores (background) for IG}
     \end{subfigure}
    \hfill
   \begin{subfigure}[b]{0.45\textwidth}
    \centering 
    \includegraphics[width=\textwidth]{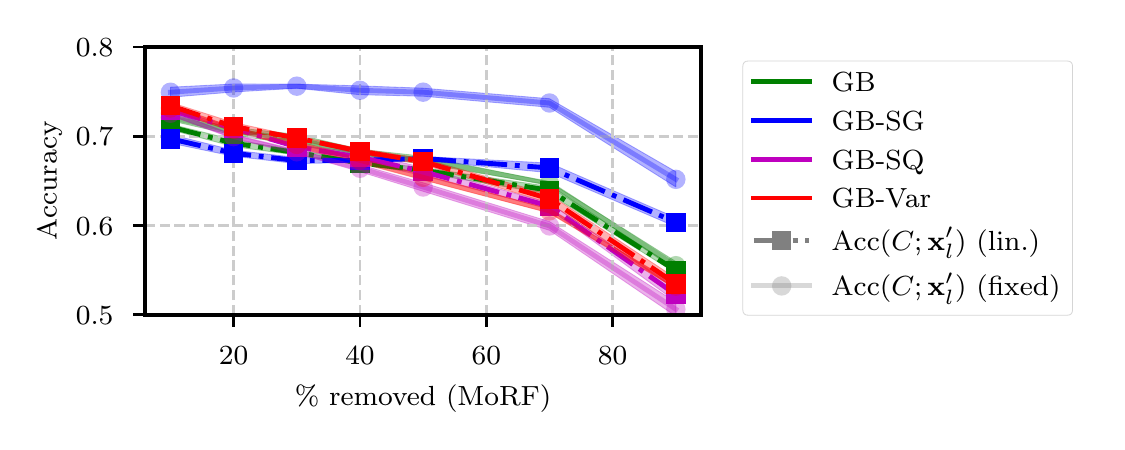}
    \label{fig:igleakinfill}
    \caption{MoRF scores with linear imputation and fixed imputation (background) for IG}
     \end{subfigure}
       \begin{subfigure}[b]{0.45\textwidth}
  \centering
     \includegraphics[width=\textwidth]{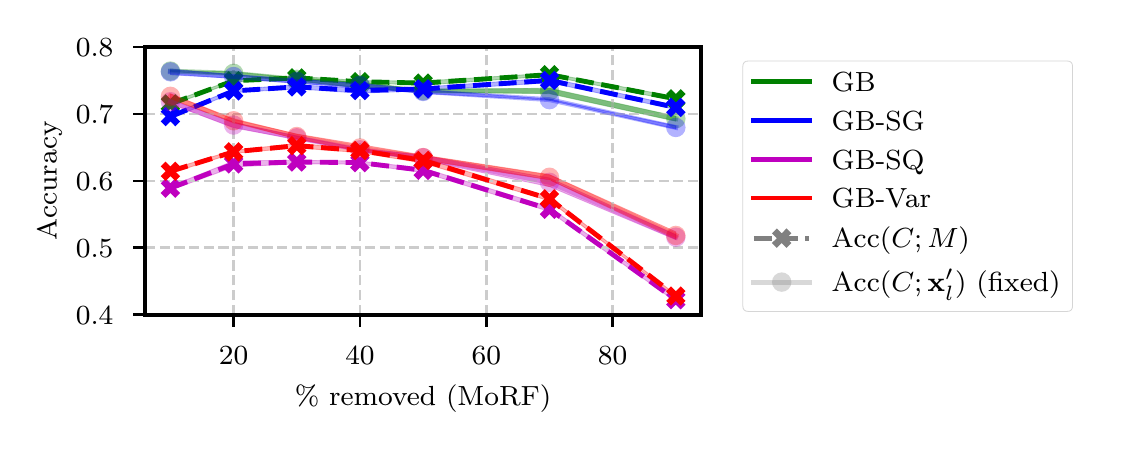}
     \caption{Mask Information and MoRF scores with fixed imputation (background) for GB}
    \label{fig:gbleak}
  \end{subfigure}
    \hfill
  \begin{subfigure}[b]{0.45\textwidth}
  \centering
     \includegraphics[width=\textwidth]{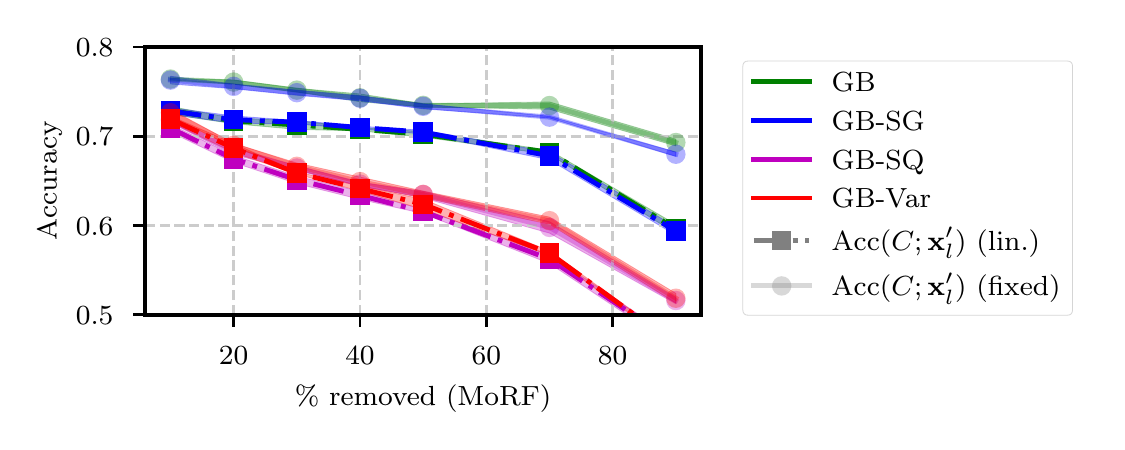}
    \label{fig:gbleakinfill}
      \caption{MoRF scores with linear imputation and fixed imputation (background) for GB}
  \end{subfigure}
  \caption{\textbf{Left}: Accuracy $\acc (C;\mask)$ obtainable only using the mask and MoRF scores (retraining) with fixed imputation. We observe that the mask-only accuracy highly corresponds to the accuracy of the retaining benchmarks. \textbf{Right}: Accuracies obtained with linear imputation vs MoRF scores (retraining) with fixed imputation. The linear imputation does not allow for mask leakage and compensates this effect. We observe that without the mask leakage, the curves lie below their couterparts and in particular for IG, their order also changes.}
  \label{fig:masking}
\end{figure}

We show the accuracy obtained using only the bitmask in \cref{fig:masking} for IG and GB. On the right side, we show the corresponding curves after linear imputation. We observe that the imputation strategy makes a considerable difference in the results and decouples the accuracy from those obtained with the bitmask only.



\subsection{ROAD Benchmark}
We extend the analysis of the ROAD benchmark and its correspondence to other benchmarks by providing the full table of Spearman rank correlations in \cref{tab:spearman}. To obtain the standard errors reported, we use a single run of each benchmark and compute the correlation coefficient. We do this $N=5$ times and compute the mean and standard deviations of the coefficients. Note that in the paper, we first took the mean of the benchmark results and computed the rank correlation on the mean to obtain a more stable estimate but no standard deviation. Thus the mean results reported in the table differ minimally to those in the main paper. 
\begin{figure*}[h]
\centering
   \includegraphics[width=.5\linewidth]{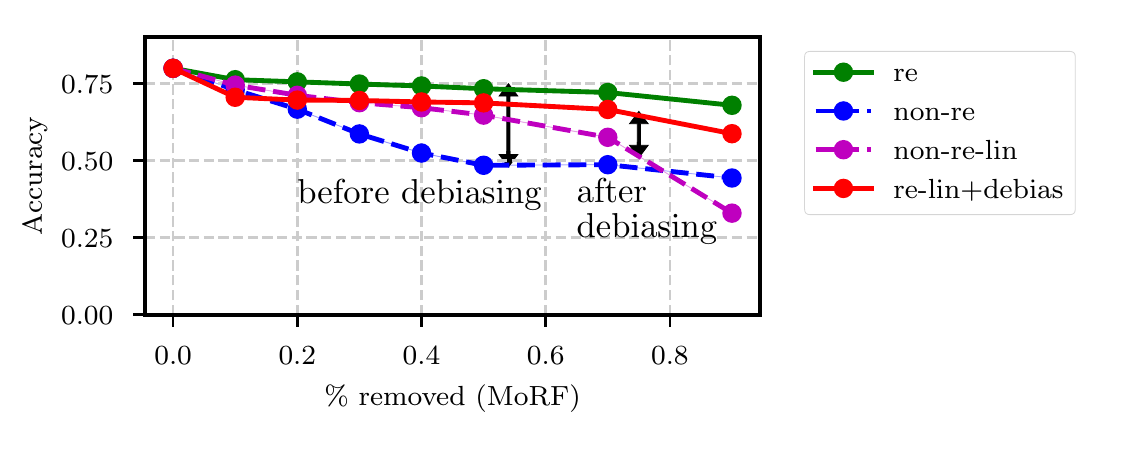}
    \caption{Comparison of using our debiasing strategies in retraining and non-retraining schemes on GB-SG. They reduces the difference gap between retraining and non-retraining schemes significantly except for the very end of the curve, where extremely few pixels are present.}
    \label{fig:comparison_gb}
    \vspace{-15pt}
\end{figure*}

We report the run times again in \cref{tab:runtimes} with their standard errors for completeness. 
\newcommand{\wstd}[2]{\makecell{#1\small\\$\pm$#2}}
\begin{table}
\centering
\scalebox{0.7}{\begin{tabular}{|c|c|ccc|cc|ccc|cc|}
\cline{3-12}
\multicolumn{2}{c|}{\multirow{3}{*}{}} & \multicolumn{3}{c|}{Retrain} & \multicolumn{2}{c|}{Non-Retrain} & \multicolumn{3}{c|}{Retrain} & \multicolumn{2}{c|}{Non-Retrain}\\
\multicolumn{2}{c|}{}  & \multicolumn{3}{c|}{MoRF} & \multicolumn{2}{c|}{MoRF} & \multicolumn{3}{c|}{LeRF} & \multicolumn{2}{c|}{LeRF}\\ 

\cline{3-12} \multicolumn{2}{c|}{} & fix$^\dagger$    & lin   & lin+debias & lin$^{*}$ & fix &  fix    & lin   & lin+debias & lin & fix \\ 
\hline
\multirow{3}{*}{\begin{tabular}[c]{c@{}c@{}} Retrain\\ MoRF\end{tabular}} & fix$^\dagger$ &  \wstd{1.00}{0.00} && && &&& &&\\
&lin & \wstd{0.69}{0.02}     & \wstd{1.00}{0.00} & && &&& &&\\
&lin+deb & \wstd{0.64}{0.02}   &  \wstd{0.97}{0.01} & \wstd{1.00}{0.00} && &&& &&\\
\hline
\multirow{2}{*}{\begin{tabular}[c]{c@{}c@{}} Non-Retrain\\ MoRF\end{tabular}} & lin$^{*}$ &  \wstd{\textbf{0.65}}{0.01} &  \wstd{\textbf{0.86}}{0.01}   & \wstd{\textbf{0.82}}{0.01} & \wstd{1.00}{0.00} & &&& &&\\
& fix  & \wstd{\textbf{0.14}}{0.02} & \wstd{\textbf{0.41}}{0.02} &  \wstd{\textbf{0.50}}{0.02}   & \wstd{0.45}{0.01}  & \wstd{1.00}{0.00} &&& &&\\
\hline
\multirow{3}{*}{\begin{tabular}[c]{c@{}c@{}} Retrain\\ LeRF\end{tabular}} & fix &  \wstd{\textbf{-0.09}}{0.02}& \wstd{0.41}{0.02} &  \wstd{0.49}{0.03}   & \wstd{0.33}{0.01}  & \wstd{0.58}{0.01} & \wstd{1.00}{0.00} && &&  \\
&lin & \wstd{0.14}{0.01}& \wstd{\textbf{0.57}}{0.01} &  \wstd{0.66}{0.01}   & \wstd{0.52}{0.01}  & \wstd{0.74}{0.01}  & \wstd{0.86}{0.01} &\wstd{1.00}{0.00} & &&\\
&lin+deb &\wstd{0.14}{0.01}& \wstd{0.58}{0.01} &  \wstd{\textbf{0.68}}{0.02}   & \wstd{0.55}{0.01}  & \wstd{0.77}{0.01} & \wstd{0.85}{0.01} & \wstd{0.99}{0.01} & \wstd{1.00}{0.00}  &&\\ 
\hline
\multirow{2}{*}{\begin{tabular}[c]{c@{}c@{}} Non-Retrain\\ LeRF\end{tabular}} & lin  & \wstd{0.19}{0.01}  &  \wstd{0.59}{0.01}& \wstd{0.67}{0.01}  & \wstd{\textbf{0.53}}{0.01} & \wstd{0.79}{0.01} & \wstd{0.82}{0.01} & \wstd{0.95}{0.01} &  \wstd{0.95}{0.01} & \wstd{1.00}{0.00} &\\
&fix &\wstd{0.49}{0.01}  &  \wstd{0.53}{0.01}& \wstd{0.44}{0.01}  & \wstd{0.68}{0.01} & \wstd{\textbf{0.01}}{0.01} & \wstd{0.05}{0.01} & \wstd{0.13}{0.01} &  \wstd{0.14}{0.01} & \wstd{0.18}{0.01} & \wstd{1.00}{0.00}\\ \hline
\end{tabular}}
\caption{Rank Correlations between all methods used with standard deviations computed by considering the rankings obtained through five consecutive runs as independent. Results indicated in bold correspond to those reported in the main paper. The ROAR benchmark is marked by $^\dagger$ and our ROAD by $^{*}$.
\label{tab:spearman}}
\end{table}

\begin{table}[]
    \centering
    \resizebox{0.5\textwidth}{!}{ 
    \begin{tabular}{r|c|c|c|c}
        Method & \makecell{r-fix\\ (ROAR)} & r-lin & \makecell{non-r-lin\\ (ROAD)} &  non-r-fix\\
        \hline
        Time & 5986$\pm$5\,s & 6412$\pm$13\,s & 42.3$\pm$0.2\,s & 33.8$\pm$0.3\,s\\
        \hline
        Relative & 100\,\% &  107\,\% & 0.7\,\% & 0.6\,\%\\
    \end{tabular}
    }
    \caption{Runtimes for benchmarking a single explanation method (IG-Base) on the same hardware including standard deviations of $N{=}5$ runs.}
    \label{tab:runtimes}
\end{table}

\section{NOTATION}
Finally, we provide a short overview of our notation in \cref{tab:notation}.
\begin{table}[h]
    \centering
    \resizebox{.61
    \textwidth}{!}{ 
    \begin{tabular}{r|l}
        $c$ & Number of classes in the classification problem\\
        $C$ & class label random variable \\
        $d$ & Dimensionality of the input\\
        $\bm{e}$ & local feature attribution $\bm{e}\left(f, \bm{x} \right)$ \\
        $f$ & Non-linear prediction model \\
        $H$ & Entropy of a discrete random variable\\
        $k$ & Number of features removed\\
        $I$ & Mutual information \\
        $\scatterOP$ & Imputation operator (redistributes features to their positions and fills in gaps)\\
        $\mask$ & Binary mask in $\left\{0,1\right\}^d$\\
        $\mathcal{M}$ & Mask selection operator (takes out relevant features)\\$\bm{x}$ & Input features in $\mathbb{R}^d$ \\
        $\xl$ & Low importance features only in $\mathbb{R}^{k-d}$ \\
        $\xlp$ & Imputed Low importance features in $\mathbb{R}^{d}$ \\
        $\xh$ & High importance features only in $\mathbb{R}^{k}$ \\ 
        $\xhp$ & Imputed High importance features in $\mathbb{R}^{d}$ \\
    \end{tabular}
    }
    \caption{Overview of the notation used in this work.\label{tab:notation}}
\end{table}
\typeout{}
\bibliographystyle{apalike}
\bibliography{main}